\documentclass{article}

\usepackage{microtype}
\usepackage{graphicx}
\usepackage{subfig}
\usepackage{booktabs} % for professional tables
\usepackage[ruled, vlined]{algorithm2e}
\usepackage{placeins}
\usepackage{hyperref}

\usepackage[accepted]{icml2021}

\icmltitlerunning{Intermediate Layer Optimization for Inverse Problems}
\usepackage{amsthm}
\usepackage{amsfonts}
\usepackage{amssymb}
\usepackage{pifont}
\usepackage{amsmath}
\usepackage{multirow}
\usepackage{amssymb}
\usepackage{caption}
\usepackage{subfig}
\usepackage{multirow}
\usepackage{graphicx}
\usepackage[export]{adjustbox}
\newtheorem{theorem}{Theorem}
\newtheorem{lemma}{Lemma}
\newtheorem{corollary}{Corollary}
\newtheorem{definition}{Definition}
\newtheorem{remark}{Remark}

\usepackage{cleveref}
\newenvironment{delayedproof}[1]
 {\begin{proof}[\raisedtarget{#1}Proof of \Cref{#1}]}
 {\end{proof}}
\newcommand{\raisedtarget}[1]{%
  \raisebox{\fontcharht\font`P}[0pt][0pt]{\hypertarget{#1}{}}%
}

\usepackage[abbreviations, symbols]{glossaries-extra}
\glsxtrnewsymbol[description={$\delta$-cover of $\Theta$ w.r.t. $||\cdot||_q$ norm}]{N}{$N(\delta$, $\Theta$, $||\cdot||_q$)}
\glsxtrnewsymbol[description={$\delta$-packing of $\Theta$ w.r.t. $||\cdot||_q$ norm}]{M}{$M(\delta$, $\Theta$, $||\cdot||_q$)}
\glsxtrnewsymbol[description={$k$-dimensional ball of radius $r$ w.r.t. $||\cdot ||_q$ norm}]{B}{$B_q^k(r)$}
\glsxtrnewsymbol[description={Minkowski sum of the sets $S_1, S_2$, i.e. the set $\{x + y | x \in S_1, y \in S_2\}$}]{Minkowski}{$S_1\oplus S_2$}
\glsxtrnewsymbol[description={Gaussian complexity of set $T$}]{Gaussian Complexity}{$\mathcal G(T)$}
\glsxtrnewsymbol[description={Set $\{1, ..., M\}$}]{Set}{$[M]$}
\makeglossary

\newcommand{\xmark}{\ding{55}}
\newcommand{\R}{\mathbb R}
\newcommand{\N}{\mathcal N}
\newcommand{\ballpdr}[3]{B_{#1}^{#2}(#3)}
\renewcommand{\wp}{\textrm{w.p.}}
\newcommand{\sgn}{\textrm{sgn}}
\newcommand{\vol}{\mathrm{vol}}
\newcommand{\argmin}{\mathrm{argmin}}

\newcommand{\norm}[1]{\| #1 \|}

\DeclareCaptionFont{tpt}{\fontsize{25pt}{11pt}\selectfont #1}
\DeclareCaptionFont{ltpt}{\fontsize{35pt}{11pt}\selectfont #1}
\begin{document}

\twocolumn[
\icmltitle{Intermediate Layer Optimization \\ for Inverse Problems using Deep Generative Models}



\icmlsetsymbol{equal}{*}

\begin{icmlauthorlist}
\icmlauthor{Giannis Daras}{equal,to}
\icmlauthor{Joseph Dean}{equal,to}
\icmlauthor{Ajil Jalal}{to}
\icmlauthor{Alexandros G. Dimakis}{to}
\end{icmlauthorlist}

\icmlaffiliation{to}{The University of Texas at Austin}
\icmlcorrespondingauthor{Giannis Daras}{giannisdaras@utexas.edu}
\icmlcorrespondingauthor{Joseph Dean}{josephdean98@utexas.edu}
\icmlcorrespondingauthor{Ajil Jalal}{ajiljalal@utexas.edu}
\icmlcorrespondingauthor{Alexandros G. Dimakis}{dimakis@austin.utexas.edu}


\vskip 0.3in]



\printAffiliationsAndNotice{\icmlEqualContribution}

\begin{abstract}
We propose \textbf{I}ntermediate \textbf{L}ayer \textbf{O}ptimization (\textbf{ILO}), a novel optimization algorithm for solving inverse problems with deep generative models. Instead of optimizing only over the initial latent code, we progressively change the input layer obtaining successively more expressive generators. To explore the higher dimensional spaces, our method searches for latent codes that lie within a small $l_1$ ball around the manifold induced by the previous layer. Our theoretical analysis shows that by keeping the radius of the ball relatively small, we can improve the established error bound for compressed sensing with deep generative models. We empirically show that our approach outperforms state-of-the-art methods introduced in StyleGAN-2 and PULSE for a wide range of inverse problems including  inpainting, denoising, super-resolution and compressed sensing.
\end{abstract}

\section{Introduction}

We study how deep generators can be used as priors to solve inverse problems like inpainting, super-resolution, denoising
and compressed sensing from random projections. 
Image reconstruction methods can be either supervised~\cite{pathak2016context, richardson2020encoding, yu2018generative} or unsupervised~\cite{pulse, bora2017compressed, pajot2018unsupervised}, see the recent survey~\cite{ongie2020deep} for a unified presentation. 
Such inverse problems naturally appear in many applications including medical imaging, single pixel reconstruction and other domains~\cite{lustig2007sparse, lustig2008compressed, chen2008prior, duarte2008single, qaisar2013compressive, hegde2009compressive}.

We focus on unsupervised image reconstruction techniques
that rely on a pre-trained generator, building on the general framework introduced in CSGM~\cite{bora2017compressed}. 
The central optimization problem that appears in unsupervised image reconstruction is the inversion of a deep generative model, 
i.e. finding a latent code that explains the measurements. This can be performed for different generators, e.g. DCGAN or more recently the powerful StyleGAN-2~\cite{stylegan,stylegan2} as shown in the excellent results obtained by PULSE~\cite{pulse}. 
Unfortunately, inverting a generator with even 4 layers is NP-hard~\cite{lei2019inverting} so approximate inversion methods are needed.

The CSGM framework~\cite{ bora2017compressed} used gradient descent to minimize the measurement mean squared error (MSE) and showed good empirical performance for numerous inverse problems including inpainting and compressed sensing with random Gaussian measurements using DCGAN. However, this \textit{does not} work as well for deeper generators e.g. BigGAN as discussed in~\citet{ylg}.
PULSE~\cite{pulse} improved the CSGM framework focusing specifically on super-resolution, by refining the latent space optimization and using the  StyleGAN-2~\cite{stylegan, stylegan2} generator.

We propose a novel optimization method for solving general inverse problems using a technique we call \textbf{Intermediate Layer Optimization} (ILO). Our method adaptively changes which layer is optimized, moving from the initial latent code to intermediate layers closer to the pixels. By optimizing intermediate layers we expand the range of the generator to better satisfy the measurements. This has to be done carefully since intermediate layers can produce non-realistic images and therefore inversion must be regularized. 
% We experiment with different loss functions and utilize a perceptual loss combined with standard mean squared error. 

%\noindent \textbf{Our Contributions:}
\subsection{Our Contributions}
1. We propose a novel optimization method for solving general inverse problems by adaptively changing which layer
variables are optimized. Our method extends PULSE~\cite{pulse}
beyond super-resolution, to all inverse problems with differentiable forward operators.  \\
2. To avoid over-expanding the range of the generator to non-realistic images, we only search for latent codes within a small $l_1$ ball around the manifold induced by the previous layer. Conceptually, our method generalizes the framework introduced in~\citet{dhar2018modeling}; instead of allowing sparse deviations only in the image space, we allow small deviations from the manifold of any layer of the generator. \\
3. We theoretical analyze our framework by establishing sample complexity and error bounds.  We show that by restricting the radius of the latent searches, we can improve the established error bound of CSGM~\cite{bora2017compressed}. \\
4. Experimentally, our method significantly outperforms the previous state-of-the-art techniques for solving inverse problems with deep generative models for a wide range of tasks including inpainting, denoising and super-resolution.\\
5. To illustrate the power of inverse problems with general differentiable forward operators, we use a classifier as a measurement process. Specifically, we show how we can use a classifier to bias generators to produce human images that look like ImageNet classes like frogs, corals and goldfishes. Our method uses gradients from classifiers trained to achieve robustness to adversarial attacks as proposed in~\cite{santurkar2019image}, but guiding generative latent codes as opposed to pixels directly.\\
6. We open-source all our code to encourage further research in this area: \href{https://github.com/giannisdaras/ilo}{https://github.com/giannisdaras/ilo}. A demo of our code is available \href{https://colab.research.google.com/drive/1qbDHqOGpLH_F5k2HAmU5Xt4YKyYc4kc_?usp=sharing}{under this URL}.
\begin{figure*}[!htp]

% \captionsetup[subfigure]{labelformat=empty,font=Large,labelfont=Large}
\captionsetup[subfigure]{labelformat=empty,font=tpt}
\captionsetup{justification=centering}
% \captionsetup[subfloat]{farskip=2pt,captionskip=1pt}
\begin{center}
\begin{adjustbox}{width=0.9\columnwidth, center}
\begin{tabular}{cccccc}
\subfloat{\includegraphics[height=2.5in, width=2.5in]{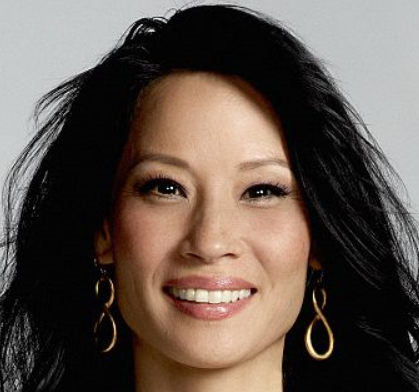}}& 
\subfloat{\includegraphics[width=2.5in]{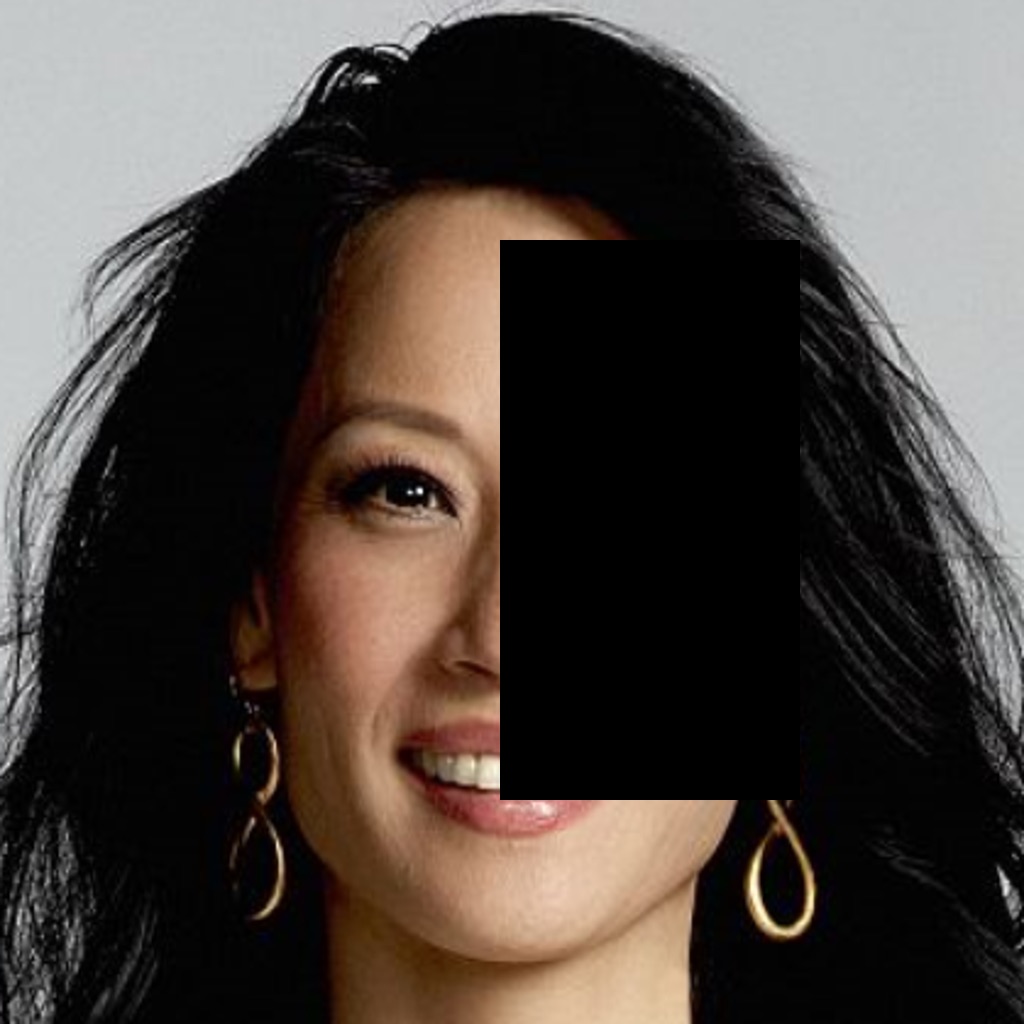}} & 
\subfloat{\includegraphics[width=2.5in]{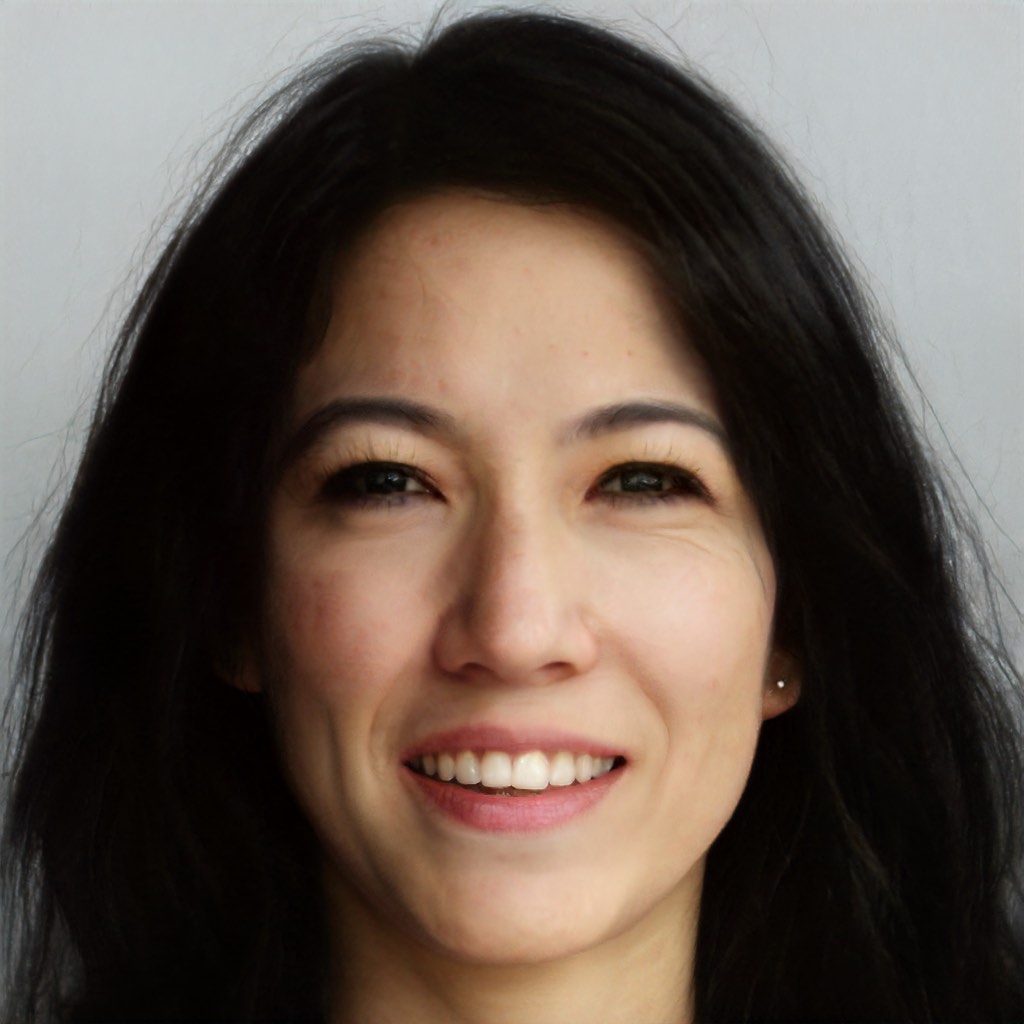}} & 
\subfloat{\includegraphics[width=2.5in]{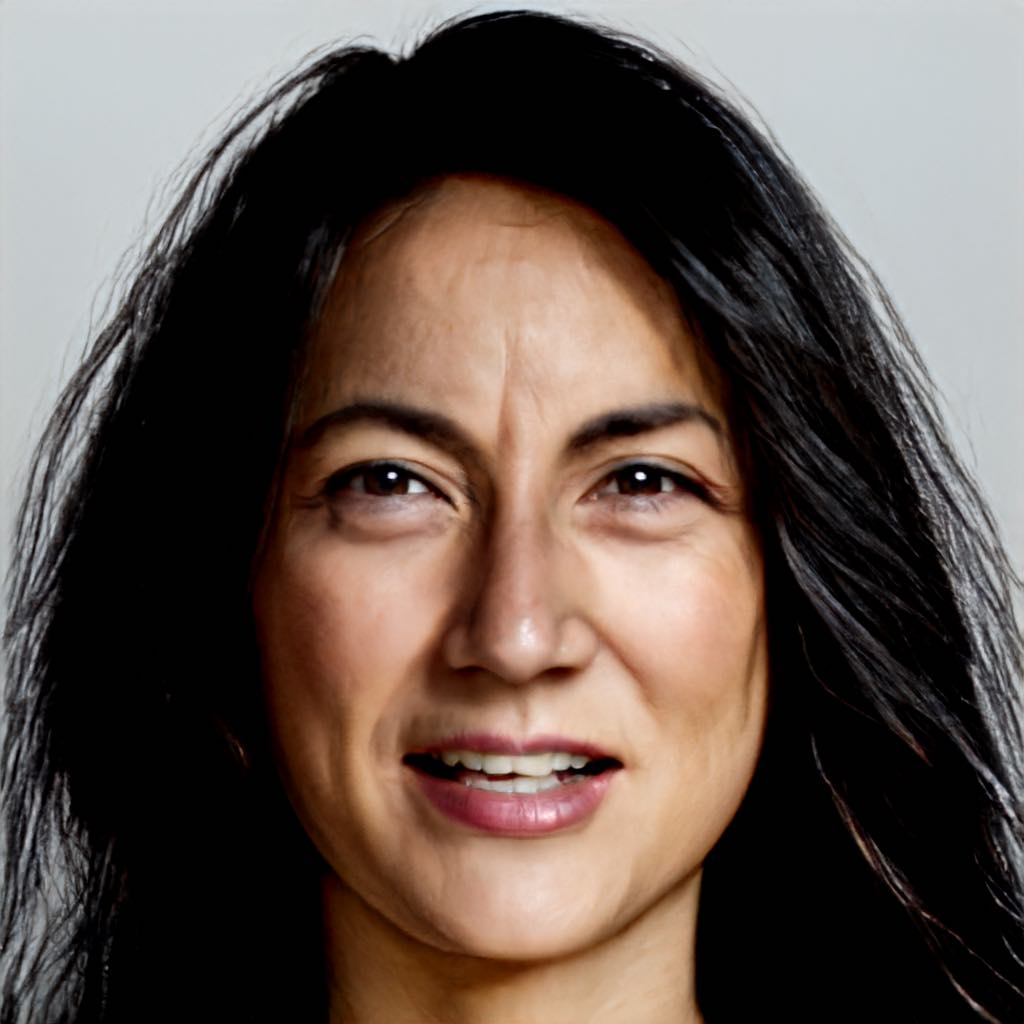}} & 
\subfloat{\includegraphics[width=2.5in]{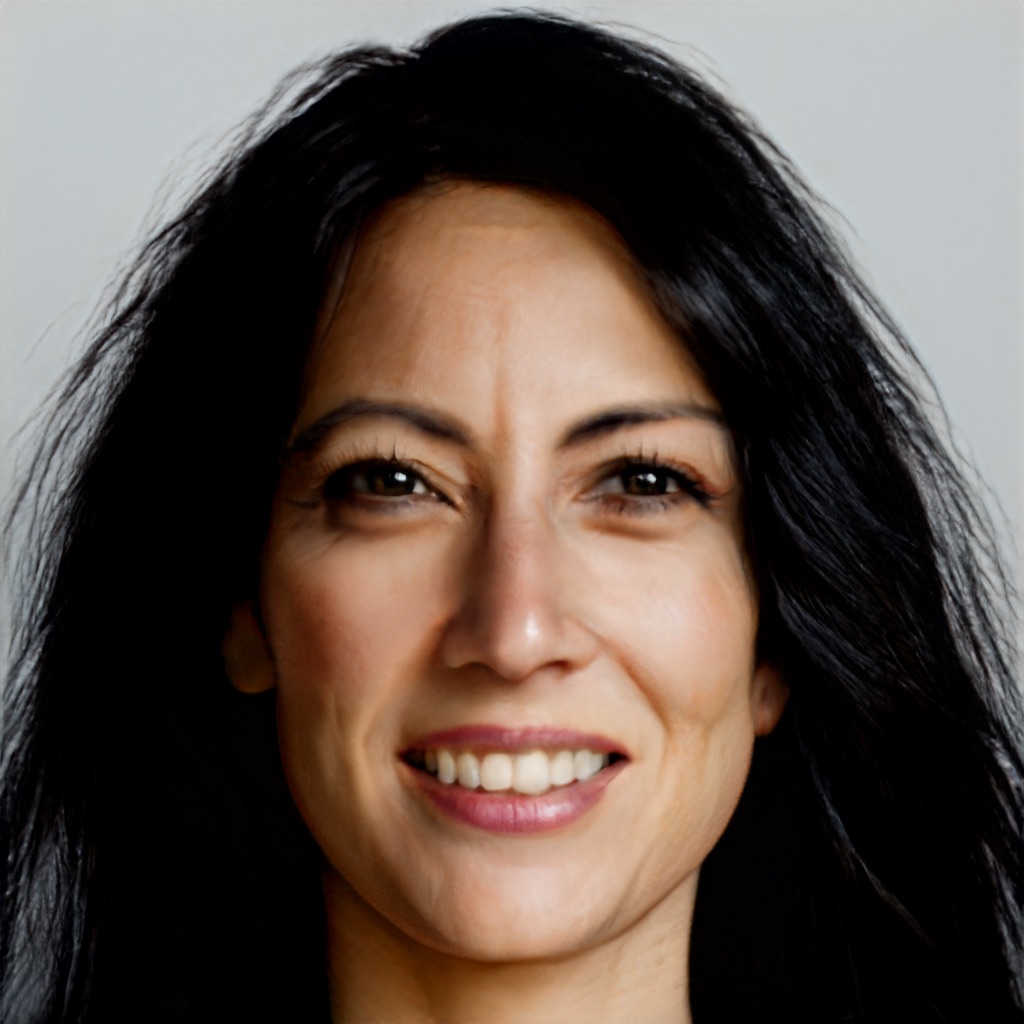}} & 
\subfloat{\includegraphics[width=2.5in]{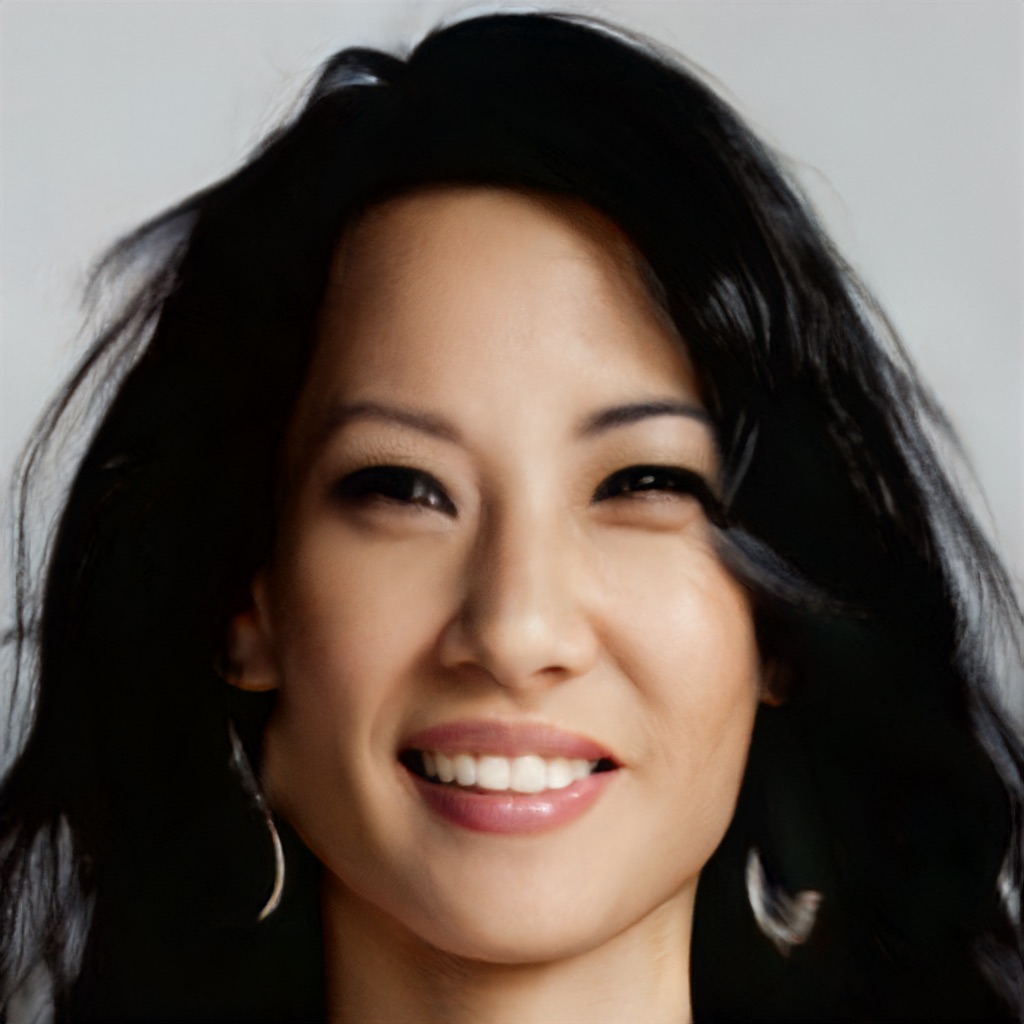}}\\[-2ex]

\subfloat{\includegraphics[width=2.5in]{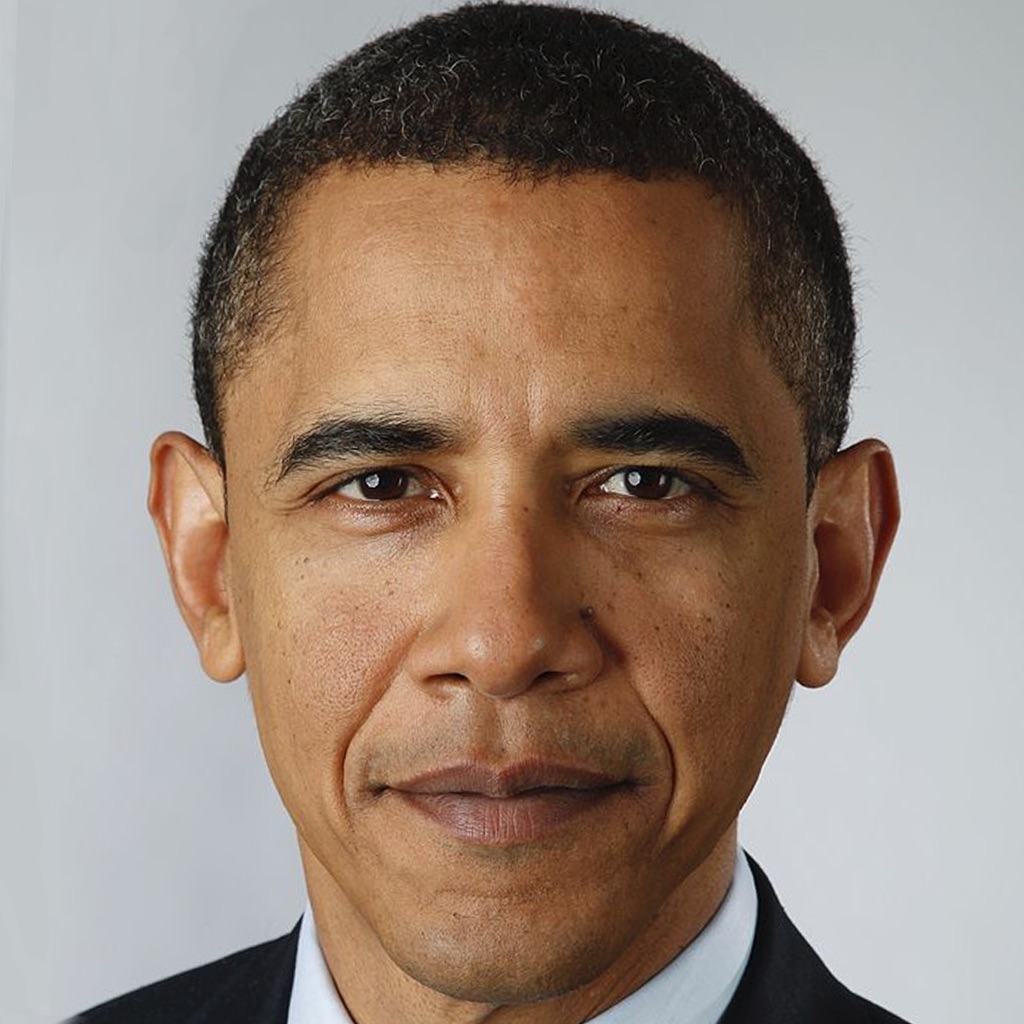}} & 
\subfloat{\includegraphics[width=2.5in]{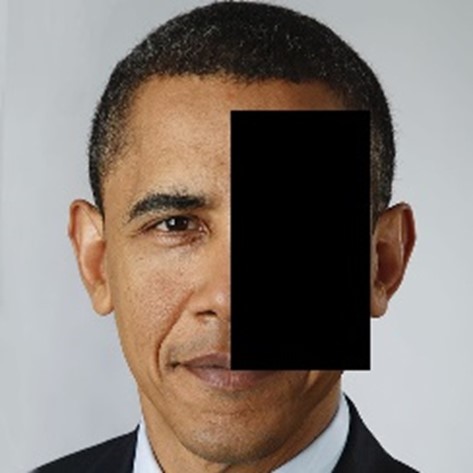}} & 
\subfloat{\includegraphics[width=2.5in]{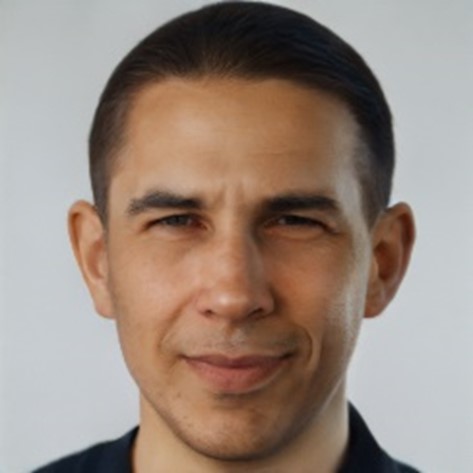}} & 
\subfloat{\includegraphics[width=2.5in]{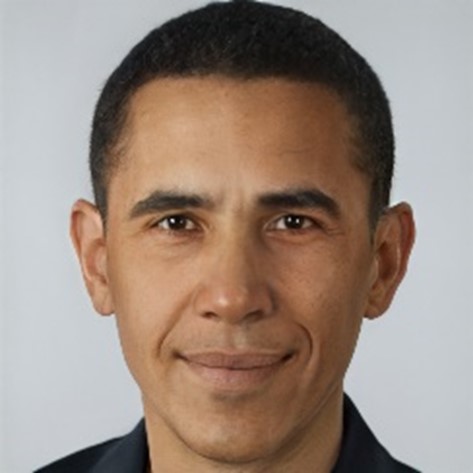}} & 
\subfloat{\includegraphics[width=2.5in]{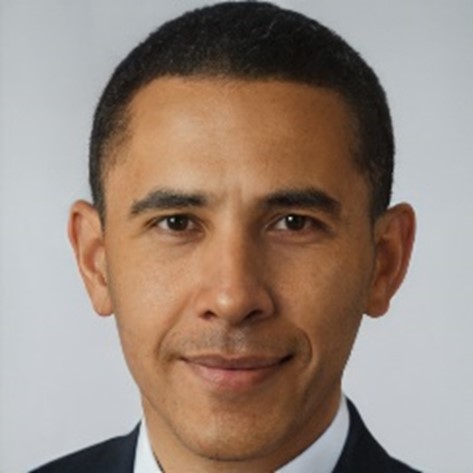}} & 
\subfloat{\includegraphics[width=2.5in]{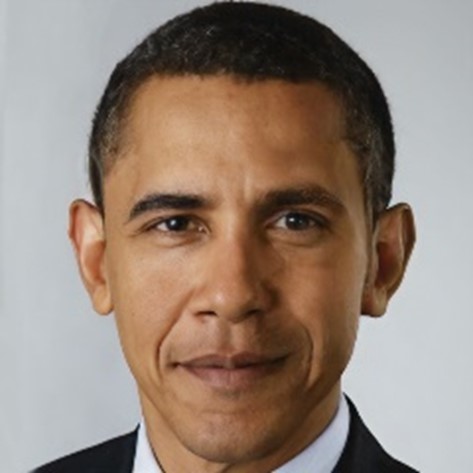}}\\[-2ex]

\subfloat{\includegraphics[width=2.5in]{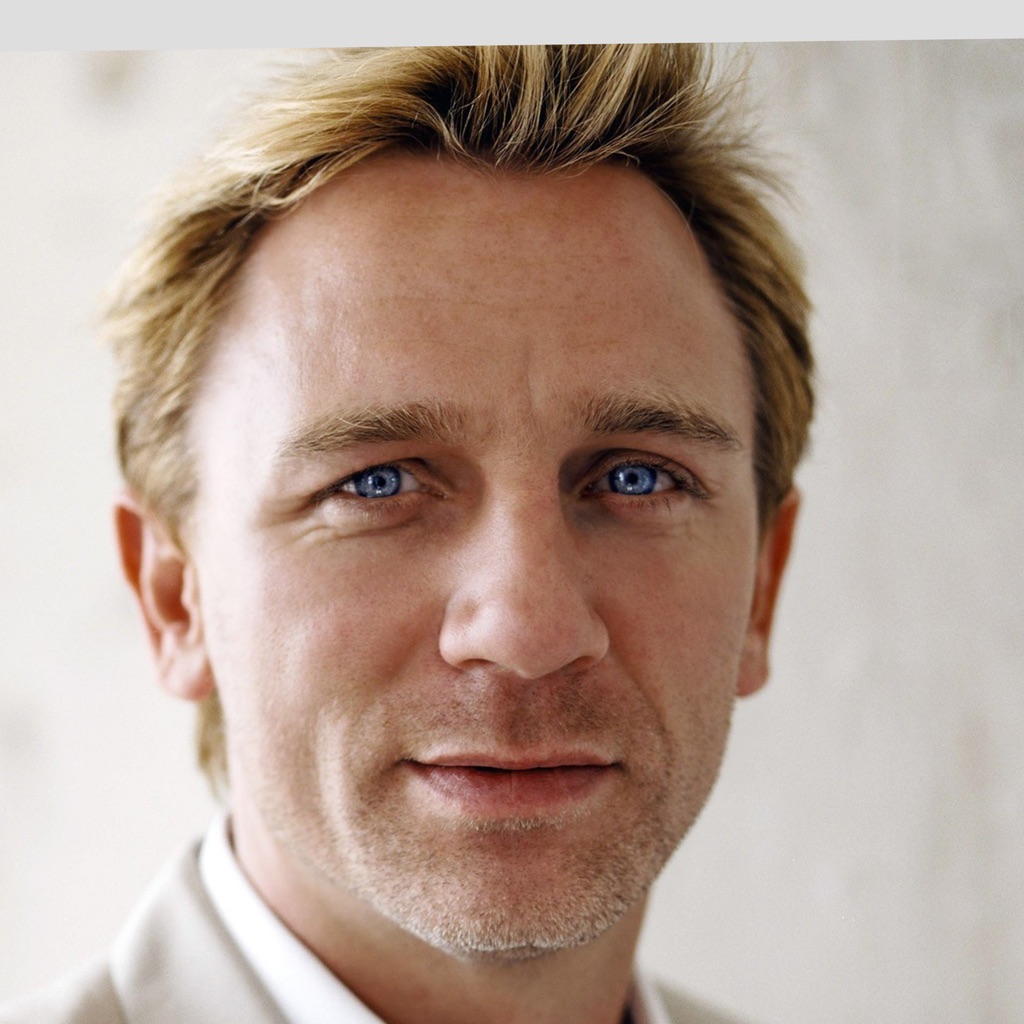}} & 
\subfloat{\includegraphics[width=2.5in]{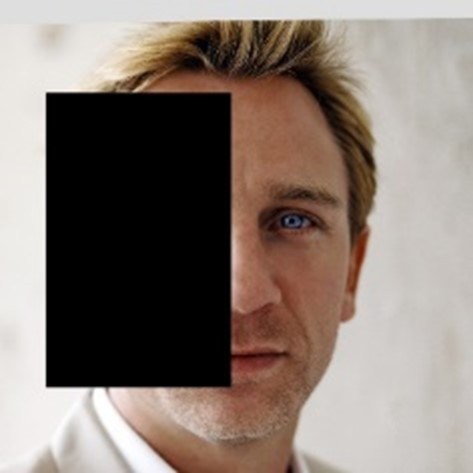}} & 
\subfloat{\includegraphics[width=2.5in]{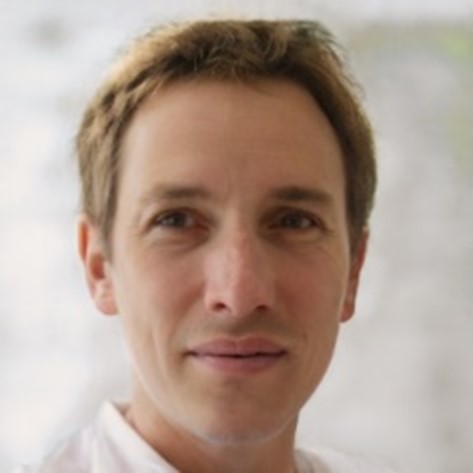}} & 
\subfloat{\includegraphics[width=2.5in]{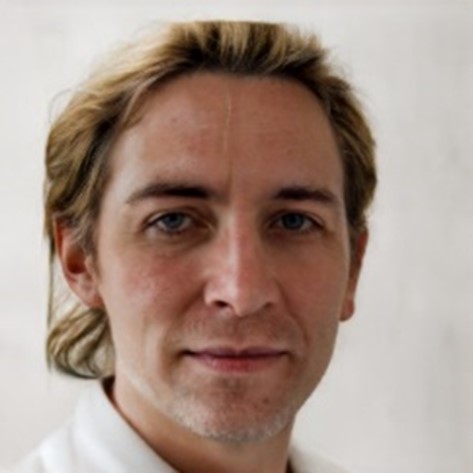}} & 
\subfloat{\includegraphics[width=2.5in]{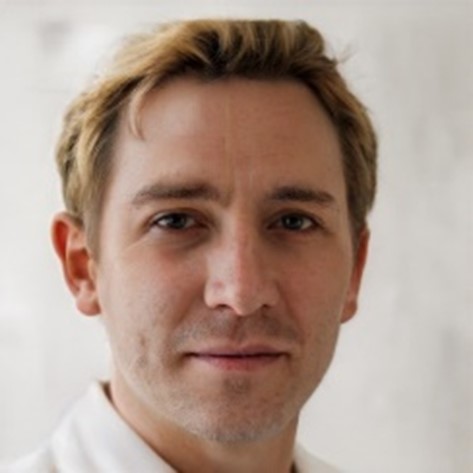}} & 
\subfloat{\includegraphics[width=2.5in]{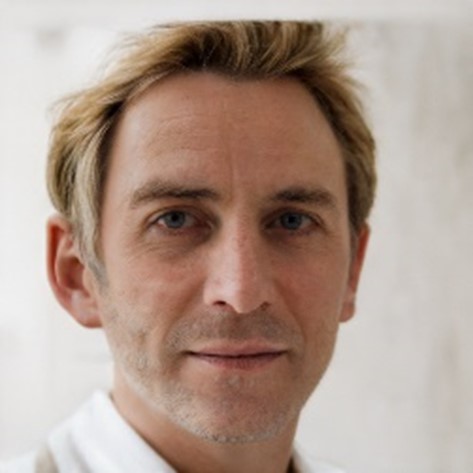}}\\[-2ex]

\subfloat{\includegraphics[width=2.5in]{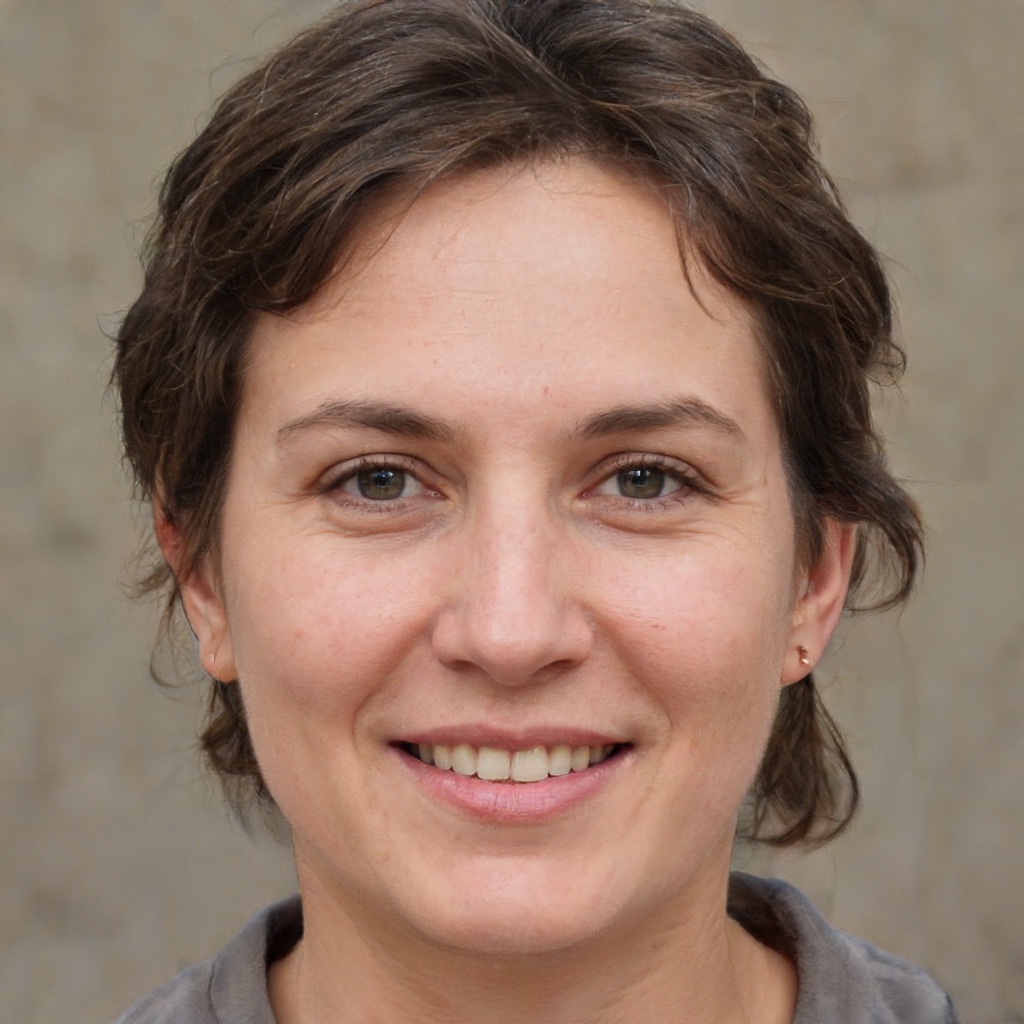}} &
\subfloat{\includegraphics[width=2.5in]{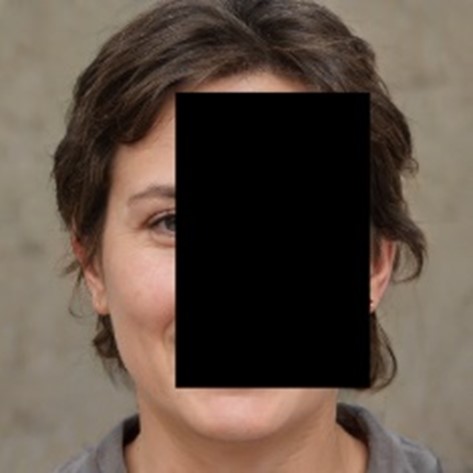}} &
\subfloat{\includegraphics[width=2.5in]{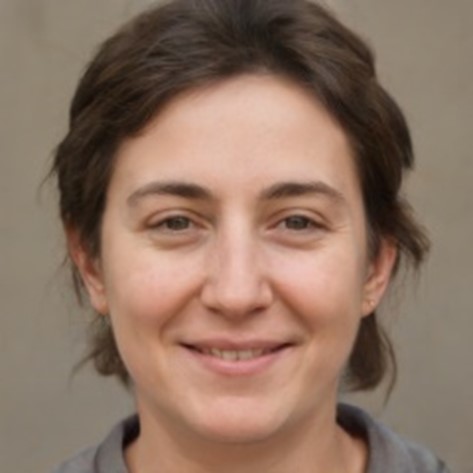}} &
\subfloat{\includegraphics[width=2.5in]{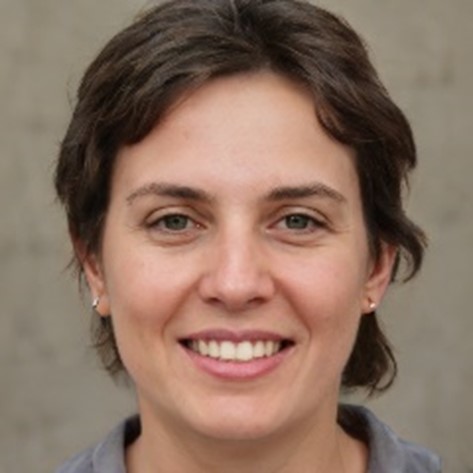}} &
\subfloat{\includegraphics[width=2.5in]{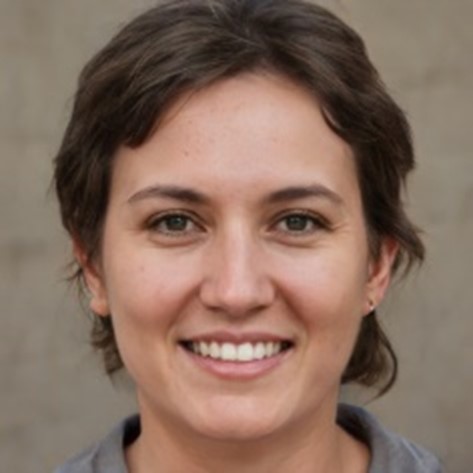}} &
\subfloat{\includegraphics[width=2.5in]{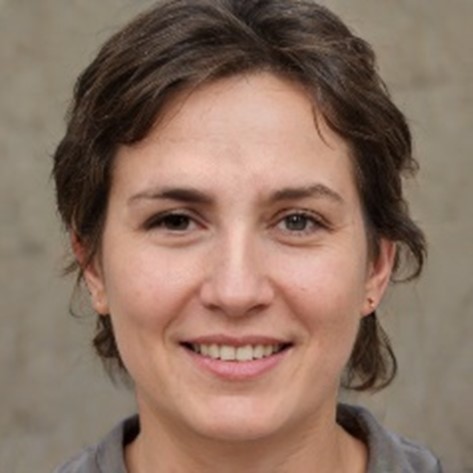}}\\[-2ex]

\subfloat{\includegraphics[width=2.5in]{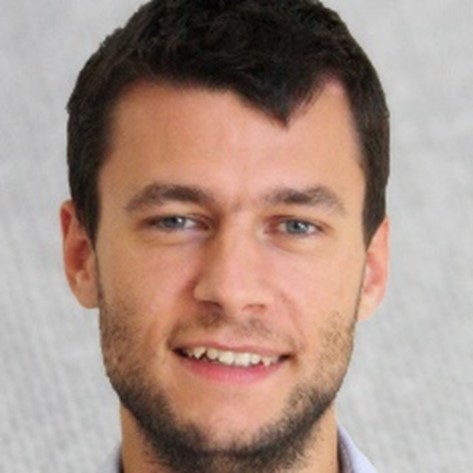}} & 
\subfloat{\includegraphics[width=2.5in]{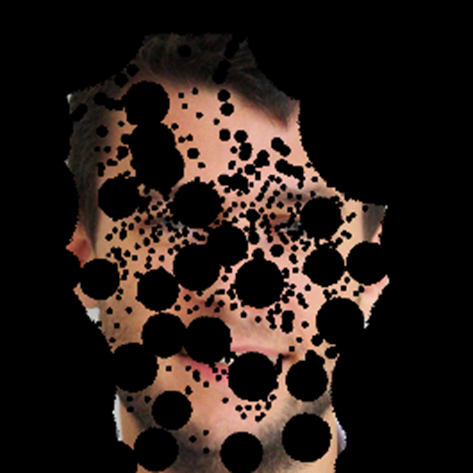}} & 
\subfloat{\includegraphics[width=2.5in]{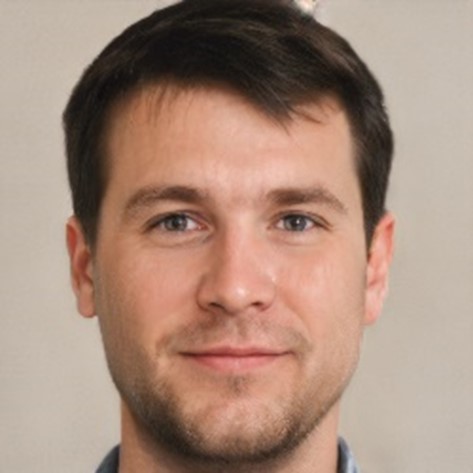}} & 
\subfloat{\includegraphics[width=2.5in]{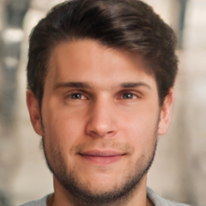}} & 
\subfloat{\includegraphics[width=2.5in]{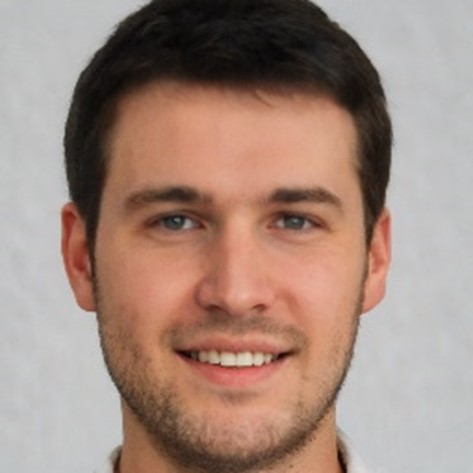}} & 
\subfloat{\includegraphics[width=2.5in]{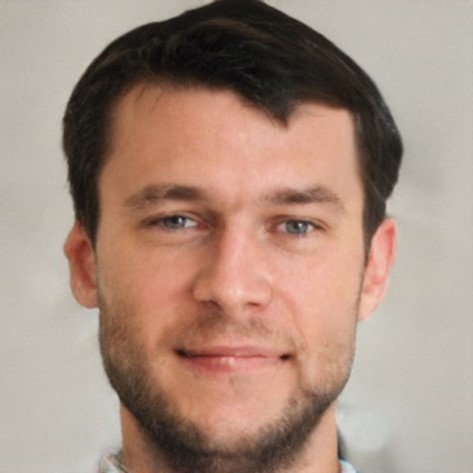}}\\[-2ex]

\subfloat[Original]{\includegraphics[width=2.5in]{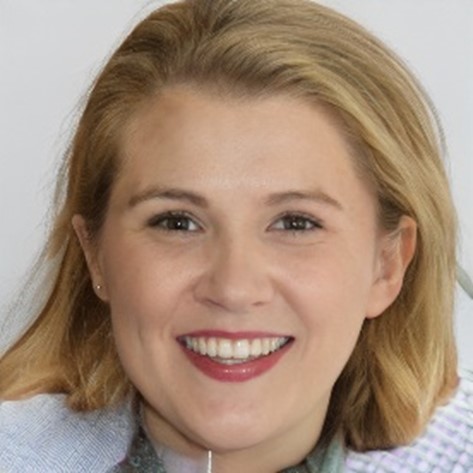}} &
\subfloat[Observation]{\includegraphics[width=2.5in]{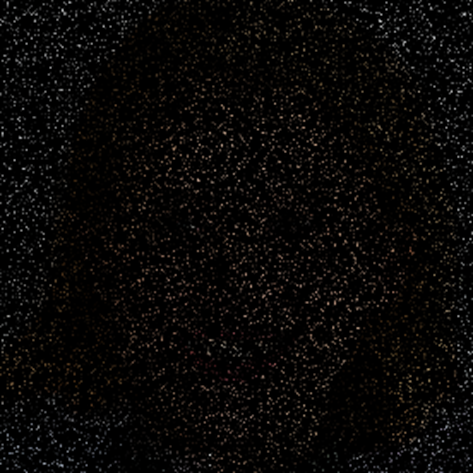}} &
\subfloat[CSGM MSE (PULSE)]{\includegraphics[width=2.5in]{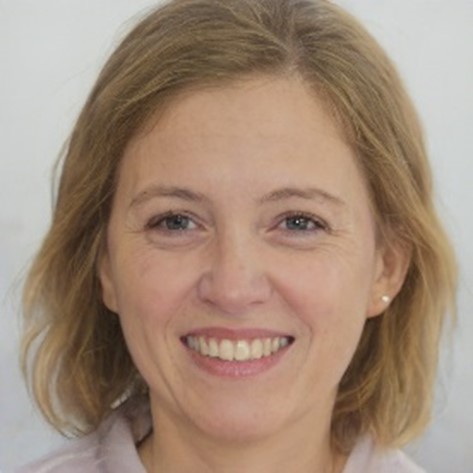}} &
\subfloat[CSGM LPIPS]{\includegraphics[width=2.5in]{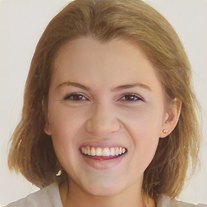}} & 
\subfloat[CSGM LPIPS+MSE]{\includegraphics[width=2.5in]{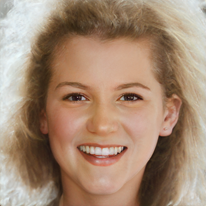}} &
\subfloat[Ours]{\includegraphics[width=2.5in]{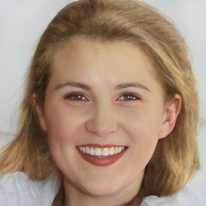}}\\
\end{tabular}
\end{adjustbox}
\caption{\footnotesize 
Results on the inpainting task. Rows 1, 2, 3 and 5 are real images (outside of the test set, collected from the web) while rows 4, 6 are StyleGAN-2 generated images.
Column 2: the first five images have masks that were chosen to
remove important facial features. The last row is an example of randomized inpainting, i.e. a random $1\%$ of the total pixels is observed. Columns 3-5: reconstructions using the CSGM~\cite{bora2017compressed} algorithm with the StyleGAN-2 generator and the optimization setting described in PULSE~\cite{pulse}. While PULSE only applies to super-resolution, we extend it using MSE, LPIPS and jointly MSE+LPIPS loss. The experiments of Columns 3-5 form an ablation study of the benefits of each loss function. Column 6: reconstructions with ILO (ours). As shown, ILO consistently gives better reconstructions of the original image. Also, many biased reconstructions can be corrected by our method. In the last two rows, recovery of the image is still possible from very few pixel observations using our method.}
\label{inpainting_res}
\end{center}
\end{figure*}

\section{Algorithm}
\subsection{Setting}
The key step in our approach is to decompose pre-trained generative models as compositions of feed-forward neural networks. Given a (pre-trained) generative model $G(z) \in \R^n$ that produces images from latent codes $z \in \R^k$, we decompose it as a $G = G_2 \circ G_1$ where $G_1: \R^k \to \R^p$ and $G_2: \R^p \to \R^n$. As usual, the latent vectors $z^k \in \R^k$ 
were sampled according to a simple distribution $P_z$, typically Gaussian and independent. 

Our observations are formed by a known measurement matrix
\begin{gather}
y=Ax + \textit{noise},
\end{gather}
where $A: \R^{m \times n}$ where $x \in \R^n$ is the real image we want to recover. We emphasize that our algorithm can be applied when the measurement process is a general differentiable operator $y=\mathcal{A}(x)$ but our theory only applies to linear inverse problems. Since we will be working with latent vectors in different layers we indicate the dimension as a superscript, so $z^k$ denotes an initial latent vector in $\R^k$ and $z^p$ an intermediate vector in $\R^p$.

\subsection{Approach}

Our approach is described in Algorithm \ref{optimization_algorithm}. The first step of our method is the same as in CSGM~\cite{bora2017compressed}; we optimize over a $k$-dimensional latent code, $z^k$, which is the input of the first layer of the generator. In practice, to obtain the solution of line 1 of Algorithm \ref{optimization_algorithm}, we pick an initial $z^k$ from the latent distribution of the generator and we optimize the loss function $||AG(z^k) - Ax||$ using gradient descent. Once we solve this optimization problem, we obtain a solution, $\hat z^k$, that we map to the $p$-dimensional space using $G_1$. By doing that, we get an intermediate latent representation, $\hat z^p = G_1(\hat z^k)$. 

From that point onwards, our algorithm proceeds in rounds. At the beginning of each round, we optimize on the $p$-dimensional input space of $G_2$ but we only allow solutions that lie within an $l_1$ ball centered at $\hat z^p$. Intuitively, we allow deviations from the range of $G_1$ to increase the expressitivity of the model, but we restrict those deviations to avoid overfitting on the measurements (see Experiments section). 

Once we obtain the solution of line 4 of Algorithm \ref{optimization_algorithm}, i.e. once we find the latent code, $\tilde z^p$, that best explains the measurements and lies inside an $l_1$ ball of the previous latent, we project this solution back to the range of the generator. To do that, we search for the latent code $z^k$ such that $G_1(z^k)$ is as close as possible to $\tilde z^p$ (line 5 of Algorithm \ref{optimization_algorithm}). This problem is solved by initializing a latent vector $z^p$ to $\hat z^p$ and then minimizing using gradient descent the loss $||G_1(z^k) - \tilde z^p||$. The solution of this problem forms a new $\hat z^k$ vector which is in turn projected again to the intermediate code $\hat z^p = G_1(\hat z^k)$. Our algorithm attempts to explore the set we call the \textit{extended range:} the range of vectors realizable by the previous layer, dilated by an $l_1$ ball of sparse deviations. Within this set we would like to find the latent vector that best explains the measurements. 

We emphasize that our theoretical analysis provides performance bounds for the global optimum in this extended range, while our algorithm is based on projected gradient descent for a non-convex problem and therefore can be stuck in local optima. 
It may be possible to prove that such local optimization algorithms obtain global minima under generator weight assumptions as achieved in the pioneering work of ~\citet{hand2018global,hand2018phase} for CSGM, but this remains open for future work.

%Our method attempts to explore the Minkowski sum of an $l_1$ ball with the range of $G_1$. In other words, it tries to find the intermediate latent vector that, when fed to $G_2$, explains best the measurements among all the other latent vectors that lie within an $l_1$ ball around any point of the range. The exploration of this set is done in a greedy fashion; we start exploring the set around the CSGM solution, which is guaranteed to lie in the range and with repetitive projections and local exploration we attempt to explore sparse deviations around the range. 

\section{Theoretical Analysis}

\subsection{Preliminaries}

We begin our theoretical discussion by revisiting some important elements of the theory of compressed sensing with deep generative models. 

\begin{definition}[S-REC~\cite{bora2017compressed}]
Let $S \subseteq \R^n$. For some parameters $\gamma, \delta > 0$, a matrix $A \in \R^{m \times n}$ is said to satisfy S-REC$(S, \gamma, \delta)$ if $\forall x_1, x_2 \in S$, we have that:
\begin{gather}
    ||A(x_1 - x_2)||_2 \geq \gamma ||x_1 - x_2||_2 - \delta.
    \label{srec}
\end{gather}
\end{definition}

The S-REC condition, introduced in CSGM~\cite{bora2017compressed}, guarantees that if two vectors, $x_1, x_2 \in \R^n$, are very different (right side of the equation), then their measurements will be significantly different as well (left side of the equation). In CSGM, the set $S$ of interest is the range of the generator. Therefore, S-REC is a key property for proving small reconstruction error when observing $Ax$. \citet{bora2017compressed} show that if 1) $A$ is a matrix with i.i.d. Gaussian entries drawn from $\mathcal N(0, \frac{1}{m})$ and 2) $m = \frac{1}{a^2}\Omega\left(k \log\left( \frac{L_1 \cdot L_2 r_1}{\delta}\right)\right)$, then with probability $1 - e^{-a^2\Omega(m)}$, S-REC$(G_2(G_1(B_1^k(r_1))), 1 - a, \delta))$ is satisfied.

\begin{algorithm}[]
\SetAlgoLined
\begin{small}
% \KwResult{Write here the result }
\tcp*[h]{CSGM solution} \\
\nl $\hat z^k \gets \argmin_{z^k \in \ballpdr{2}{k}{r_1}} ||AG(z^k) - Ax||_2$ \\
\nl $\hat z^p \gets G_1(\hat z^k)$ \\
\nl \For{$t\gets0$ \KwTo $r$}{
\tcp*[h]{Best solution within an $l_1$ ball centered around the prev. solution} \\
\nl $\tilde z^p \gets \argmin_{z^p \in \hat z^p \oplus \ballpdr{1}{p}{r_2}}||AG_2(z^p) - Ax||$ \newline
\tcp*[h]{Projection back to the range} \\
\nl $\hat z^k \gets \argmin_{z^k \in B_2^k(r_1)}||G_1(z^k) - \tilde z^p||$ \\
\nl $\hat z^p \gets G_1(\hat z^k)$
 }
\tcp*[h]{Return the best solution within an $l_1$ ball of some point in the range} \\
\nl \Return{$G_2(\tilde z^p)$}
 \caption{ILO for one layer of the generator}
 \label{optimization_algorithm}
 \end{small}
\end{algorithm}

\subsection{Intermediate Layer Optimization}

Our theoretical result is a sample complexity bound for the reconstruction algorithm that optimizes in the full extended range of the generative model, similar in style to the CSGM~\cite{bora2017compressed} result. 
 
Let $\ballpdr{q}{k}{r_1}$ denote a ball of radius $r_1$ measured in $l_q$ norm and $\oplus$ denote the Minkowski sum operation, i.e. given sets  $S_1, S_2$, the set
$$ S_1\oplus S_2= \{x + y | x \in S_1, y \in S_2\}.$$
If the initial vector $z^k$ lies in a ball of radius $r_1$, denoted as $\ballpdr{2}{k}{r_1}$, the range of the first generator is  $G_1(\ballpdr{2}{k}{r_1})$. We are expanding this set to create the \textit{extended range}:
$$
G_1(\ballpdr{2}{k}{r_1}) \oplus \ballpdr{1}{p}{r_2}.
$$
Our result is showing that minimizing the measurements in this extended range gives a reconstruction that is close to the best reconstruction that the extended generator $G_2$ can produce. This result is obtained with high probability over the random measurement matrix $A$, if the number of measurements is sufficiently large:

%--notes:
%$G_1(\ballpdr{2}{k}{r_1}) \oplus \ballpdr{1}{p}{r_2}||Ax - AG_2(z^p)||$
%--- 
 %\glsxtrnewsymbol[description={$k$-dimensional ball of radius $r$ w.r.t. $||\cdot ||_q$ norm}]{B}{$B_q^k(r)$}
%\glsxtrnewsymbol[description={Minkowski sum of the sets $S_1, S_2$, i.e. the set $\{x + y | x \in S_1, y \in S_2\}$}]{Minkowski}{$S_1\oplus S_2$}

 %are motivated by the previous discussion. Specifically, we observe that as we move to high dimensional latent spaces \textbf{1)} \textit{we cannot afford linear dependence on the dimension}, \textbf{2)} \textit{we can afford small radius and hence polynomial dependence on it} (since, worst case, an infinitesimal radius would recover the solution of CSGM).

% Print glossary
%\printunsrtglossary[type=symbols]

%\subsubsection{Theorem}
%We are now ready to introduce our main theorem.
\begin{theorem}
Let $G=G_2\circ G_1$ with $G_1:\R^k \to \R^p$ be an $L_1$-Lipschitz function and $G_2:\R^p \to \R^n$ be an $L_2$-Lipschitz function. Let $A \in R^{m\times n}$ be the measurements matrix with $A_{ij} \sim \N(0, 1/m)$ i.i.d. entries.

Let $K$ be a parameter of our choice 
where $K\leq \sqrt{p}$, and $r_2 = \frac{K \delta}{L_2}$.
Consider the true optimum in the extended range
\begin{gather}
    \bar z^p = \argmin_{z^p \in G_1(\ballpdr{2}{k}{r_1}) \oplus \ballpdr{1}{p}{r_2}}||x - G_2(z^p)||,
    \label{optimal_z}
\end{gather}
and the measurements optimum in the extended range
\begin{gather}
    \tilde z^p = \argmin_{z^p \in G_1(\ballpdr{2}{k}{r_1}) \oplus \ballpdr{1}{p}{r_2}}||Ax - AG_2(z^p)||.
    \label{optimal_measurements}
\end{gather}
Then, if the number of measurements is sufficiently large:
\begin{gather}
m = \frac{1}{(1-\gamma)^2}\Omega\left(k \log \frac{L_1L_2r_1}{\delta} + K^2\log p \right),
\end{gather}
then with probability at least 
$1-e^{-\Omega((1-\gamma)^2 \cdot m)}$, we have the following error bound:
\begin{gather}
||x - G_2(\tilde z^p)|| \leq \left( 1 +  \frac{4}{\gamma}\right)||x - G_2(\bar z^p)|| \nonumber \\ +  \delta\cdot \frac{\log(4K)}{\gamma}\cdot \frac{\sqrt{p}}{K} \log\frac{\sqrt p}{K}.
\label{error_bound}
\end{gather}
\label{main-theorem-l1}
\end{theorem}

We will now try to develop intuition about the theorem. We begin by explaining the sets involved in Equations \eqref{optimal_z}, \eqref{optimal_measurements}.
% To do so, we first need to discuss about the notion of the range of $G_1$. During training, latent vectors were sampled according to a distribution $P_Z$.
% We denote with $B_2^k(r_1)$ an $l_2$ ball that contains with exponentially high probability (on $k$) all the vectors that were fed to the generator during the training.
We consider $B_2^k(r_1)$ to be a set containing all the latent codes of the first layer of the generator that could be potentially pre-images of any sensed signal $x$. We refer to $B_2^k(r_1)$ as the domain of $G$ and to $G_1(B_2^k(r_1))$ as the range of $G_1$. The \textit{extended range} contains all vectors that lie within an $l_1$ ball of radius $r_2$ from some point in the range of $G_1$. This is the set $G_1(B_2^k(r_1)) \oplus B_1^p(r_2)$ that both minimizations are performed in.

Let's now consider the error bound of \eqref{error_bound}. First, $\bar z^p$ is the latent code in the extended range that best explains the image $x$. We refer to this as the true optimum latent code. Next, $\tilde z^p$, is the measurements optimum, i.e. the latent code in the extended range that best explains the measurements $Ax$.
It is important to realize that a reconstruction algorithm only has access to this measurement error and can never compute $\bar z^p$. Our goal is to show that $\tilde z^p$ produces an image close to the one produced by $\bar z^p$.

Our theorem states that given enough measurements $m$, the measurements optimum is nearly as good as the true optimum (see \eqref{error_bound} and Remark \ref{scaling}).

% Now, $\bar z^p$ on the Theorem is the latent code in the extended domain of $G_2$ that best explains the image $x$. Since we never observe $x$, we can never solve the problem of Equation \eqref{optimal_z}. Instead, we can find $\tilde z^p$, which is the latent code in the extended domain of $G_2$ that best explains the measurements, $Ax$. The theorem states that once you choose a $\delta$, you are allowed to extend the range of $G_1$ with a ball of radius $\frac{K \delta}{L_2}$ (where $K$ is a free parameter chosen from the set $\{1, ..., \sqrt p\}$). Then, if you have $K^2\log p$ more measurements than the measurements CSGM would require, you are guaranteed that $G_2(\tilde z^p)$ reconstructs $x$ almost as good as $\bar z^p$ (see remarks for scaling of the error term).

% First, similar to CSGM~\cite{bora2017compressed}, for the first we consider that we are optimization over an $l_2$ $k$-dimensional ball of radius $r_1$. 

\begin{remark}[Choice of $K$]
The size of the extended range affects the required number of measurements (\eqref{optimal_measurements}) and our error bound (see \eqref{error_bound}).  Observe that the size of the extended range is directly controlled by $K$, since, for any fixed $\delta$, we set $r_2 = \frac{K\delta}{L_2}$.
As $K$ increases, we explore a bigger set and both terms on the right side of \eqref{error_bound} become smaller. However, measurements scale quadratically with $K$.
We can set $K$ to scale approximately as $\sqrt{k}$ (see Remark \ref{scaling} for details on how all the quantities can scale). For that choice of $K$, observe that our result requires measurements that scale linearly on $k$ (and only logarithmic in $p$) while the CSGM result requires measurements that scale linearly on $p$. The costs for the small increase in the measurements, are 1) the additive error scales with $\sqrt{p}$, 2) we are restricted to exploring a small radius. 

In practice, these can be tuned as hyperparameters and our experiments show that even small expansions significantly outperform CSGM in numerous inverse problems. \end{remark}

\begin{remark}[CSGM sample bound applied directly on the intermediate layer]
We compare to the result we obtain by applying CSGM to the intermediate layer generator. 
That would yield measurements that scale as: 
$$m= \Omega\left( k\log \left( \frac{L_1 L_2 r_1}{\delta}\right) + p \log \left( \frac{L_2 r_2}{\delta} \right)\right).$$ 
These many measurements result in an additive error term of $O(\delta)$. Our new bound requires fewer measurements when the free parameter $K$ is smaller than $\sqrt{p}$. 
\end{remark}

\begin{remark}[Parameter Scaling]
There are various ways to set the parameters in our bounds, depending on the scaling of sizes of the intermediate layers and the Lipschitz constants.
For typical piecewise linear networks with $d$ layers and maximum $n$ neurons in each layer, we know that the end-to-end Lipschitz constant $L \leq L_1 \cdot L_2$ might scale as $n^d$ for bounded maximum weights. Hence, as in CSGM, we may set $r_1$ to scale as $n^d$. The error term $||x - G_2(\bar z^p)||$ scales linearly with $n$. Hence, we need to choose $\delta, K$ such that the additive term in inequality \eqref{error_bound} scales sublinearly. We may set $\delta$ to scale as $\frac{1}{\sqrt p}$. To get the same order of measurements as CSGM, we may set $K$ to scale as $\sqrt{k}$. For that choice of parameters, the radius for the intermediate search, i.e. $r_2$ scales as $\sqrt \frac{k}{p} n^{-d_2}$, where $d_2$ is the depth of $G_2$.
\label{scaling}
\end{remark}

\subsection{Sketch of the proof}
% \noindent \textbf{Proof Ideas:}
The central novelty of our proof is how we upper bound the 
metric entropy of the epsilon nets used to cover the extended range of the generator, i.e. the set $G_1(B_2^k(r_1)) \oplus B_1^p(r_2)$. First, we observe that if $S_1$ is a epsilon net for $G_1(B_2^k(r_1))$ and $S_2$ is an epsilon net for $B_1^p(r_2)$, then a simple bound for the size of an epsilon net on the extended range will have at most $|S_1|\cdot |S_2|$ elements. 

CSGM uses a volumetric argument to upper bound the size of the epsilon net for $S_1$. Our key idea is that using the same method to bound the size of the cover for the $l_1$ ball is sub-optimal for small radii. Instead, we use Maurey's empirical method or the related Sudakov's minoration inequality ~\cite{maurey,wainwright2019high} yielding logarithmic (instead of linear) dependence on the dimension $p$. Maurey's bound poses technical challenges that we need to address when extending the chaining argument of the CSGM proof. With Maurey's method, successive nets in the chaining 
can have significantly higher metric entropy for large radii.  To minimize the additive error in our bound during chaining, we 
switch from volumetric epsilon-nets to Maurey's method at the right selected scale. The full proof of our Theorem can be found in the Appendix.

% \begin{remark}[Multiple splits]
% The theoretical result is presented for a single intermediate optimization problem. However, in practice, we may optimize over multiple intermediate layers. Our result extends trivially in that case.
% \end{remark}

\subsection{S-REC for partial circulant matrices}

We extend the theory of matrices that satisfy the S-REC condition beyond i.i.d. Gaussian measurements. 
To establish that a family of random matrices satisfies this condition (and hence obtains sample complexity bounds), three conditions must be proved with high probability~\cite{bora2017compressed,baraniuk2008simple}: (1) The random matrix $A$ should satisfy the Johnson-Lindenstrauss (JL) lemma on a suitable $\epsilon$-net, (2) The matrix operator norm should be bounded: $\norm{A}_{op} \leq \sqrt{n}$, and (3) for a fixed vector $x$, $\norm{Ax} \leq 2 \norm{x}$.

Here we establish that randomly signed partial circulant matrices satisfy the S-REC condition for a number of measurements scaling similarly to Gaussian i.i.d. measurements. 

\begin{lemma}
Consider the setting of Theorem~\ref{main-theorem-l1}. Let $g = [g_1, \cdots, g_n]$ be a vector with i.i.d. Gaussian entries of variance $1/m$, let $F \in \R^{m\times n}$ be a partical circulant matrix that has $g$ in its first row, and let $D\in \R^{n\times n}$ be a diagonal matrix with uniform $\pm 1$ entries along its diagonal. Then for $m = \Omega\left( \frac{1}{(1-\gamma)^2}(k \log \frac{L_1 L_2 r_1}{\delta } + K^2 \log p)  \log^4(n)\right)$,  $FD$ satisfies S-REC$(G_2(G_1(\ballpdr{2}{k}{r_1}) \oplus \ballpdr{1}{p}{r_2}), 1-\gamma, \delta \cdot \frac{\log(4K)}{\gamma}\cdot\frac{\sqrt{p}}{K} \log\frac{\sqrt{p}}{K})$ with probability $1 - e^{-\Omega(m)}.$
\label{s_rec_circulant}
\end{lemma}

Our proof of this lemma can be found in the Appendix and relies on 
previous results establishing JL properties for partial circulant matrices post-multiplied by random diagonal matrices~\cite{krahmer2011new,hinrichs2011johnson}. 

There is an important computational benefit in such structured measurement matrices. We are sensing high resolution images that are $1024\times1024$ for 3 color channels resulting in signal  dimension $n$ being $3$ million. If measurements are at ten percent (a typically challenging compressed sensing regime), that results to $m\times n$  matrices that are $300k \times 3m$ which require gigabytes to store and hit GPU memory limitations. Therefore random Gaussian measurement matrices cannot be implemented for high resolution imaging. Partial circulant matrices require orders of magnitude less memory due to their structure and matrix-vector products can be computed much faster using FFT. We expect that these benefits will have a key role for future high-resolution imaging systems.

%From the above conditions, JL is the tricky part since circulant matrices do not satisfy JL.
%However, it is known that RIP matrices whose columns have a sign flip satisfy JL\cite{krahmer2011new}. That is, if $A$ is a matrix with $RIP$, and $S$ is a Bernoulli $\pm 1$ diagonal matrix, then $AS$ will have JL.
%\cite{huang2018improved} claim that a partial circulant matrix with $m \gtrsim k \log^2(k) \log(n)$ will satisfy $RIP(k)$. I don't know if this proof is correct because it has 0 citations, but it is well known that $m \gtrsim k \log^2(k) \log^2(n)$ certainly works.\cite{krahmer2014suprema}

\section{Experiments}

\subsection{Algorithmic adaptations to StyleGAN}
Up to this point, we have presented and theoretically analyzed the ILO algorithm. Our method is not tied to any specific architecture and it only assumes access to a generative model and the underlying domain of the latent space of the initial layer. In this section, we present empirical innovations on how to use our framework with the state-of-the-art generative model StyleGAN-2~\cite{stylegan2}.

% How stylegan works
StyleGAN-2 has several peculiarities that need to be taken into account for the design of a compressed sensing algorithm. First, in StyleGAN-2 the initial latent code $z^k \in \R^k$ is not fed directly to the model. Instead, it is first mapped through a multilayer linear network, \textit{the mapping network}, to an intermediate representation $w^k \in \R^k$. We refer to the domains of $z^k, w^k$ as $\mathcal Z, \mathcal W$ respectively. During training, a $z^k$ is sampled according to a distribution on $\mathcal Z$, it gets transformed through the mapping network to a $w^k \in \mathcal W$ and one copy of $w^k$ is fed to each one of the $18$ layers of StyleGAN-2. Additionally, each one of the layers receives a noise vector $u^k$ (unique for each layer). 
%drawn from a different probability space $\mathcal U$.

\subsubsection{Optimization setting}

The first thing to decide is which intermediate layer will be used to split the StyleGAN-2 generator. We observe that we obtain better results with multiple splits. We consider the generator of StyleGAN-2 as a composition of layers $G_1 \circ G_2 \circ ... \circ G_{18}$ and we run Algorithm \eqref{optimization_algorithm} in rounds, where in each round the initial layer is discarded. 

To ensure that we stay in an $l_1$ ball around the manifold at each layer, we use Projected Gradient Descent (PGD)~\cite{nesterov2003introductory}. To implement the projection to an $l_1$ ball around the current best solution (see line 4 of Algorithm \eqref{optimization_algorithm}), we use the method of \citet{duchi2008efficient}. Guided by our theory, we increase the maximum allowed deviation as we move to higher dimensional latent spaces. The radii of the balls are tuned separately as hyperparameters, for a full description see the Appendix.

For all inverse problems, it is helpful to allow the $w^k$ vectors to deviate~\cite{pulse}, i.e. we can optimize over a sequence $\{w_i^k\}_{i=1}^{18}$. The deviations are typically regularized with an additional term in the loss function, which captures the geodesic distance of the vectors. PULSE reports that optimizing only over the first five noise vectors, i.e. $\{u_i^k\}_{i=1}^5$, yields better reconstructions for super-resolution comparing to optimizing over the whole sequence. 
We show that this is not necessarily true if this optimization is performed sequentially. 
Our method starts by optimizing only the first five noise vectors (as in PULSE), but we gradually allow optimization of the rest of the latent vectors as we move to higher dimensional latent spaces.

%Concretely, for every new intermediate layer $G_j:\R^p \to \R^n$, ILO optimizes over: i) a  vector in $\R^p$, initialized to the output of the previous layer, $G_{j-1}: \R^k \to \R^p$, ii) the latent vectors that affect $G_{j}$, i.e. $\{w_i\}_{i=j}^{18}$ and iii) a subset of the noises that can affect $G_{j}$, i.e. a subset of $\{u_i\}_{i=j}^{18}$.

\subsection{Loss functions and adaptation to general inverse problems}

Here we consider the effect of different loss functions in solving  general inverse problems.
It has been observed that LPIPS yields optimal performance with image size $256\times 256$~\cite{stylegan2}. Therefore, we downsample images from $1024\times1024$ to $256\times 256$ pixels. If the given image is inpainted, missing pixels are mixed with observed pixels during this downsampling. We observe that this blending leads to distorted reconstructions when using the LPIPS loss. Hence, for inpainting  under scarce measurements we use only the MSE loss. We note that unlike the previously proposed methods, ILO can work for inpainting with extremely few observed pixels -- even with less than $1\%$ of the whole image.  If we observe a significant portion of the image, then we use both LPIPS and MSE. To address these distortion issues, we minimize the perceptual distance between the generated image and a superimposed reconstruction, i.e. we replace the missing pixels of the observed image with the ones generated by StyleGAN prior to downsampling. 

For super-resolution, we use a weighted average of LPIPS and MSE (as in inpainting with sufficient measurements). To compare the high-resolution and low-resolution images, we first downsample with cubic interpolation~\cite{keys1981cubic} as in PULSE.
We also consider the problem of denoising, where Gaussian noise is added to the image. As usual, we assume knowledge to the forward operator $\mathcal A(x)$. Simply inverting a noisy high-resolution image creates grainy reconstructions due to the expressive power of StyleGAN-2. We address this in the optimization process by adding gaussian noise to the generated images before using them in the loss function.
We call this new technique \textbf{Stochastic Noise Addition} (SNA). 

\subsection{Results}
We show that ILO obtains state-of-the-art unsupervised performance for solving inverse problems with deep generative models in four different settings: inpainting, super-resolution, denoising and compressed sensing with circulant matrices. We compare with different variants of the CSGM algorithm using optimization and loss function innovations introduced in PULSE and StyleGAN. Unless stated otherwise, we will denote with CSGM + MSE the optimization procedure described in PULSE for the StyleGAN generator. Through a wide variety of experiments, we observe that ILO largely outperforms alternative techniques, both in terms of visual
quality and in terms of true MSE error. We measure the latter on images sampled randomly from Celeba-HQ~\cite{celeba, celebahq}. Finally, to show the benefits of extending the range of the generator, we illustrate how one can use an adversarially robust classifier to guide the generation of human faces that look like objects from ImageNet~\cite{imagenet_cvpr09}.

\begin{figure*}[!htp]
\captionsetup[subfigure]{justification=centering}
\captionsetup[subfigure]{labelformat=empty}
\begin{adjustbox}{width=\textwidth, center}
\begin{tabular}{ccc}
\subfloat[]{\includegraphics[width=\textwidth]{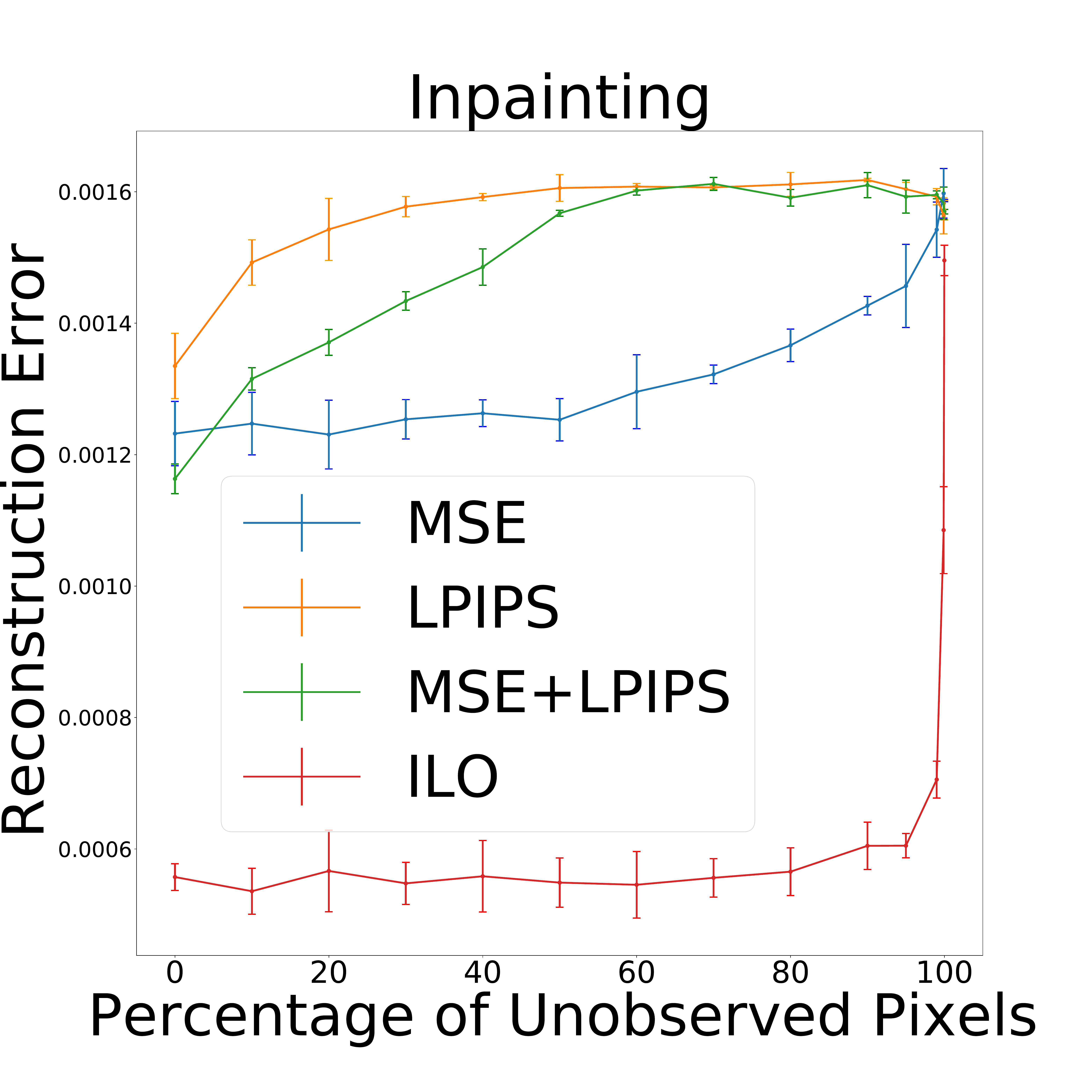}}
\subfloat[]{\includegraphics[width=\textwidth]{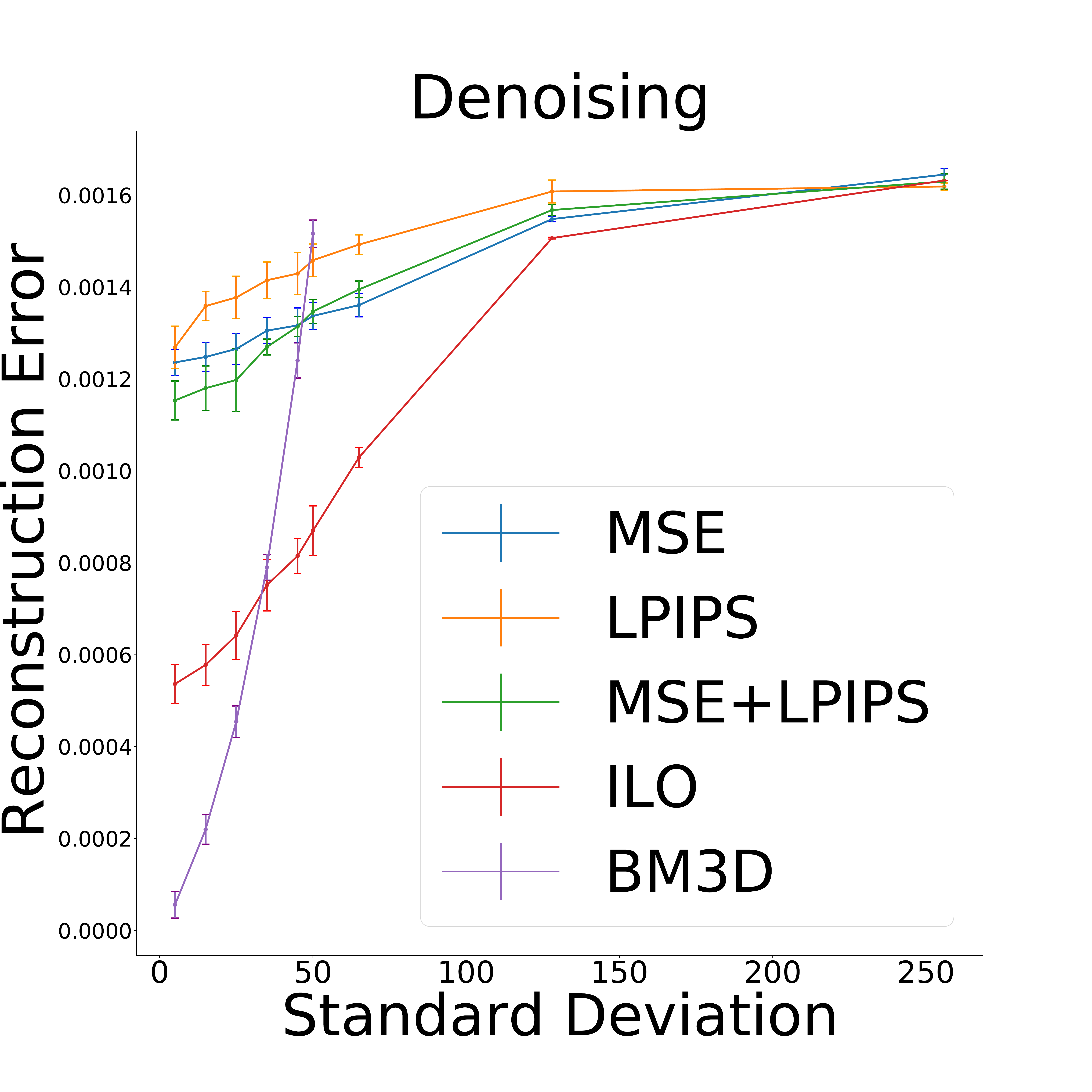}}
\subfloat[]{\includegraphics[width=\textwidth]{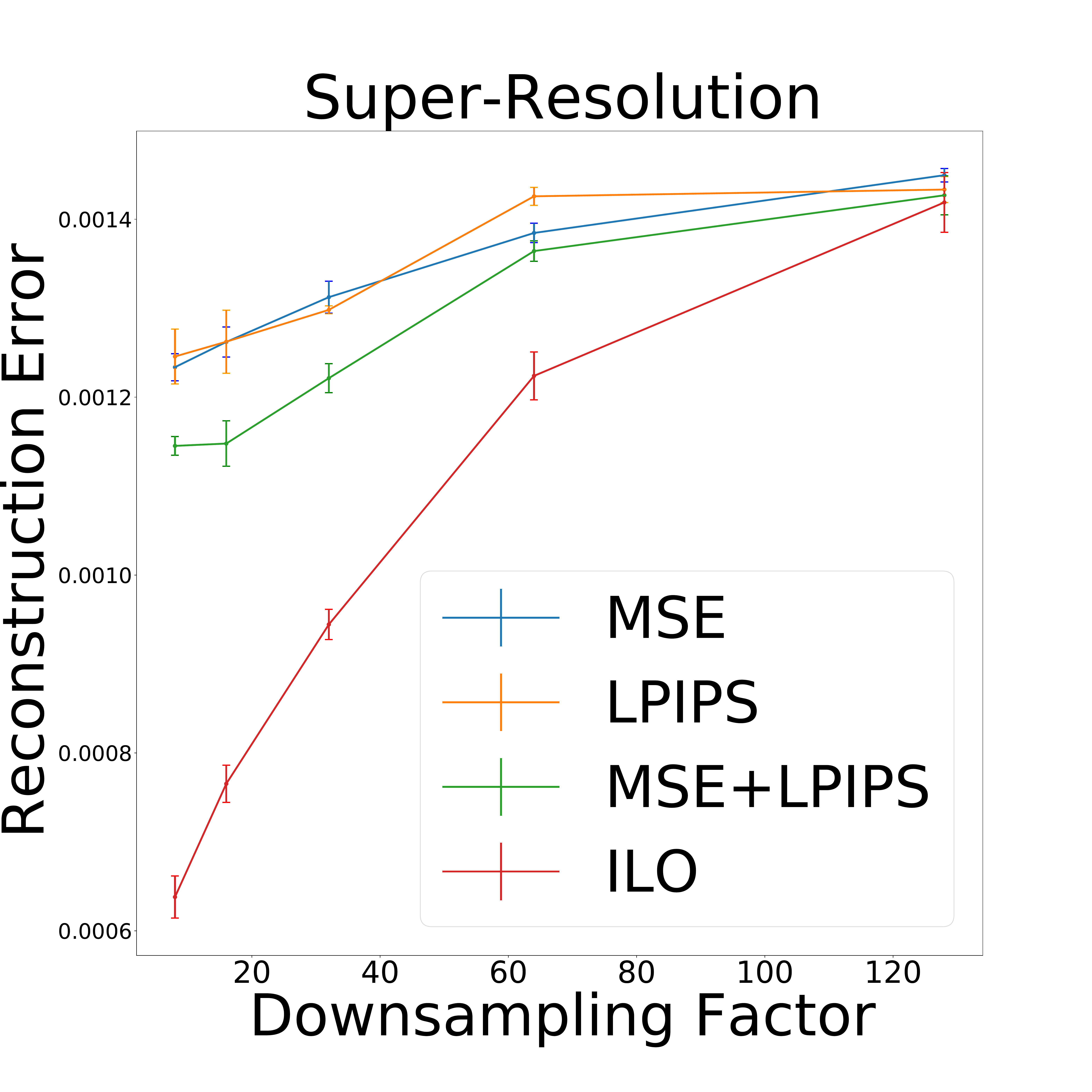}}
\subfloat[][]{\includegraphics[width=\textwidth]{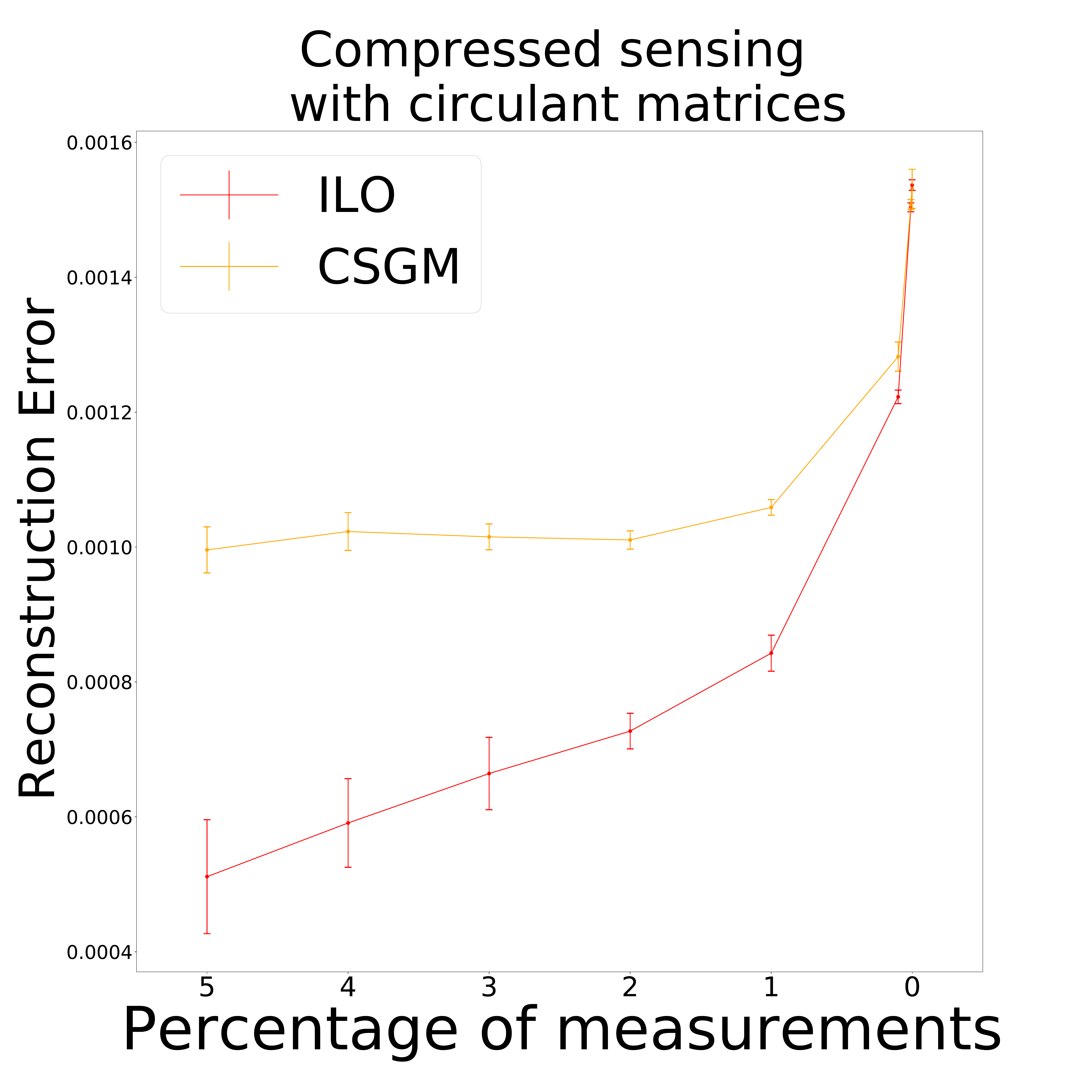}}
\end{tabular}
\end{adjustbox}
\caption{Plots showing the true MSE error on Celeba-HQ images, i.e. the MSE between the real image (that we never observe) and the reconstructed image from the measurements.
From left to right: Inpainting, Denoising, Super-resolution and Compressed sensing with partial circulant matrices. As shown, ILO significantly outperforms all previous methods except in the very noisy regime. }
\label{mse_plots}
\end{figure*}

\noindent \textbf{Inpainting:}
%\subsection{Inpainting}
% Inpainting is the process of reconstructing an image with missing pixel values that are defined by a known mask
For inpainting, the algorithm tries to complete missing pixels to a given image. The measurement process corresponds to a linear matrix
that has rows that are a subset of the identity. 
Results for inpainting are shown in Figure \ref{inpainting_res}.
We perform two types of experiments. First, we mask important facial features from real images (collected from the web) and generated images from StyleGAN-2. Next, we do randomized inpainting, i.e. we inpaint pixels of a given image independently with a pre-defined probability. We experiment with observation probabilities up to $1\%$. This is a very challenging scenario: a human observer cannot distinguish face characteristics from such few pixels, e.g. see Figure \ref{inpainting_res} last row, second column. As shown in the Figure, ILO gives reconstructions that look much closer to the hidden image than the other methods.  Our method is able to give surprisingly accurate reconstructions even under extreme scarce measurements (see last column, last row of Figure \ref{inpainting_res}). To quantify the performance of the different methods we randomly select a few images from Celeba-HQ~\cite{celeba, celebahq} and reconstruct at different levels of sparsity. Figure \ref{mse_plots} column 1 shows that ILO is $2\times$ better in terms of reconstruction error anywhere between $5\% - 100\%$ observed pixels.

% Rows 1, 2, 3 and 5 are real images (outside of the test set, collected from the web) while rows 4, 6 are StyleGAN-2 generated images.
% Column 2: the first five images have masks that were chosen to
% remove important facial features. The last row is an example of randomized inpainting, i.e. a random $1\%$ of the total pixels is observed. Columns 3-5: reconstructions using the CSGM~\cite{bora2017compressed} algorithm with the StyleGAN-2 generator and the optimization setting described in PULSE~\cite{pulse}. While PULSE only applies to super-resolution, we extend it using MSE, LPIPS and jointly MSE+LPIPS loss in columns 3-5. Columns 3-5 form an ablation study of the benefits of each loss function. As shown, the combination of ILO with our new loss consistently give better reconstructions. Focusing in the last two rows, recognizing the person is made possible from very few pixel observations using our method.
% Column 6: reconstructions with ILO (ours). ILO gives reconstructions that look much closer to the hidden image than previously proposed methods.

% Inpainting is the process of reconstructing an image with missing pixel values that are defined by a known mask. By defining the degradation operator as the element-wise multiplication with the known mask, ILO is able to generate realistic images that match the observed pixels. As shown in \ref{inpainting_res}, ILO is able to handle a wide variety of defined masks such as blocked masks or randomized mask without additional training. 

\noindent \textbf{Denoising:}
%\subsection{Denoising}
Our next experiment is on denoising. To ablate the SNA framework we introduced, we show results with and without our technique on an image with additive noise of standard deviation $\sigma=30$. Results are summarized in Table \ref{sna_ablation}.
\begin{table}[!htp]
    \centering
    \begin{tabular}{c|c|c}
    Algorithm & SNA & PSRN (dB) \\ \hline
    \multirow{2}{4em}{CSGM}  &  \xmark & 19.89 \\
    & \checkmark & 21.38 \\ \midrule
    \multirow{2}{4em}{ILO} & \xmark & 28.34 \\
    & \checkmark & \textbf{32.92}
    \end{tabular}
    \caption{Results with and without SNA for a noisy image ($\sigma=30$).}
    \label{sna_ablation}
\end{table}
Since it is clear that SNA consistently improves reconstruction, we use it in all subsequent denoising experiments.

We compare with the CSGM framework using MSE, only LPIPS or a combination of both loss functions. For ILO, we only use a weighted combination of MSE and LPIPS. We also compare with a standard denoising method, the BM3D algorithm~\cite{bm3d}.
We vary the noise standard deviation from $5$ to $256$ and clip the perturbed values to the range $[0, 255]$ (RGB). Results are shown in Figure \ref{mse_plots}, second column. We observe that ILO outperforms all the previously proposed CSGM based methods by a large margin. For the typical setting of 
$\sigma=25$, ILO is $1.8\times$ better than the best performing CSGM baseline. 
BM3D shows excellent performance, outperforming all other methods in the very low noise regime but rapidly deteriorates for harder settings. We refer the reader to the   for visual results and additional denoising experiments.

% Since our baselines perform poorly for denoising, we also compare with compare with BM3D~\cite{bm3d}, a standard denoising method. Visual results are shown in Figure \ref{denoising_results}.

\begin{figure}
\captionsetup[subfigure]{labelformat=empty,font=ltpt}
\captionsetup[subfigure]{justification=centering}
% \captionsetup[subfigure]{labelformat=empty}
\begin{adjustbox}{width=\columnwidth, center}
\begin{tabular}{ccccc}
\captionsetup{justification=centering}
\subfloat{\includegraphics[height=0.4\textwidth, width=0.4\textwidth]{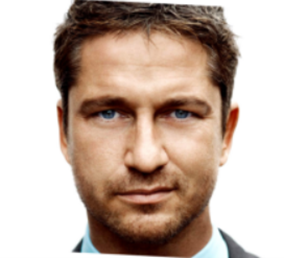}} &
\subfloat{\includegraphics[width=0.4\textwidth]{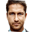}} &
\subfloat{\includegraphics[width=0.4\textwidth]{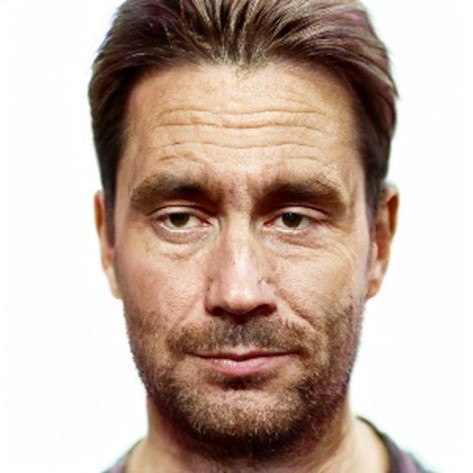}} &
\subfloat{\includegraphics[width=0.4\textwidth]{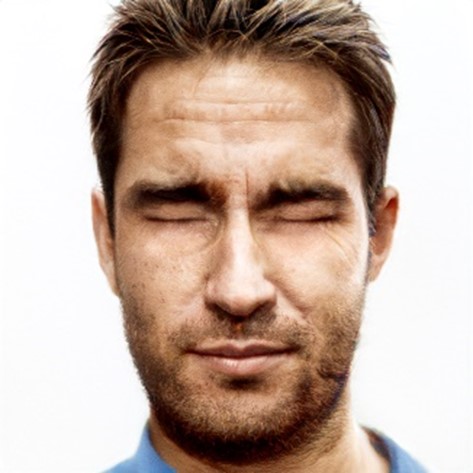}} &
\subfloat{\includegraphics[width=0.4\textwidth]{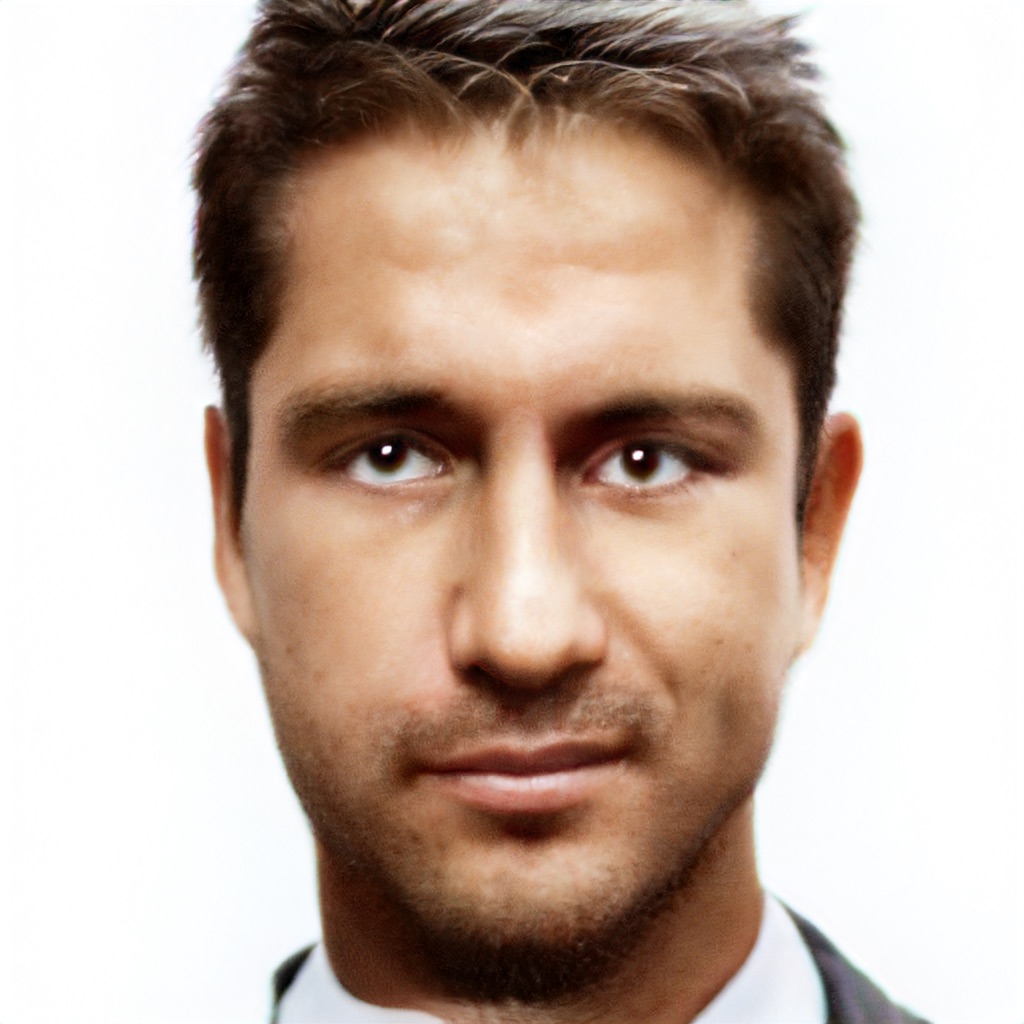}} \vspace{-4mm} \\
\subfloat{\includegraphics[height=0.4\textwidth, width = 0.4\textwidth]{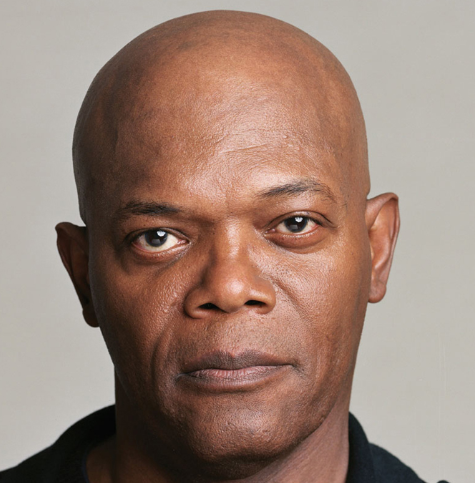}} &
\subfloat{\includegraphics[height=0.4\textwidth, width=0.4\textwidth]{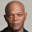}} &
\subfloat{\includegraphics[width = 0.4\textwidth]{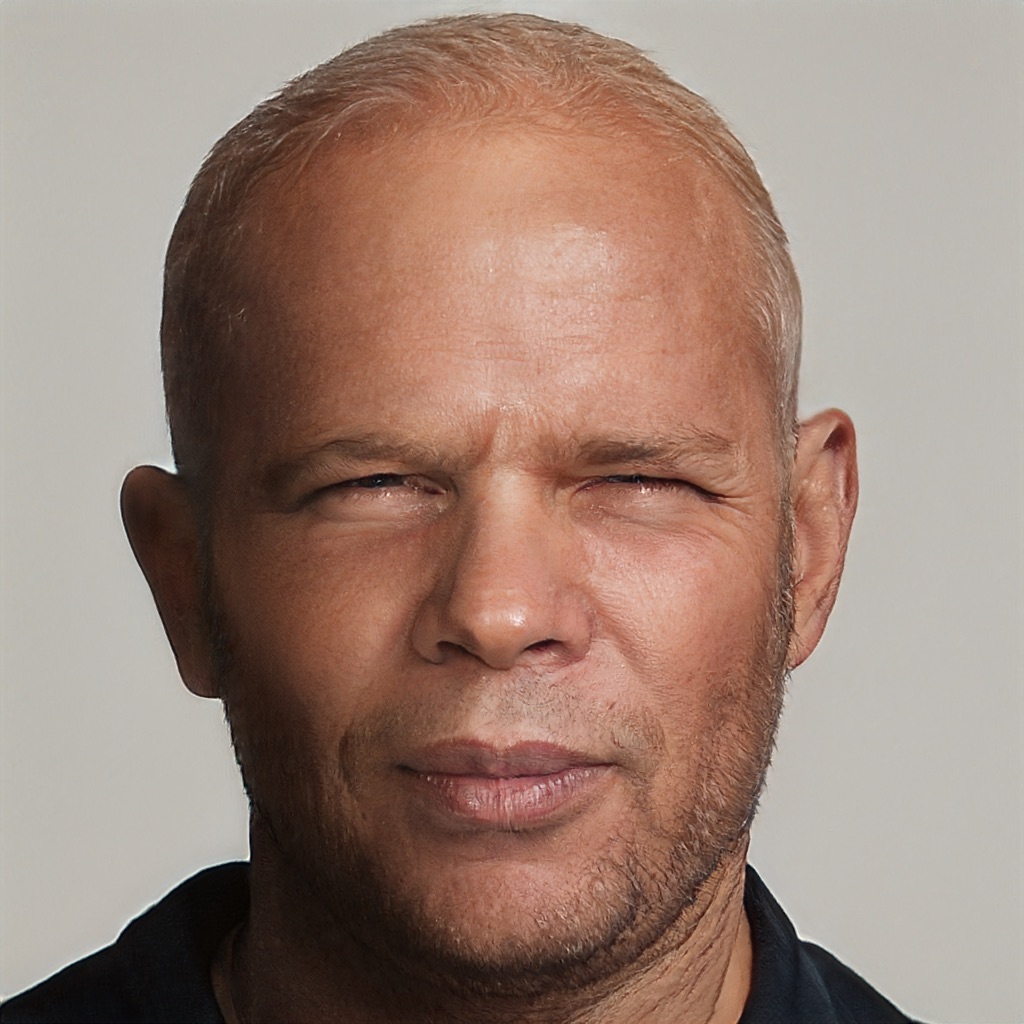}} &
\subfloat{\includegraphics[width = 0.4\textwidth]{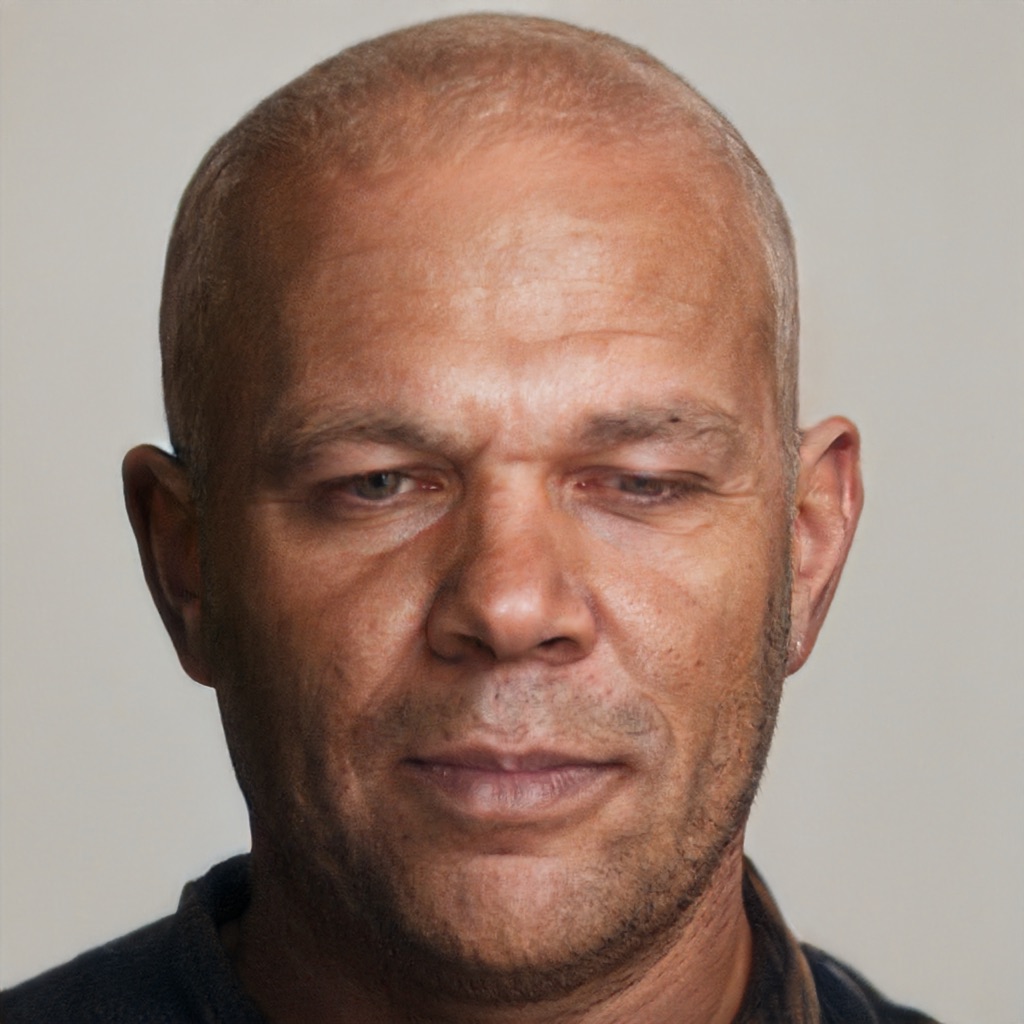}} &
\subfloat{\includegraphics[width = 0.4\textwidth]{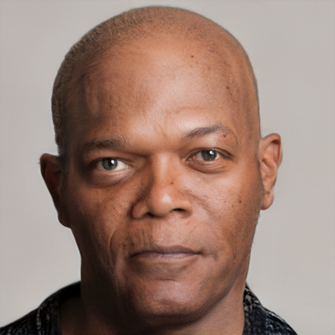}} \vspace{-4mm} \\
\subfloat[Original]{\includegraphics[width = 0.4\textwidth]{ip_images/obama.jpg}} &
% \subfloat[Original]{\includegraphics[width = 0.4\textwidth]{depixel/obama_original.jpg}} & ip_images/obama.png
\subfloat[LR ($\times 16$)]{\includegraphics[width = 0.4\textwidth]{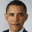}} &
\subfloat[CSGM MSE (PULSE)]{\includegraphics[width = 0.4\textwidth]{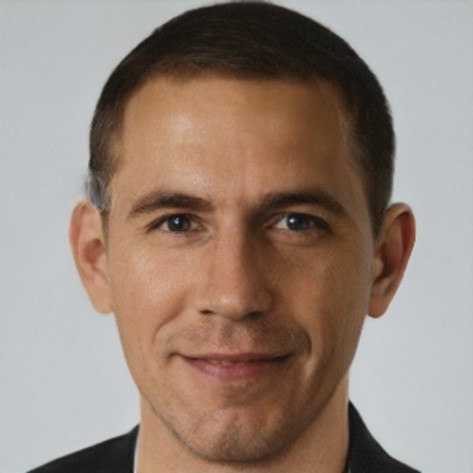}} &
\subfloat[CSGM LPIPS]{\includegraphics[width = 0.4\textwidth]{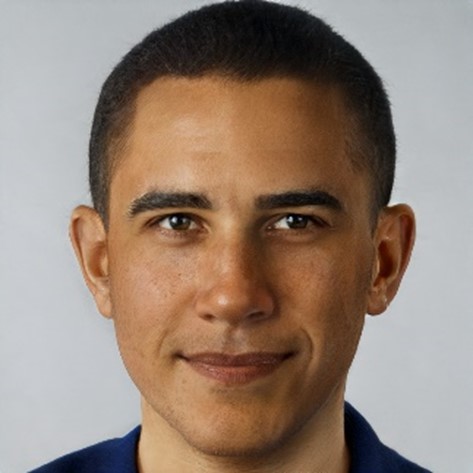}} &
\subfloat[Ours]{\includegraphics[width = 0.4\textwidth]{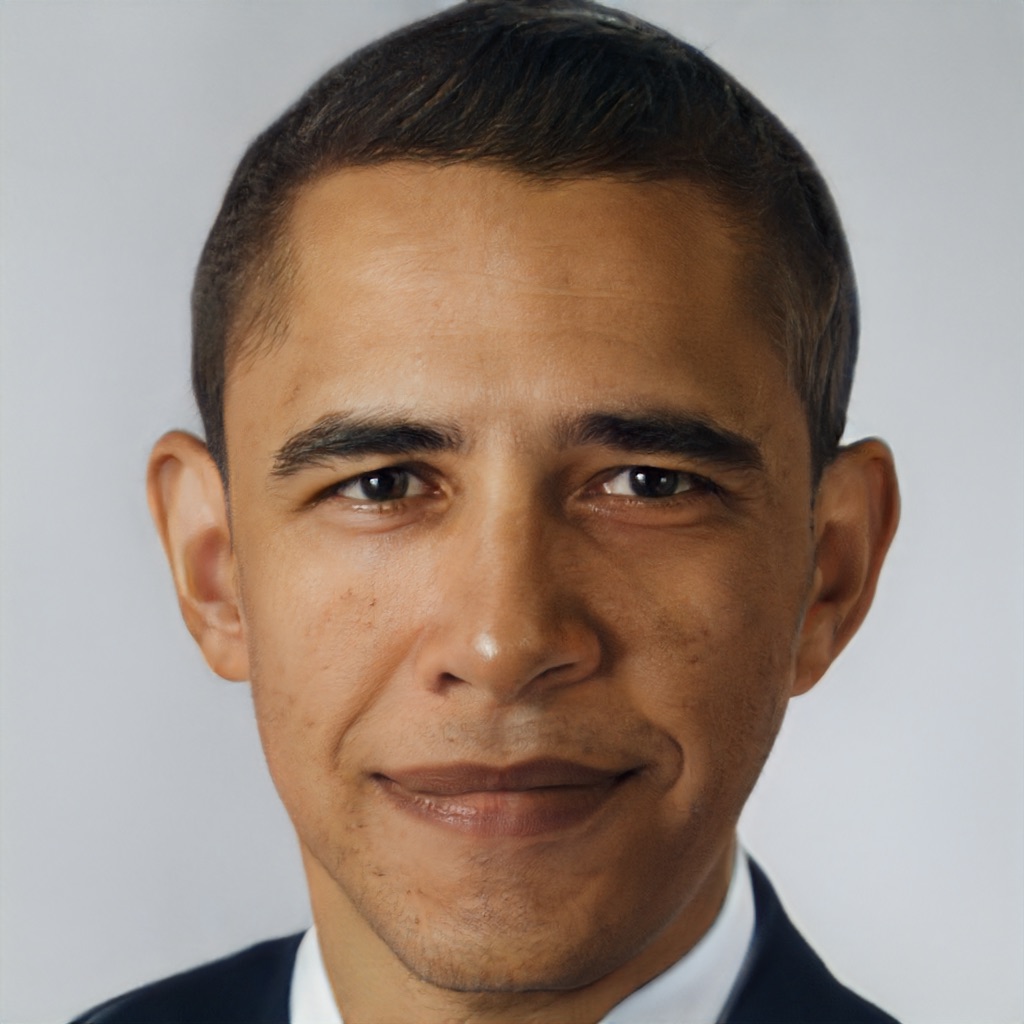}}\\
\end{tabular}
\end{adjustbox}
\caption{\small Results on the super-resolution task.
ILO (ours) gives more accurate reconstructions comparing to PULSE (third column) and other baselines. Many biased reconstructions can be corrected by applying ILO on the weighted combination of MSE and LPIPS.}
\label{super_res}
\end{figure}

\noindent \textbf{Super-resolution:}
%\subsection{Super Resolution}
We report results on super-resolution, the only task PULSE was actually designed for. We sample images from Celeba-HQ, downsample using Bicubic Downsampling (as done in PULSE) and measure the reconstruction error. Three example reconstructions are shown in Figure \ref{super_res}. We also report reconstruction error on Celeba-HQ. Results are reported in Figure \ref{mse_plots}, third column. As shown, ILO outperforms significantly all the other methods, including PULSE (CSGM + MSE). To give some examples, when the image is downscaled from $1024\times 1024$ to $64\times 64$ (scaling factor $16$), ILO is $1.65\times$ better than PULSE in terms of reconstruction error. For $32\times 32$ images, ILO is $1.4\times$ better than PULSE.

As shown, our method not only generates reconstructions that look much closer to the true image, but also appears to generate more racially diverse samples~\cite{jain2020imperfect, tan2020improving, pulse}, e.g. see third row.
% We show that ILO can be applied to super-resolution or depixelization tasks by defining a forward downsampling operator. For these experiments, we use bicubic downsampling and show how ILO can generate high-resolution images for a given low-resolution image. In Figure 2 (right) shows a comparison of various super-resolution algorithms in comparison to ILO. We note that many of the observed biases in ~\cite{pulse} attributed to the StyleGAN-2 model can be corrected by applying ILO to the inversion process.

%----

%---superresolution fig---

\noindent \textbf{Compressed sensing with partial circulant matrices:}
For an experiment with observations of random projections we used  partial circulant measurement matrices with random signs. 
Lemma \ref{s_rec_circulant} establishes that such matrices satisfy the conditions for Theorem \ref{main-theorem-l1}.  Figure \ref{mse_plots}, column 4, shows the reconstruction error when varying the number of measurement rows. When the number of measurements is $5\%$ of the dimension $n$, ILO performs $2\times$ better than CSGM in terms of reconstruction error.

\noindent \textbf{Out of Distribution generation:}
Our method can generate images that lie outside of the range of the pre-trained generator. By choosing the radius of the $l_1$ ball for each layer, we control the trade-off between how natural (comparing to the dataset the model was trained on) these images look, and the out-of-distribution generation capability of our model. 

\begin{figure}
    \captionsetup[subfigure]{justification=centering}
    \captionsetup[subfigure]{labelformat=empty}
    \begin{adjustbox}{width=\columnwidth, center}
        \begin{tabular}{cccc}
            \subfloat[]{\includegraphics[width = 0.7in]{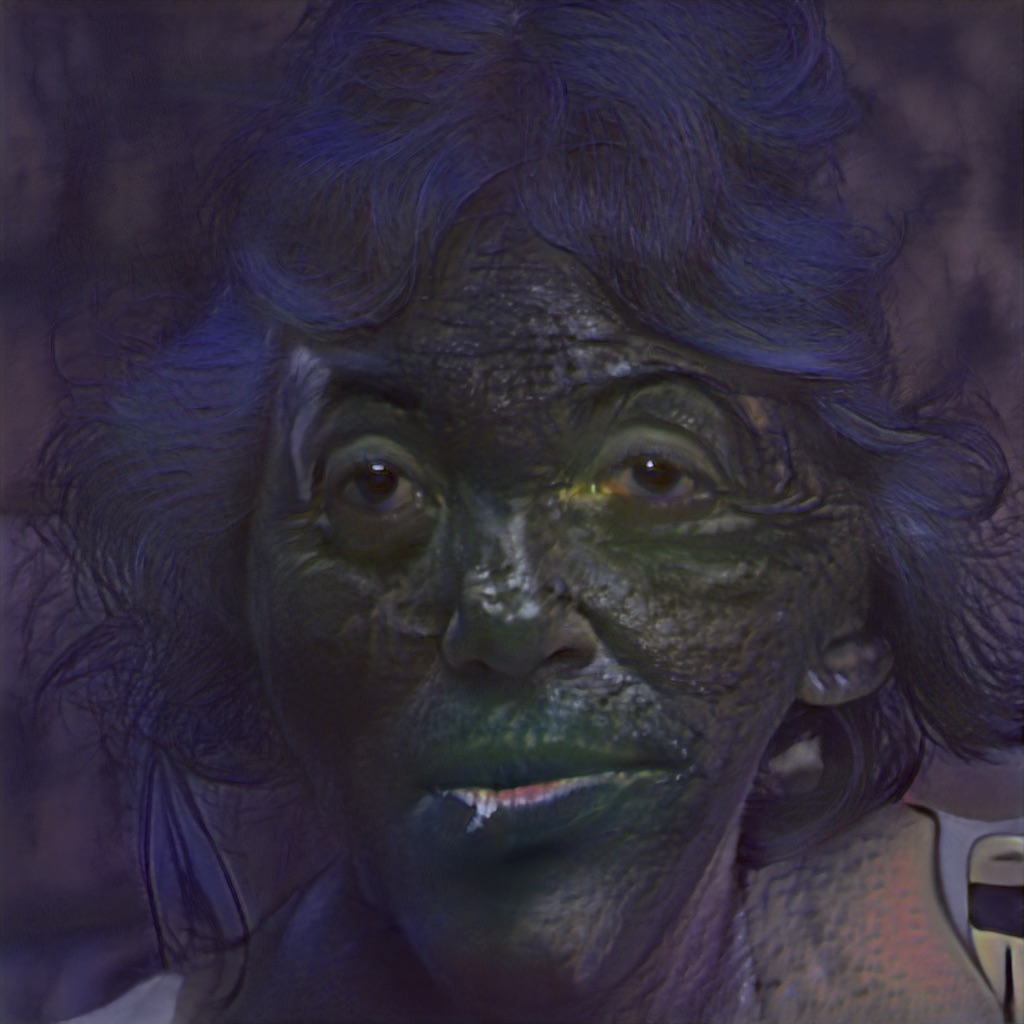}} & \hspace{-4.5mm}  \subfloat[]{\includegraphics[width = 0.7in]{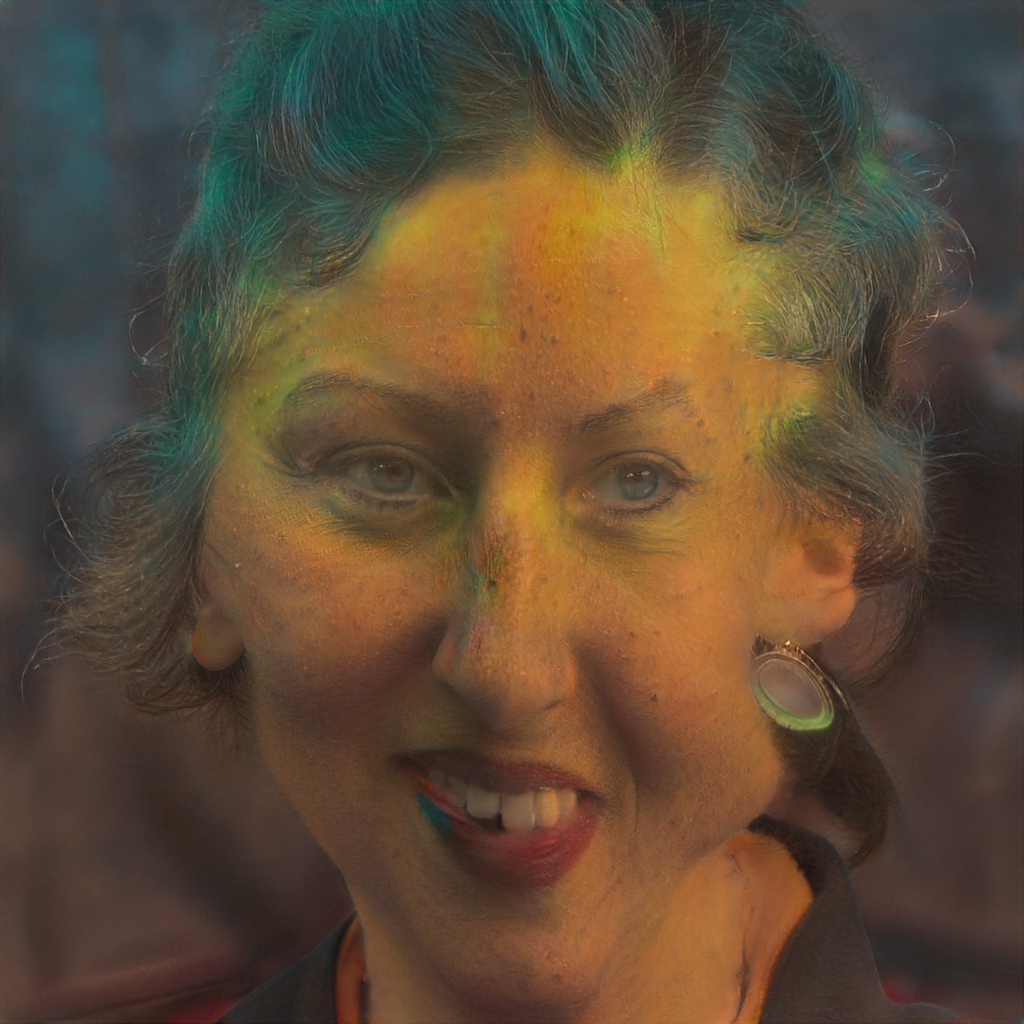}} & \hspace{-4.5mm}  \subfloat[]{\includegraphics[width = 0.7in]{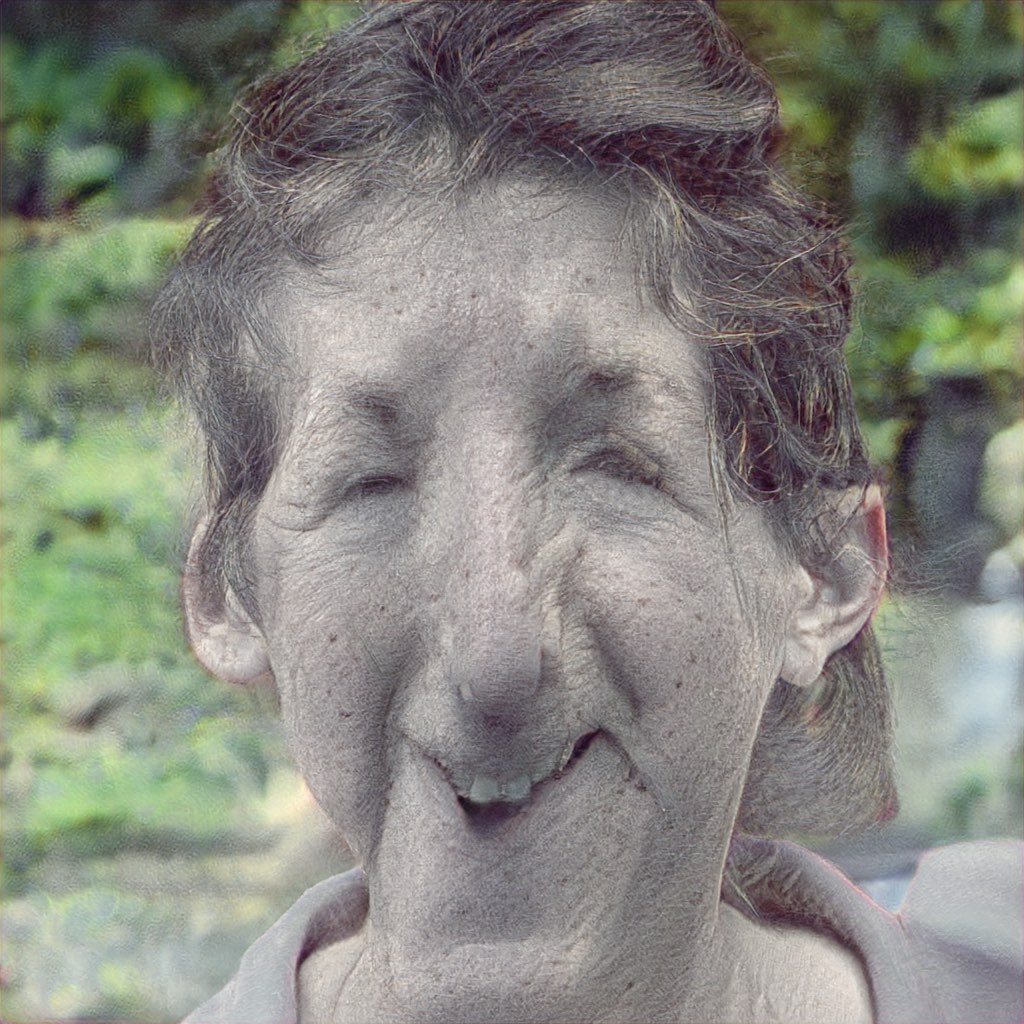}} \vspace{-8mm} \\ 
            \subfloat[]{\includegraphics[width = 0.7in]{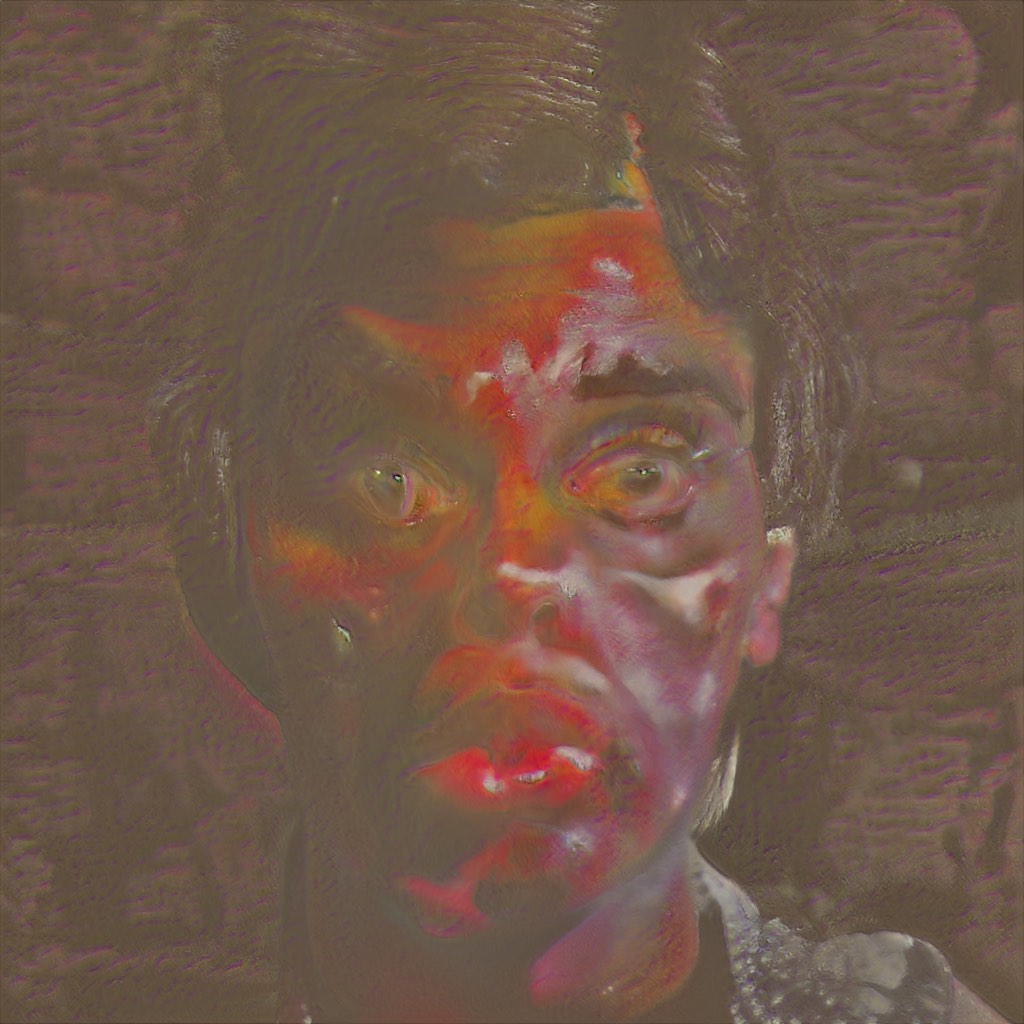}} & \hspace{-4.5mm}  \subfloat[]{\includegraphics[width = 0.7in]{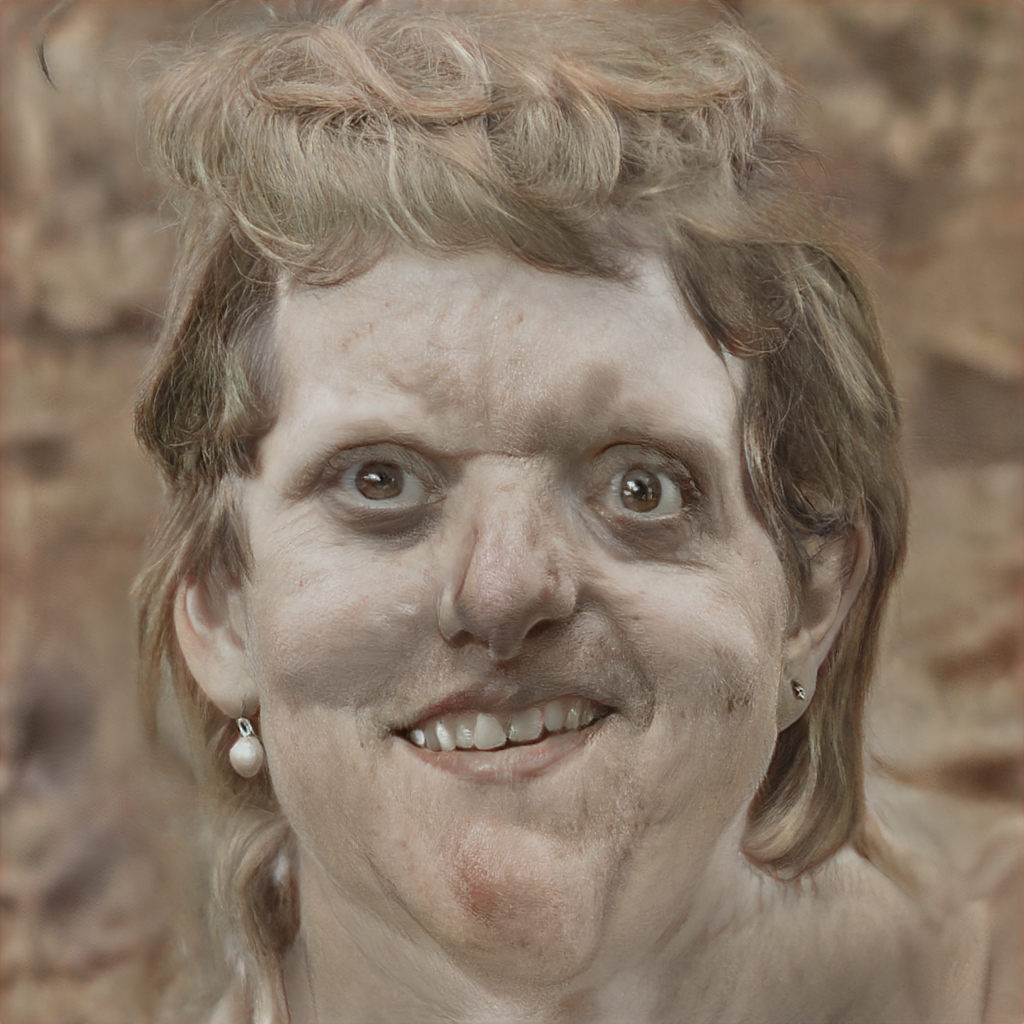}} & \hspace{-4.5mm}  \subfloat[]{\includegraphics[width = 0.7in]{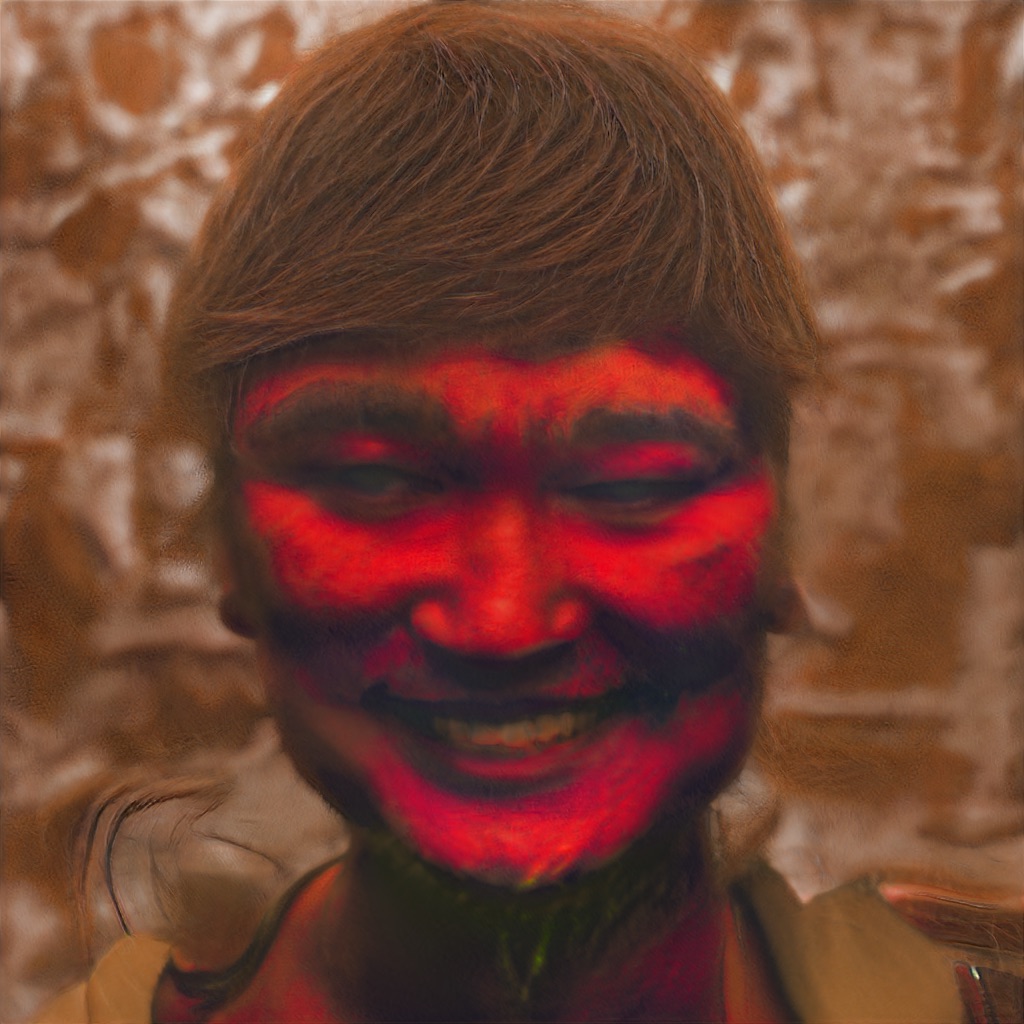}} \\
            % \subfloat[Apple]{\includegraphics[width = 0.7in]{external_classifier/apple3.png}} & \hspace{-4.5mm}  \subfloat[Banana]{\includegraphics[width = 0.7in]{external_classifier/banana3.png}} & \hspace{-4.5mm}  \subfloat[Frog]{\includegraphics[width = 0.7in]{external_classifier/frog3.png}}
        \end{tabular}
    \end{adjustbox}
    \caption{Illustration of using a classifier as a differentiable forward operator. Here we assume that the only observation is $y=\mathcal{A}(x|\textrm{class})$ where $\mathcal{A}$ is an ImageNet classifier. 
    The classes used in this Figure are (from top-left):
    Frog, Coral, Irish Wolf Dog, Goldfish, Boston Terrier Dog and Apple. 
We use a robust classifier as proposed by~\citet{santurkar2019image} and solve the inverse problem to generate images that look like these classes. The difference with~\citet{santurkar2019image} is that we perform the search using ILO in the StyleGAN-2 generator latent spaces as opposed to pixels and that keeps images closer to human faces. 
    }
    \label{fig:external_classifier}
\end{figure}

To demonstrate this, we run the following experiment; we remove entirely the loss functions that relate the generated images with a reference image (i.e. MSE and LPIPS) and we add a new classification loss term using an external classifier trained on a different domain. Essentially, we search for latent codes that lie in an $l_1$ ball around the range of intermediate layers and maximize the probability that the generated image belongs to a certain category. We consider a classifier trained on ImageNet~\cite{imagenet_cvpr09}. This optimization problem is one of the simplest methods to create adversarial examples~\cite{xiao2018generating} and hence the generated images will not be visually interesting. However, if our classifier is adversarially robust, then even optimizing directly over the pixel space leads to an interesting generative process~\cite{santurkar2019image}. We use the latent space of StyleGAN-2 to generate images of faces with fruit or animal characteristics. The radius of the $l_1$ projection at different layers controls the distance of the generated images to human faces. The results are shown in Figure \ref{fig:external_classifier}.

\noindent \textbf{Running time:}
%\subsection{Running time}
Our algorithm runs CSGM as the first step and therefore initially seems to be strictly slower. Surprisingly, ILO can find better solutions than CSGM in \textit{fewer} total steps. StyleGAN-2 typically requires $300-1000$ optimization steps (on the first layer) for a good reconstruction~\cite{stylegan, stylegan2}. However, we observe that running $50$ steps in each one of the first four layers outperforms CSGM. That said, ILO continues to improve with more iterations, also depending on task, number of measurements and hyperparameters. In practice, the obtained inverse problems required approximately $30-60$ seconds per image on a single 1080Ti GPU. For a discussion on hyperparameters, their effects on running times and comparisons to other baselines see the Appendix.
%
%In practice, the obtained inversions were computed in $30-100$ seconds per image on a single 1080Ti GPU while CSGM results were on average $2x$ faster taking $30-50$ seconds. % 

\noindent \textbf{Related Work:} There has been significant recent work in 
unsupervised methods for inverse problems using pre-trained generative models. 
Recently, \citet{inf_theoretic} showed that the sample scaling of CSGM is near-optimal in the absence of further assumptions. \citet{hand2018global} proved algorithmic convergence guarantees for solving non-convex linear inverse problems with deep generative priors under random weight assumptions. Faster recovery algorithms were proposed by \citet{raj2019ganbased, Shah_2018} while \citet{Pandit_2019} analyzed approximate message passing (AMP) for inverse problems in the high-dimensional random limit. Beyond AMP, Regularization-by-Denoising (RED) methods have shown excellent recent performance in imaging, see e.g.~\citet{sun2019block}. 
Deep generative models have been developed for MRI~\cite{mardani2018deep}
and benefited from task-awareness~\cite{kabkab2018taskaware}, meta-learning~\cite{wu2019deep} and 
specifically designed autoencoders~\cite{mousavi2019data}.

The theoretical framework we introduce is related to the ideas proposed by 
\citet{dhar2018modeling} on allowing additive sparse deviations in the generated images. In that case, the recovered signals have the form $G(z) + v$, where $G: \R^k \to \R^n$ is a deep generative model, $z \in \R^k$ is a latent variable and $v \in \R^n$ is an $l$-sparse vector. The additive term allows the recovery of signals that lie outside of the range. 
%
%Under mild assumptions on the measurements matrix,~\citet{dhar2018modeling} prove that if $m = \Omega(l \log \frac{n}{l} + k \log L)$, then there is a decoder that can reconstruct any signal of the form $G(z) + v$ with high probability. This work, albeit very interesting, has certain caveats. First, the proposed objective function is non-differentiable. Second, there is no theoretical guarantee that the relaxation of the problem they solve in practice will recover a solution close to the original problem. 
% Second, unlike the analysis of ~\citet{bora2017compressed}, the reconstruction guarantee of~\citet{dhar2018modeling} is only existential. 
%Third, it is not clear how much is the expressitivity increased by allowing sparse deviations from the range of the generator, especially for models that generate high-resolution images.
%
Our approach is very close to this framework but generalizes it since it allows sparse deviations anywhere in the latent space. 

Another related recent work is that on GAN surgery~\cite{surgery}. In that paper, the range of the generator is expanded by optimizing intermediate layers directly. The main difference is in the optimization procedure; it is not performed sequentially nor regularized by a previous search in the lower dimensional space as we propose in this paper. 
Our paper also benefits from the StyleGAN-2 architecture~\cite{stylegan, stylegan2} and builds on several key ideas from PULSE~\cite{pulse}. 

Finally, there is significant prior work on deep learning methods that do not rely on pre-trained generators, see e.g.~\cite{lucas2018using, Yu_2019, Liu_2019, Sun_2020, Sun2020, yang2019deep, Tian_2020, tripathi2018correction}. Such methods can show excellent performance but require training a network specifically for each reconstruction task. 
This is in contrast with our framework that can solve all inverse problems universally, leveraging the same pre-trained network.

\section{Conclusions and Future Work}

We proposed a novel framework for solving inverse problems leveraging pre-trained generative models.
Our method expands the range of the generator by optimizing different intermediate layers and achieves excellent performance for several tasks. On the theory side, a central open problem would be to establish global convergence of ILO, possibly following the ideas of~\cite{hand2018global,hand2018phase} or surfing~\cite{song2019surfing}.

On the empirical side, a central open problem would be the application of our framework in other  domains like medical imaging, but that would require  pre-trained generative models e.g. for high-resolution MRI images. Another open direction that is particularly exciting is the use of classifiers to generate out-of-distribution samples. 
Our generated samples show the powerful modularity of combining pre-trained generators with 
differentiable forward operators that can guide image reconstruction in a data-driven way. 

\section{Intended Use}
ILO is intended as a proof of concept for solving inverse problems by leveraging pre-trained Generative models. The intended use of our implementation using StyleGAN2 and also the classifier is purely as an art project. Our primary goal is to demonstrate that a classifier can produce images outside the range of a pre-trained generator (i.e. human faces) by leveraging intermediate layer optimization. This model is not suitable, for face recognition or any real subject identification or any real subject image manipulation.  We are not releasing the classifier transformation code in public because of the potential for abuse. Interested artists can contact us for code. 

 The training dataset of the used generator (StyleGAN) has been noted to have imbalance of white faces compared to faces of people of color. Furthermore, different reconstruction optimization methods may be biasing reconstructions, an issue we are investigating in on-going work. Our method can be used with any generative model and perhaps a model trained e.g. with FairFace would be better but this is part of our on-going research.

\clearpage

\section{Appendix}

\printunsrtglossary[type=symbols]

\subsection{Proofs}
\subsubsection{Metric entropy for the $l_1$ ball}

\begin{definition}[Covering number~\cite{wainwright2019high}]
A $\delta$-covering of a set $\mathbb T$ with respect to a metric $\rho$ is a set $\{\theta^1, ..., \theta^M \}\subset \mathbb T$ such that for each $\theta \in \mathbb T$, there exists some $i \in [N]$ such that $\rho(\theta, \theta^i) \leq \delta$. The $\delta$-covering number $N(\delta, \mathbb T, \rho)$ is the cardinality of the smallest $\delta$-cover.
\end{definition}

\begin{definition}[Packing number~\cite{wainwright2019high}]
A $\delta$-packing of a set $\mathbb T$ with respect to a metric $\rho$ is a set $\{\theta^1, ..., \theta^M \}\subset \mathbb T$ such that $\rho(\theta^i, \theta^j) > \delta$ for all distinct $i, j \in [M]$. The $\delta$-packing number $M(\delta, \mathbb T, \rho)$ is the cardinality of the largest $\delta$-packing.
\end{definition}

\begin{lemma}[\citet{wainwright2019high}]
For all $\delta > 0$, the packing and covering numbers are related as follows:
\begin{gather}
    M(2\delta, \mathbb T, \rho) \leq N(\delta, T, \rho) \leq M(\delta, T, \rho).
    \label{packing_covering}
\end{gather}
\end{lemma}

\begin{theorem}[Maurey's Empirical Method~\cite{maurey}]
Let $\ballpdr{1}{d}{r} = \{ x \in \R^d \ | \ ||x||_1\leq r\}$. Then, 
\begin{gather}
    \log N(\delta, \ballpdr{1}{d}{r}, ||\cdot ||_2) \leq \frac{r^2}{\delta ^2}\log(2d+1).
\end{gather}
\label{maureys}
\end{theorem}
A short proof of this result follows.
\begin{proof}
Fix $x \in \R^d$. 
Let $Z$ be the following RV:
\begin{gather}
    Z = \begin{cases}
    \sgn(x_i)r e_i, \quad \wp \ \frac{|x_i|}{r} \\
    0, \quad \wp 1 - \frac{||x||_1}{r}
    \end{cases}
\end{gather}

Observe that: $E[Z_i] = \sgn(x_i) r \cdot  \frac{|x_i|}{r} = x_i$ and $V[Z_i] = r^2 \cdot \frac{|x_i|}{r} = r|x_i|$.

Let \begin{gather}
    \bar Z = \frac{1}{t} \sum_{i=1}^{t}Z_i
\end{gather}
where $Z_i$ are independent copies of $Z$.

We have that:
\begin{gather}
    E[||\bar Z - x||^2] = E\left[ \sum_{j=1}^d (\bar Z_j - x_j)^2\right] \\ = \sum_{j=1}^d E\left[ (\bar Z_j - x_j)^2 \right] = \sum_{j=1}^{d} V(\bar Z_j) \\
    = \sum_{j=1}^{d}V\left( \frac{1}{t}\sum_{i=1}^{t} (Z_i)_j\right) = \\ \frac{1}{t^2}t\sum_{j=1}^{d}V(Z_j) = \frac{1}{t}\sum_{j=1}^{d} r|x_i| =\frac{r||x||_1}{t} \leq \frac{r^2}{t}.
\end{gather}
If we choose $t$ such that: $\frac{r^2}{t}\leq \delta^2$, then, we have that $E[||\bar Z - x||^2] \leq \delta^2$. Hence, for $t \geq \frac{r^2}{\delta^2}$, by the Pigeonhole Principle, we have that there is a $\bar Z$ such that: $||\bar Z - x|| \leq \delta$. In other words, the set of all possible $\bar Z$ form an $\delta-$net for $\ballpdr{1}{d}{r}$ for $t \geq \frac{r^2}{\delta^2}$. Set $t = \frac{r^2}{\delta^2}$. We will now count how many $\bar Z$ there are. For each $\bar Z$, we have $t$ choices, each one of which can take one value among $2d + 1$ values. Hence, there are $(2d+1)^t$ different $\bar Z$. Therefore, we can create an $\delta$-net of $\ballpdr{1}{d}{r}$ that has $(2d+1)^{\frac{r^2}{\delta^2}}$ elements, i.e.
$$
\log N(\delta, \ballpdr{1}{d}{r}, ||\cdot||_2) \leq \frac{r^2}{\delta^2}\log(2d+1).
$$
\end{proof}

The same result (up to constants) for the size of the $\epsilon$-net for an $l_1$ ball follows from Sudakov's minoration inequality.

\begin{theorem}[Sudakov minoration~\cite{sudakov1969gaussian, wainwright2019high}]
Let $\{X_\theta, \theta \in T\}$ be a zero-mean Gaussian process on $T\subset R^d$. Then,
\begin{gather}
    \mathcal G(T) \geq \frac{\delta}{2} \sqrt{\log M(\delta/2, T, \rho_X)},
    \label{sudakov}
\end{gather}
with $\rho_X(\theta_1, \theta_2) = \sqrt{E[(X_{\theta_1} - X_{\theta_2})^2]}$.
\end{theorem}

\begin{corollary}
\begin{gather}
    \log N(\delta, \ballpdr{1}{d}{r}, ||\cdot ||_2)\leq \frac{16r^2}{\delta^2}\log d.
\end{gather}
\end{corollary}

\begin{proof}
Observe that:
\begin{gather}
    \mathcal G(B_1^d(r)) = E_w\left[ \sup_{||u||_1 \leq r}u^Tw\right] \\
    \leq r E_w\left[ ||w||_{\infty} \right] \\
    \leq 2r\sqrt{\log d}.
\end{gather}

By Inequality \eqref{sudakov},
\begin{gather}
    \mathcal G(T) \geq \frac{\delta}{2}\sqrt{\log M(\delta/2, T, \rho_X)} \Rightarrow \\
    \log M(\delta/2, B_1^d(r), ||\cdot ||_2) \leq 16 \frac{r_1^2}{\delta^2}\log d.
\end{gather}
It follows from the definition of the covering number that:
$$
M(\delta, T, ||\cdot ||_2) \leq M(\delta/2, T, ||\cdot||_2).
$$
By Inequality \eqref{packing_covering}, we also have:
\begin{gather}
    N(\delta, T, ||\cdot||_2) \leq M(\delta, T, ||\cdot ||_2).
\end{gather}
Hence,
\begin{gather}
    \log N(\delta, \ballpdr{1}{d}{r}, ||\cdot ||_2)\leq \frac{16r^2}{\delta^2}\log d.
\end{gather}
\end{proof}

\begin{theorem}[Volume rations and metric entropy~\cite{wainwright2019high}]
Let $\Theta$ be an arbitrary set. Then,
\begin{gather}
    \frac{\vol(\Theta)}{\vol(\delta \ballpdr{q}{d}{1})} \leq N(\delta, \Theta, ||\cdot||_q) \leq \frac{\vol\left(\frac{2}{\delta}\Theta \oplus \ballpdr{q}{d}{1}\right)}{\vol(\ballpdr{q}{d}{1})}
\end{gather}
\label{volumetric_argument}
\end{theorem}

\begin{corollary}
\begin{gather}
    \log N(\delta, \ballpdr{1}{d}{r}, ||\cdot||_2) \leq d\log\frac{4r}{\delta}
\end{gather}
\label{vol}
\end{corollary}

\begin{proof}
By Theorem \ref{volumetric_argument}, we have that:
\begin{gather}
    N(\delta, B_1^d(r), ||\cdot ||_2) \leq \frac{\vol\left( \frac{2}{\delta} B_1^d(r) \oplus B_2^d(1)\right)}{\vol(B_2^d(1))} \\
    \frac{\vol\left( \frac{2}{\delta} B_2^d(r) \oplus B_2^d(1)\right)}{\vol(B_2^d(1))}
    \leq \left(\frac{2r}{\delta} + 1\right)^d \\
    \leq \left( \frac{4r}{\delta}\right)^d.
\end{gather}
\end{proof}

\begin{remark}
Observe that by Theorem \ref{maureys} and Corollary \ref{vol}, we get two different upper bounds regarding the covering of the $l_1$-ball. With Maurey's method, the covering number depends logarithmically in the dimension but polynomially on $\frac{1}{\epsilon}$. On the other hand, the volumetric argument gives polynomial dependence on the dimension and logarithmic dependence on $\frac{1}{\epsilon}$. The Maurey's bound is tighter when $\epsilon = \Omega\left( \frac{r}{\sqrt{d}}\right)$.
\end{remark}

\subsubsection{S-REC}
\begin{lemma}[S-REC for nested $l_1$-ball]
Let $G=G_2\circ G_1$ with $G_1:\R^k \to \R^p$ be an $L_1$-Lipschitz function and $G_2:\R^p \to \R^n$ be an $L_2$-Lipschitz function. Let $A \in R^{m\times n}$ be a random matrix with $A_{ij} \sim \N(0, 1/m)$ i.i.d. entries.

Then, if 
\begin{gather}
    m = \frac{1}{(1-\gamma)^2}\Omega\left(k \log \frac{L_1L_2r_1}{\delta} + K^2 \log p \right) \\
    \quad r_2 = \frac{K \cdot \delta}{L_2}, \quad 1 < K < \sqrt{p}
\end{gather}
w.p. $1-e^{-\Omega((1-\gamma)^2m)}$, we have that $A$ satisfies S-REC$(G_2(G_1(\ballpdr{2}{k}{r_1}) \oplus \ballpdr{1}{p}{r_2}), \gamma, \log(4K) \cdot \frac{\sqrt p}{K} \cdot \log\frac{\sqrt p}{K})$.
\label{s_rec_for_nested_l1}
\end{lemma}

\begin{proof}
Using Theorem \ref{volumetric_argument}, we get that:
\begin{gather}
    N\left(\frac{\delta}{L_1 \cdot L_2}, \ballpdr{2}{k}{r_1}, ||\cdot||_2\right) \leq \nonumber \\ \left(\frac{2L_1L_2r_1}{\delta} + 1\right)^k \leq \left( \frac{4L_1L_2r_1}{\delta}\right)^k
\end{gather}
Using the fact that $G_2\circ G_1$ is $L_1L_2$ Lipschitz, we get that:
\begin{gather}
    N(\delta, G(\ballpdr{2}{k}{r_1}), ||\cdot||_2) \leq 
    \left(\frac{4L_1L_2r_1}{\delta}\right)^k
    \label{covering1}
\end{gather}

Using Maurey's Empirical Method (see Theorem \ref{maureys}), we get that:
\begin{gather}
    \log N\left(\frac{\delta}{L_2}, \ballpdr{1}{p}{r_2}, ||\cdot||_2\right) \leq \frac{r_2^2 L_2^2}{\delta^2}\log(2p+1).
\end{gather}
Setting $r_2 = \frac{K \cdot \delta}{L_2}$ and using the fact that $G_2$ is $L_2$-Lipschitz, we get:
\begin{gather}
    \log N\left(\delta, G_2( \ballpdr{1}{p}{r_2}), ||\cdot||_2\right) \leq K^2 \log 3p
    \label{covering2}
\end{gather}

By \eqref{covering1}, \eqref{covering2}, we get that:
\begin{gather}
    \log N\left(\delta, G_2(G_1(\ballpdr{2}{k}{r_1}) \oplus \ballpdr{1}{p}{r_2}), ||\cdot||_2\right)  \leq \nonumber \\ 
    k \log \frac{4L_1L_2r_1}{\delta} + K^2 \log 3p.
\end{gather}

By JL lemma, if $m = \frac{1}{a^2} \Omega\left( k \log \frac{4L_1L_2r_1}{\delta} + K^2 \log 3p \right)$, then w.p. $1 - e^{-\Omega(a^2m)}$, we have that:
\begin{gather}
    ||AG_2(\hat z_2^p) - AG_2(\hat z_1^p)|| \geq \nonumber \\ (1-a)||G_2(\hat z_2^p) - G_2(\hat z_1^p)||, \ \forall \hat z_1^p, \hat z_2^p \in S
\end{gather}
where $S$ is a minimal $\delta$-net of $G_2(G_1(\ballpdr{2}{k}{r_1})  \oplus \ballpdr{1}{p}{r_2})$.

Let $z_1^p, z_2^p \in G_1(\ballpdr{2}{k}{r_1})) \oplus  \ballpdr{1}{p}{r_2} $
and $\hat z_1^p = \argmin_{\tilde z_1^p \in S}||z_1^p - \tilde z_1^p||$, $\hat z_2^p = \argmin_{\tilde z_2^p \in S}||z_2^p - \tilde z_2^p||$.
Then, 
\small
\begin{gather}
    ||AG_2(z_2^p) - AG_2(z_1^p)||\geq ||AG_2(\hat z_2^p) - AG_2(\hat z_1^p)|| \nonumber \\  - ||AG_2(z_2^p) - AG_2(\hat z_2^p)||  -||AG_2(z_1^p) - AG_2(\hat z_1^p)|| \\
    \geq (1-a)||G_2(\hat z_2^p) - G_2(\hat z_1^p)|| - ||AG_2(z_2^p) - AG_2(\hat z_2^p)||  \nonumber \\ -||AG_2(z_1^p) - AG_2(\hat z_1^p)|| \\
    \geq (1-a)||G_2(z_2^p) - G_2(z_1^p)|| - \nonumber \\ (1-a)\left( ||G_2(z_2^p)  - G_2(\hat z_2^p)|| + ||G_2(z_1^p) - G_2(\hat z_1^p)||\right) \nonumber \\  - ||AG_2(z_2^p) - AG_2(\hat z_2^p)||  -||AG_2(z_1^p) - AG_2(\hat z_1^p)|| \\
    \geq (1-a)||G_2(z_2^p) - G_2(z_1^p)|| -  2\delta \nonumber \\ - ||AG_2(z_2^p) - AG_2(\hat z_2^p)||  -||AG_2(z_1^p) - AG_2(\hat z_1^p)||
\end{gather}
\normalsize
By Lemma \ref{measurements_dist_in_the_net}, we have that w.p. $1-e^{-\Omega( m)}$, $||AG_2(z_2^p) - AG_2(\hat z_2^p)|| + ||AG_2(z_1^p) - AG_2(\hat z_1^p)||  = O\left(\log(4K) \cdot \frac{\sqrt p}{K} \cdot \log\frac{\sqrt p}{K} \right) \cdot \delta$.  Let $a = 1-\gamma$.
Hence,
\begin{gather}
    ||AG_2(z_2^p) - AG_2(z_1^p)||\geq \gamma ||G_2(z_2^p) - G_2(z_1^p)||  \nonumber \\ 
    - \log(4K) \cdot \frac{\sqrt p}{K} \cdot \log\frac{\sqrt p}{K} \cdot \delta.
\end{gather}
\end{proof}

\begin{lemma}[]
Let $G=G_2\circ G_1$ with $G_1:\R^k \to \R^p$ be an $L_1$-Lipschitz function and $G_2:\R^p \to \R^n$ be an $L_2$-Lipschitz function.
Let $A \in R^{m\times n}$ be a random matrix with $A_{ij} \sim \N(0, 1/m)$ i.i.d. entries. Let $M_0$ be a $\frac{\delta}{L_2}$ net of $G_1(B_2^k(r_1)) \oplus B_1^p(r_2)$ such that $\log|M_0| \leq k\log\left(\frac{4L_1L_2r_1}{\delta}\right) + K^2\log 3p$.

Then, if 
\begin{gather}
    m = \Omega\left( k\log\left(\frac{4L_1L_2r_1}{\delta}\right) + K^2\log p \right), \nonumber \\ \quad r_2 = \frac{K\cdot \delta}{L_2}, \quad 1 < K < \sqrt{p}.
\end{gather}
then for any $x \in G_2(G_1(B_2^k(r_1)) \oplus B_1^p(r_2))$, if $x' = \argmin_{\hat x \in G_2(M_0)}||x - \hat x||$, w.p. $1-e^{\Omega(m)}$, we have that:
\begin{gather}
    ||A(x - x')|| = O\left( \log(4K) \cdot \frac{\sqrt p}{K} \cdot \log\frac{\sqrt p}{K} \right) \cdot \delta.
\end{gather}
\label{measurements_dist_in_the_net}
\end{lemma}

\begin{proof}
From Lemma 8.2 of \cite{bora2017compressed}, we have that if $\epsilon \geq 2 + \frac{4}{m}\log\frac{2}{f}$, then
\begin{gather}
    P(||Ax|| \geq (1+\epsilon)||x||) \leq f.
\end{gather}

Let $N_0 \subseteq N_1 \subseteq ... \subseteq N_l$ be a chain of minimal $\delta_i$-nets of $G_2(G_1(B_2^k(r_1)) \oplus B_1^p(r_2))$.

Let also:
\begin{gather}
    T_i = \{ x_{i+1} - x_{i} | x_{i+1}\in N_{i+1}, x_{i} \in N_i\}.
\end{gather}

By union bound,
\begin{gather}
    P(||At|| \leq (1+\epsilon_i)||t||, \quad \forall i \in [0, ..., l-1], \quad \forall t \in T_i) \geq \nonumber \\ 
    1 - \sum_{i=0}^{l-1}|T_i|f_i,
\end{gather}
where $\epsilon_i = 2 + \frac{4}{m}\log \frac{2}{f_i}$.
We want to choose $f_i$ such that $\sum_{i=0}^{l-1}|T_i|f_i$ decays exponentially with $m$.

First notice that:
\begin{gather}
    \log|T_i| \leq \log|N_{i+1}| + \log|N_i|
\end{gather}
To develop bounds for $\log|N_i|, \log|N_{i+1}|$ we first need to decide how $\delta_i$ decays and then whether we are going to use Maurey's method or the volumetric argument. 
% Intuitively, at first, it is better to use Maurey's method because we are dealing with nets such that $\frac{\delta_i}{r_2}$ is relatively big. However, as $i$ increases, it is better to use the vol. argument. 

We choose $\delta_i = \frac{\delta}{2^i}$. Now assume $m=K^2\log(3p) + k\log\left( \frac{L_1L_2r_1}{\delta}\right)$.

For $0\leq i < \log\frac{\sqrt{p}}{K}$ we will use Maurey's method. 
\begin{gather}
    \log |T_i| \leq 2\log|N_{i+1}| \\
    \leq 2 \left( \left(\frac{K \delta}{\delta_{i+1}}\right)^2\log(3p) + k\log\left( \frac{L_1L_2r_1}{\delta_i}\right)\right) \\
    \leq 2\cdot \left(4^{i+1}K^2\log(3p) + k\log\left( \frac{L_1L_2r_1}{\delta}
    \right) + k\left( i + 1\right)\right) \\
    \leq  2\cdot \left(4^{i+1}K^2\log(3p) + 2k\log\left( \frac{L_1L_2r_1}{\delta}
    \right)(i+1)\right) \\
    \leq 2 \cdot 4^{i + 1}m.
    \label{maurey_size}
\end{gather}

To get probability that decays exponentially with $m$, we choose:
\begin{gather}
    \log f_i = -3\cdot 4^{i+1}m \\ 
    \epsilon_i = O(1) + 3\cdot 4^{i+1}.
\end{gather}

For $\log\frac{\sqrt{p}}{K} \leq i \leq l -1$, we will use the volumetric argument.
\begin{gather}
    \log |T_i| \leq p\log\left(4K\frac{\delta}{\delta_i}\right) + p\log\left(4K\frac{\delta}{\delta_{i+1}}\right) + \nonumber \\ k\log\left( \frac{L_1L_2r_1}{\delta_i}\right) + k\log\left( \frac{L_1L_2r_1}{\delta_{i+1}}\right)\\
    \leq 2p\log\left( 4K\right) + p(2i + 1) + \nonumber \\ 2k\log\left(\frac{L_1L_2r_1}{\delta} \right) + k(2i+1)\\
    \leq 2p\log\left( 4K\right) + 3pi + 2k\log\left(\frac{L_1L_2r_1}{\delta} \right) + 3ki\\
    \leq 5ip\log(4K) + 5ik\log\left(\frac{L_1L_2r_1}{\delta} \right)
\end{gather}

We choose:
\begin{gather}
    \log f_i = -6ip\log(4K) - 6ik\log\left( \frac{L_1L_2r_1}{\delta}\right)\\   \epsilon_i \leq O(1) + \log(4K)\frac{ip}{m} + i.
\end{gather}
Notice that:
\begin{gather}
    \log |T_i|f_i \leq -ip\log(4K) -ik\log\left( \frac{L_1L_2r_1}{\delta}\right)\leq -im.
\end{gather}

For that choice of parameters, observe that:
\begin{gather}
    P(||At|| \leq (1+\epsilon_i)||t||, \quad \forall i \in [0, ..., l-1], \quad \forall t \in T_i) \\ 
    = 1 - e^{-\Omega(m)}.
\end{gather}

Let $x$ be the image we want to recostruct and $x_i$ be the closest point of that image to the $\delta_i$ net. Then, 
\begin{gather}
    x - x_0 = \sum_{i=0}^{l-1} \left(x_{i+1} - x_{i}\right) + x - x_{l} \Rightarrow \\
    ||Ax - Ax_0|| \leq \sum_{i=0}^{l-1}||Ax_{i+1} - Ax_{i}|| + ||Ax - Ax_{l}||.
\end{gather}
Now w.h.p. $||Ax_{i+1} - Ax_{i}|| \leq (1+\epsilon_i) ||x_{i+1}-x_{i}||$. Therefore, w.h.p.:
\begin{gather}
    ||Ax - Ax_0|| \leq \sum_{i=0}^{l-1}(1+\epsilon_i)||x_{i+1}-x_{i}|| + ||Ax - Ax_{l}|| \\
    \leq \sum_{i=0}^{l-1}(1+\epsilon_i)\delta_i + ||Ax - Ax_l|| \\
    \leq \sum_{i=0}^{\log \frac{\sqrt{p}}{K} - 1}\left( O(1) +  3\cdot 4^{i+1}\right)\frac{\delta}{2^i} + \nonumber \\ + \sum_{i=\log \frac{\sqrt{p}}{K}}^{l-1}\left(O(1) + \log(4K)\frac{ip}{K^2\log 3p} + i\right)\frac{\delta}{2^i} + \nonumber \\
    + ||Ax - Ax_l|| \\ 
    \leq O\left( \log(4K) \cdot \frac{\sqrt p}{K} \cdot \log\frac{\sqrt p}{K} \right) \cdot \delta + ||Ax - x_l||.
\end{gather}

Observe that:
\begin{gather}
    ||Ax - Ax_{l}|| \leq ||A|| \cdot ||x - x_l|| \\ 
    \leq 2\sqrt{n}||x - x_l|| \\
    \leq 2\frac{\sqrt{n}}{2^l}\delta.
\end{gather}
For $l = \log n$, we have that $||Ax - Ax_l|| \leq \delta$.
Hence,
\begin{gather}
    ||Ax - Ax_0|| \leq O\left( \log(4K) \cdot \frac{\sqrt p}{K} \cdot \log\frac{\sqrt p}{K} \right) \cdot \delta.
\end{gather}
\end{proof}

\subsubsection{Proof of main theorem}
\begin{delayedproof}{main-theorem-l1}
Let $\delta_{l_1} = \left(\log(4K) \cdot \frac{\sqrt{p}}{K} \log\frac{\sqrt p}{K}\right)\delta$.
Then, 
\begin{gather}
    ||G_2(\bar z^p)  - G_2(\tilde z^p)|| 
    \leq \\ \frac{||AG_2(\tilde z^p) - AG_2(\bar z^p)|| + \delta_{l_1}}{\gamma}\\
    \leq \frac{||Ax - AG_2(\bar z^p)|| + ||Ax - AG_2(\hat z^p)|| + \delta_{l_1}}{\gamma} \\
    \leq \frac{2||Ax - AG_2(\bar z^p)|| + \delta_{l_1}}{\gamma} \\
    \leq \frac{4||G_2(\bar z^p) -x|| +  \delta_{l_1}}{\gamma}.
\end{gather}

Finally, observe that:
\begin{gather}
    ||G_2(\tilde z^p)  - x|| \leq ||G_2(\bar z^p) - x|| + ||G_2(\bar z^p) - G_2(\tilde z^p)|| \\
    \leq \left( 1 +  \frac{4}{\gamma}\right)||x - G_2(\bar z^p)|| +  \frac{\delta_{l_1}}{\gamma}.
    \label{gammascaling}
\end{gather}
\end{delayedproof}

\begin{remark}
Similar to the analysis of the CSGM paper (see Lemma 4.3), $\gamma$ is a constant that we control and we may set it to $\gamma=\frac{4}{5}$ to get the same scaling term with CSGM.
\end{remark}

\subsubsection{Proof of Lemma~\ref{s_rec_circulant}}
\begin{lemma}
Consider the setting of Theorem~\ref{main-theorem-l1}. Let $g = [g_1, \cdots, g_n]$ be a vector with i.i.d. Gaussian entries of variance $1/m$, let $F \in \R^{m\times n}$ be a partical circulant matrix that has $g$ in its first row, and let $D\in \R^{n\times n}$ be a diagonal matrix with uniform $\pm 1$ entries along its diagonal. Then for $m = \Omega\left( \frac{1}{(1-\gamma)^2}(k \log \frac{L_1 L_2 r_1}{\delta } + K^2 \log p)  \log^4(n)\right)$,  $FD$ satisfies S-REC$(G_2(G_1(\ballpdr{2}{k}{r_1}) \oplus \ballpdr{1}{p}{r_2}), 1-\gamma, \delta \cdot \frac{\log(4K)}{\gamma}\cdot\frac{\sqrt{p}}{K} \log\frac{\sqrt{p}}{K})$ with probability $1 - e^{-\Omega(m)}.$
\end{lemma}

\begin{proof}
The proof follows from the proof of Lemma~\ref{s_rec_for_nested_l1} above and Theorem 3.1 in~\cite{krahmer2011new}. The proof of Lemma~\ref{s_rec_for_nested_l1} requires the Johnson-Lindenstrauss guarantee for a set of size $2^{O(m)}$, and invoking Theorem 3.1 in~\cite{krahmer2011new}, this is guaranteed to hold for the matrix $FD$.

The proof of Lemma~\ref{s_rec_for_nested_l1} also requires $\| FD\|_{op}\leq \sqrt{n}$. This is also guaranteed by noting that
\begin{align*}
    \norm{FD}_{op} & \leq \norm{FD}_F \leq \sqrt{n}, \; w.p. 1 - e^{-\Omega(m)}.
\end{align*}
\end{proof}

\subsection{Code}
Our source-code is available under the following url: \href{https://github.com/giannisdaras/ilo}{https://github.com/giannisdaras/ilo}.

Our code is implemented in PyTorch~\cite{pytorch}. Our code is based on the following open-source implementations of StyleGAN-2: \href{https://github.com/rosinality/stylegan2-pytorch}{https://github.com/rosinality/stylegan2-pytorch}, \href{https://github.com/NVlabs/stylegan2}{https://github.com/NVlabs/stylegan2}. We also draw inspiration from the open-source implementation of PULSE: \href{https://github.com/tg-bomze/Face-Depixelizer}{https://github.com/tg-bomze/Face-Depixelizer}. A Tensorflow~\cite{abadi2016tensorflow} implementation is in the works. 

Our current repository includes:
\begin{itemize}
    \item Detailed instructions on how to setup the environment and download the dependencies.
    \item Code for image pre-processing, such as random inpainting, interactive masks, noise addition, automatic face alignment, etc.
    \item Examples on how to run inpainting, denoising, super-resolution and compressed-sensing with circulant matrices for custom images.
    \item Code for out-of-distribution generation on $1000$ ImageNet~\cite{imagenet_cvpr09} classes using a robust classifier. We use a robust classifier from the \texttt{robustness}~\cite{robustness} library.
    \item Code for evaluating the performance of ILO and previous methods on Celeba-HQ~\cite{celeba, celebahq}.
    \item Tools to visualize performance and track experiments.
    \item Code for generating GIF files by collecting frames during the optimization.
\end{itemize}

Our code is GPU/CPU compatible. 
\subsection{Experimental details}
We performed all our experiments on a single GPU. As mentioned in the paper, obtaining a solution for a single inverse problem requires less than a minute on a single 1080Ti. All the experiments can be reproduced in less than a day on a single GPU.

Unless mentioned otherwise, we use Adam~\cite{kingma2014adam} optimizer with an initial learning rate of $0.1$ for each layer. 
%We use the cosine scheduler for our learning rate, as proposed by \cite{stylegan2}.
%cycle our learning rate (similar to ~\citet{smith2017cyclical}) when we switch to the next layer. 
During a single layer optimization, learning rate ramps up linearly and is ramped down using a cosine scheduler, as proposed by \cite{stylegan2}.

Loss functions are changed for each task as explained in the paper. For all tasks, we use a geodesic loss with coefficient $0.01$. For random inpainting, we use both MSE and LPIPS when we have more than $20\%$ observed pixels, otherwise we only use MSE. When both MSE and LPIPS are used,  we search co-efficients in the set $\{0.5, 1, 2, 5\}$ for each of the terms. For inpainting with continuous black boxes, we used both MSE and LPIPS. For the experiments of Figure 1 of the main paper, we used the same co-efficient for both MSE and LPIPS. 

Our optimization algorithm is Projected Gradient Descent~\cite{nocedal2006numerical}. First, we project each latent code to the unit sphere. Next, when optimizing over deeper layers, we use $l_1$ projection to stay close to the manifold induced by the previous layers. The projection in that case includes the solution of the previous layers, the latent codes (i.e. $w_i$) and the noises, (i.e. $u_i$). We tune seperately the $l_1$ radii for each one of the optimization variables and for each one of the layers. Empirically, we find that the following radii for the first four layers works decently for most of the tasks/images:
\begin{itemize}
    \item Radius of noises: $300, 2000, 2000, 4000$.
    \item Radius of latent codes: $300, 500, 1000, 2000$.
    \item Radius of previous solutions: $500, 1000, 2000$.
\end{itemize}

Projection to the $l_1$ ball allows for optimization on deep layers of the generator (that is not possible without projections).  By doing that we get better reconstruction that comes with the cost of increased number of optimization steps. Generally, tuning the radii for each layer is an especially difficult procedure. Even worse, these hyperparameters do not transfer across tasks. For the first four layers, we encourage the reader to use the parameters mentioned above.

To obtain the plots of Figure \ref{mse_plots}, we sampled (randomly) 5 images from Celeba-HQ and we reported the best score for each point on the horizontal axis over 5 different runs (25 runs in total for each method for each point in the plot) with different hyperparameters. The error bars are computed across the experiments for different images. For the plots of Figure \ref{mse_plots}, we searched over the following combinations of number of steps for each layer (starting from the first): $\{300,200,200,100\}, \{300,200,100\}, \\ \{300,200,200,100,50\}, \{50, 50, 50, 50, 500\}, \\ \{100, 100, 100, 100, 100\}$. Each reported point is the average (across images) of the minimums of those runs.

\subsection{Additional Experiments}
In this section, we list additional Figures and Experiments that could not fit in the paper due to space limit. 

\begin{figure*}[!htbp]
\captionsetup[subfigure]{labelformat=empty,font=ltpt}
\captionsetup[subfigure]{justification=centering}
\begin{adjustbox}{width=0.8\textwidth, center}
\begin{tabular}{cccc}
\centering
\subfloat[]{\includegraphics[width=0.25\textwidth]{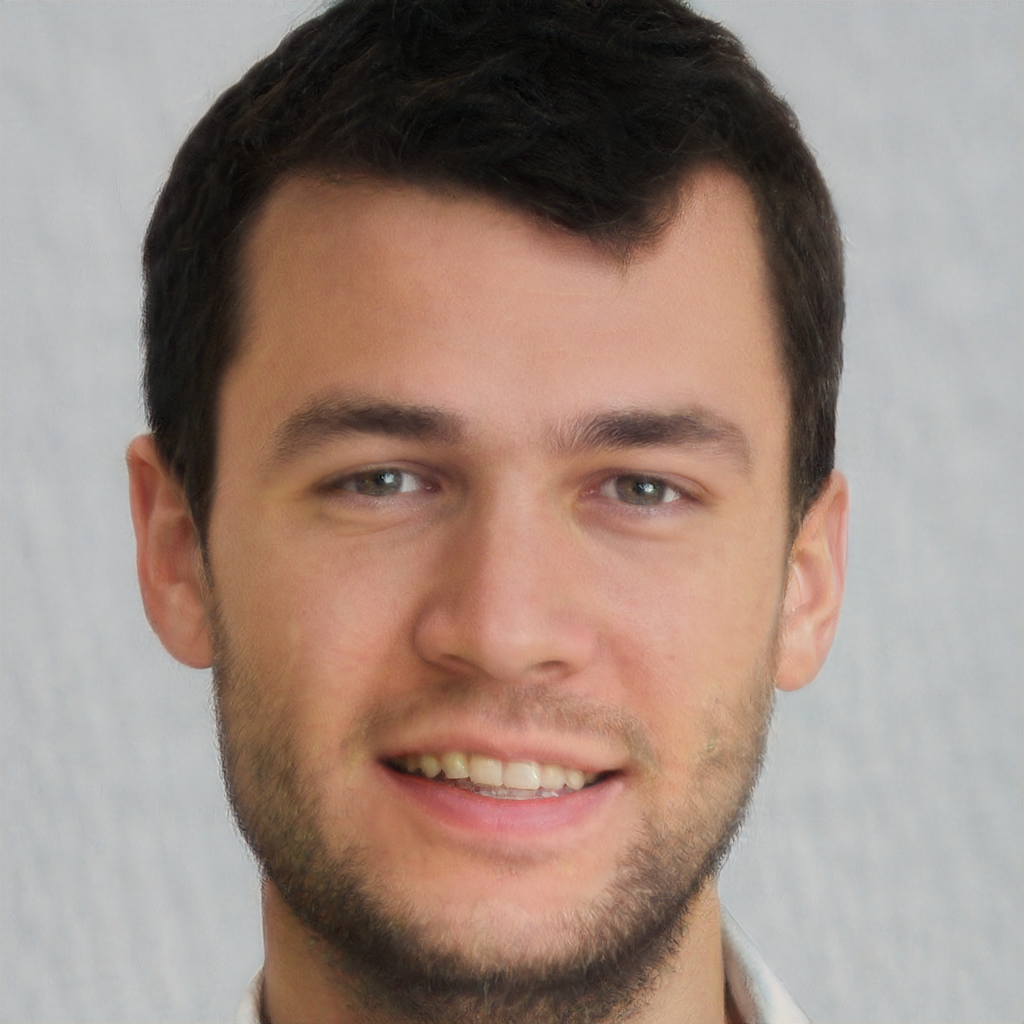}} & \hspace{-4mm}
\subfloat[]{\includegraphics[width=0.25\textwidth]{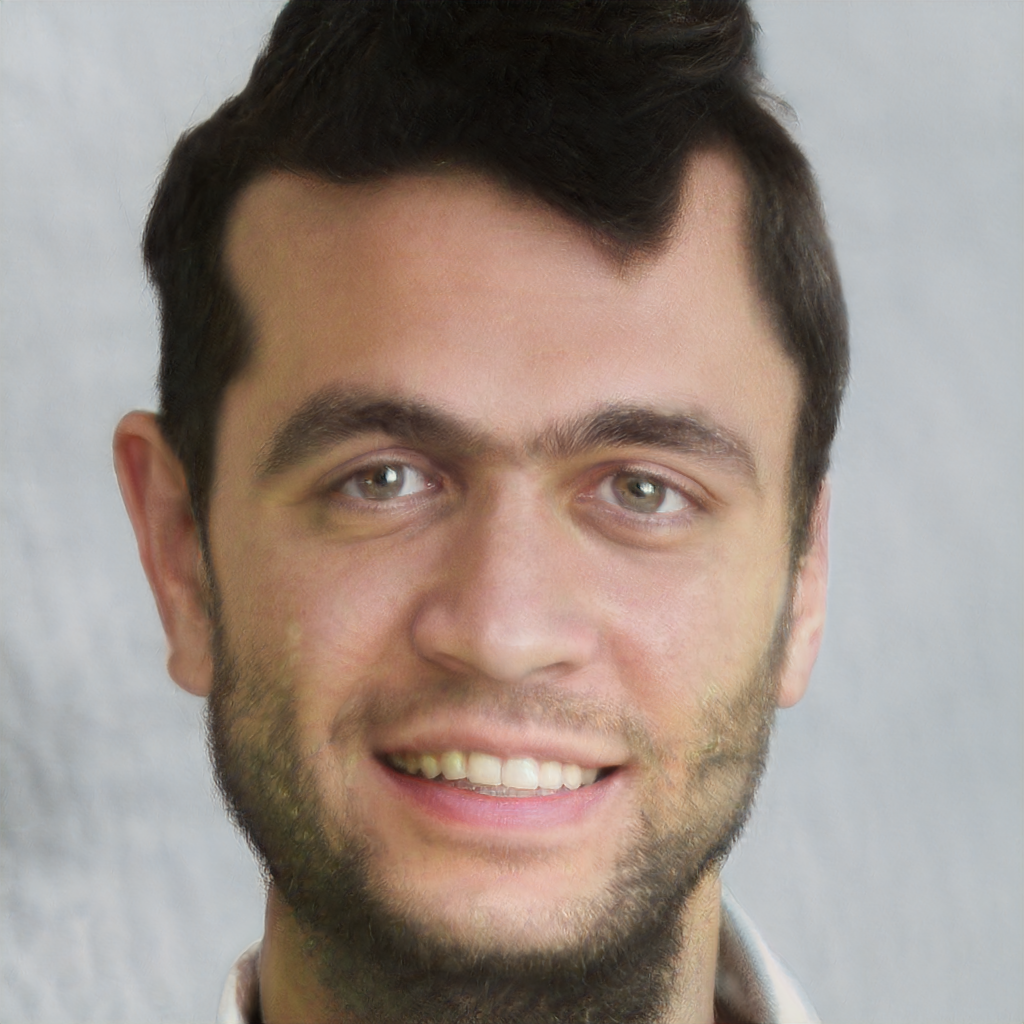}} & \hspace{-4mm}
\subfloat[]{\includegraphics[width=0.25\textwidth]{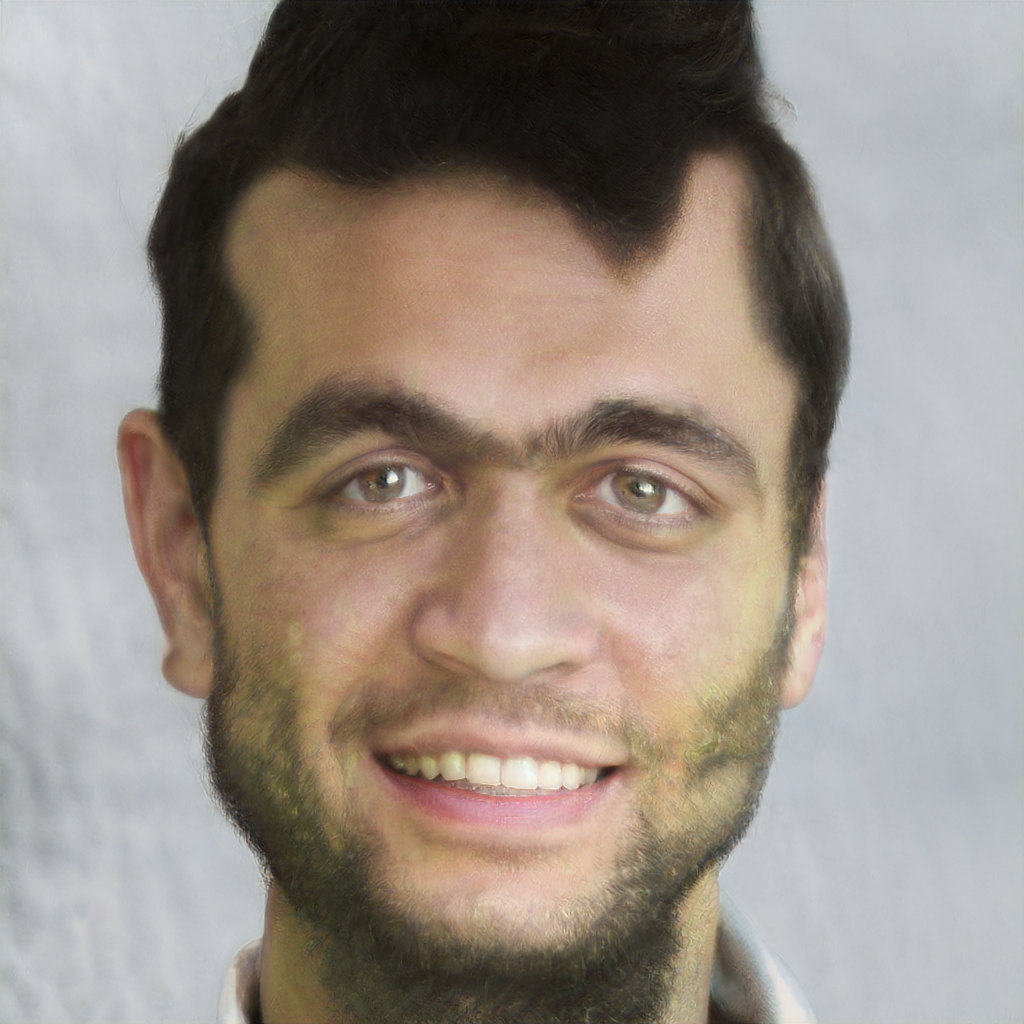}} & \hspace{-4mm}
\subfloat[]{\includegraphics[width=0.25\textwidth]{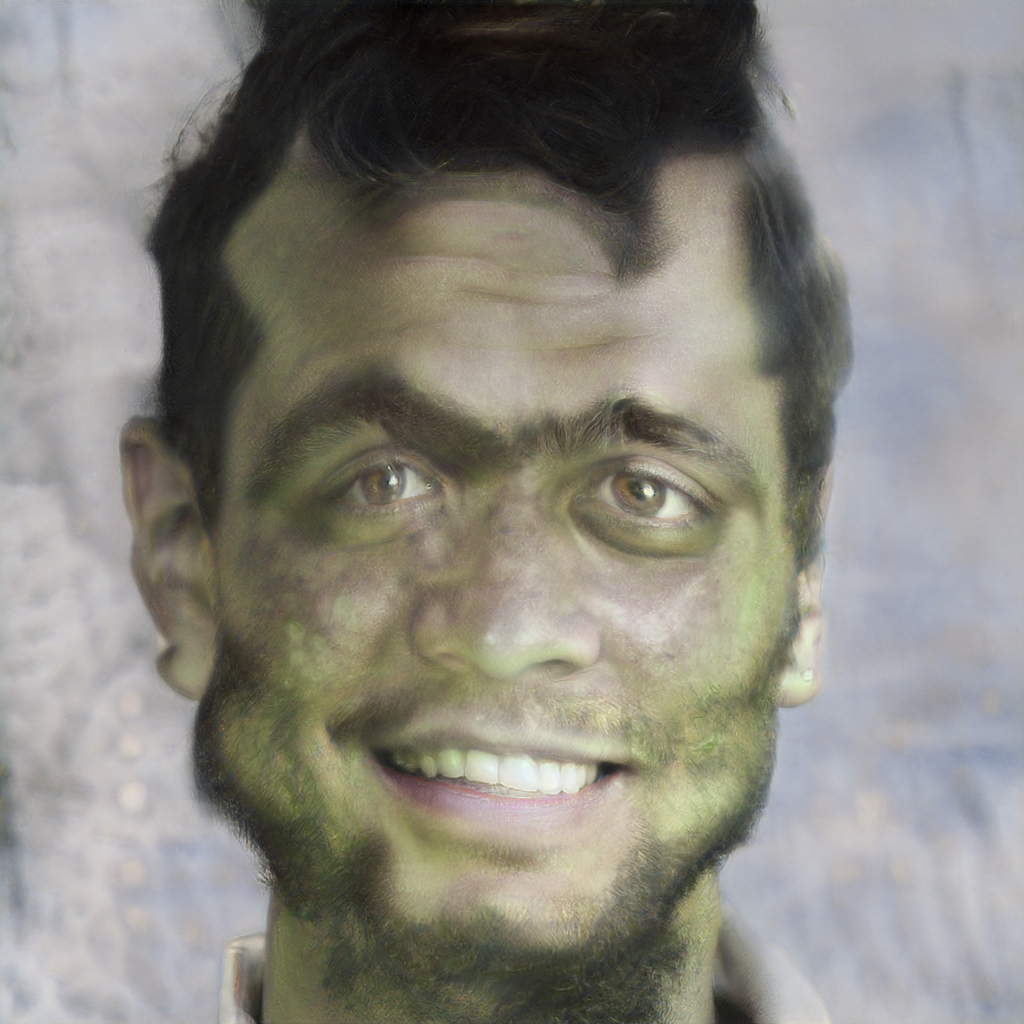}}
\end{tabular}
\end{adjustbox}
\caption{Morphing using a classifier for Bull Frog class, keeping also a loss term for distance to a well-known machine learning researcher, with his permission.}
\label{morphing}
\end{figure*}

Figure \ref{morphing} shows that by combining MSE loss to a reference image and the classification probability, we can morph a given person to an ImageNet~\cite{imagenet_cvpr09} class. We observe experimentally that better results are obtained by only using MSE and LPIPS loss during the first ILO rounds and only using the additional classification term in the deeper layers of the generator. An extra benefit of this method is that we can interpolate intermediate frames to see how actually a human face can be transformed to an imagenet class since the generator first matches the phase and then uses the classifier to alter it.

\begin{figure*}[!htp]
\captionsetup[subfigure]{labelformat=empty,font=ltpt}
\captionsetup[subfigure]{justification=centering}
\begin{adjustbox}{width=0.8\textwidth, center}
\begin{tabular}{ccccc}
\captionsetup{justification=centering}
\subfloat[Original]{\includegraphics[width=0.4\textwidth]{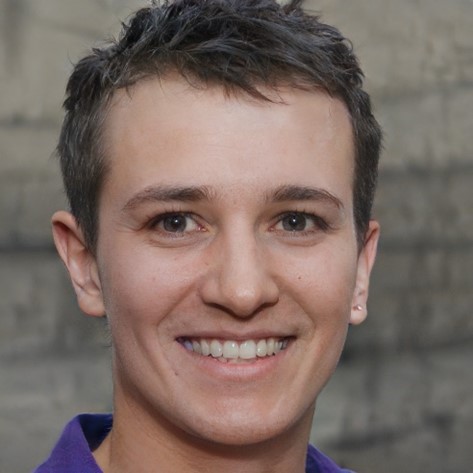}} & \hspace{-4mm}
\subfloat[Gaussian Noise 23.93 dB][Noisy (23.9dB)]{\includegraphics[width=0.4\textwidth ]{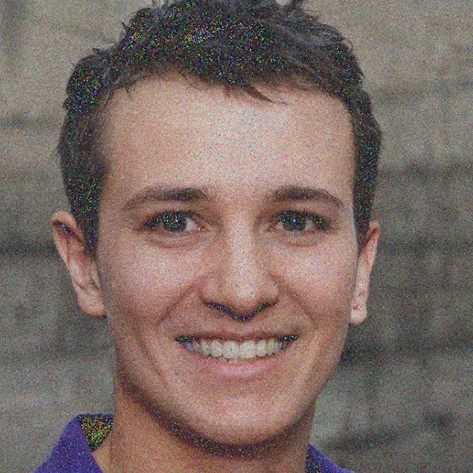}} & \hspace{-4mm}
\subfloat[MSE 26.48 dB][MSE (26.5dB)]{\includegraphics[width=0.4\textwidth]{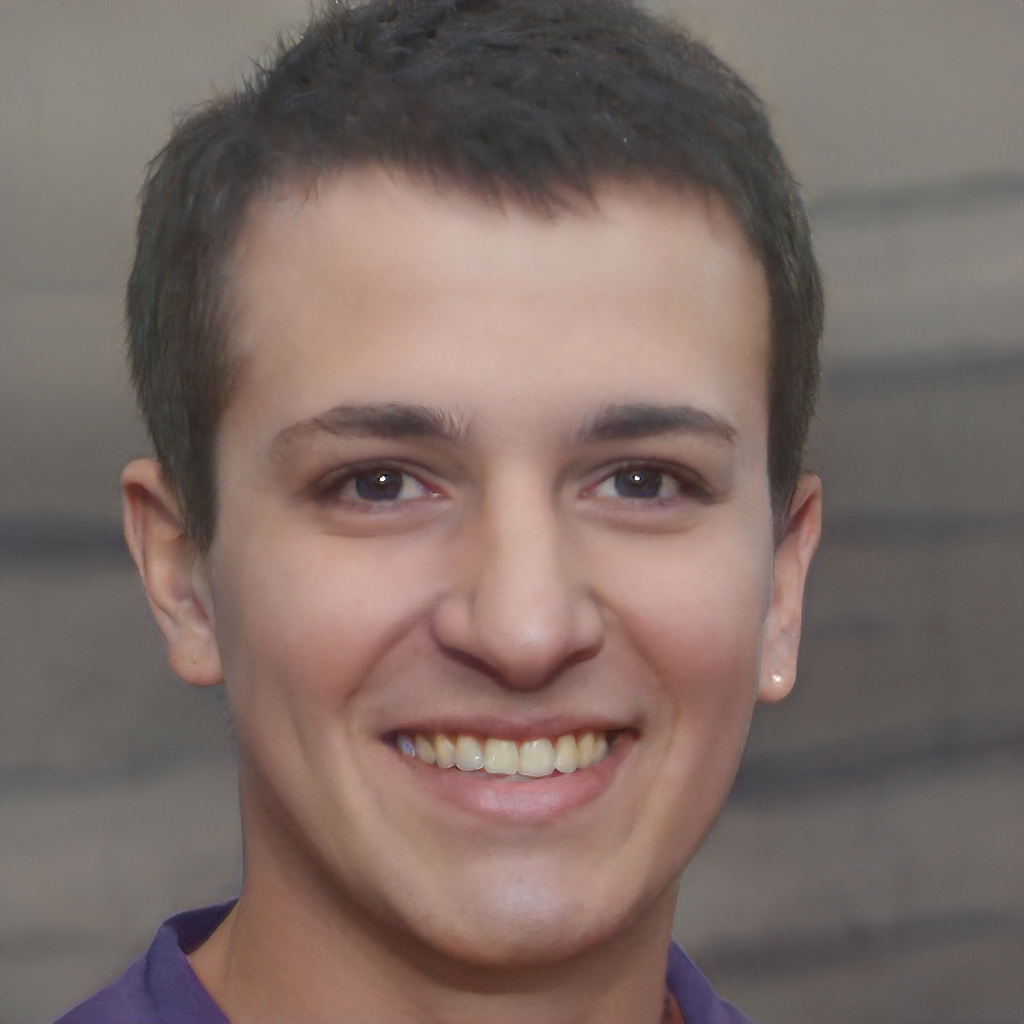}} & \hspace{-4mm}
\subfloat[BM3D 27.65 dB][BM3D (27.6dB)]{\includegraphics[width=0.4\textwidth]{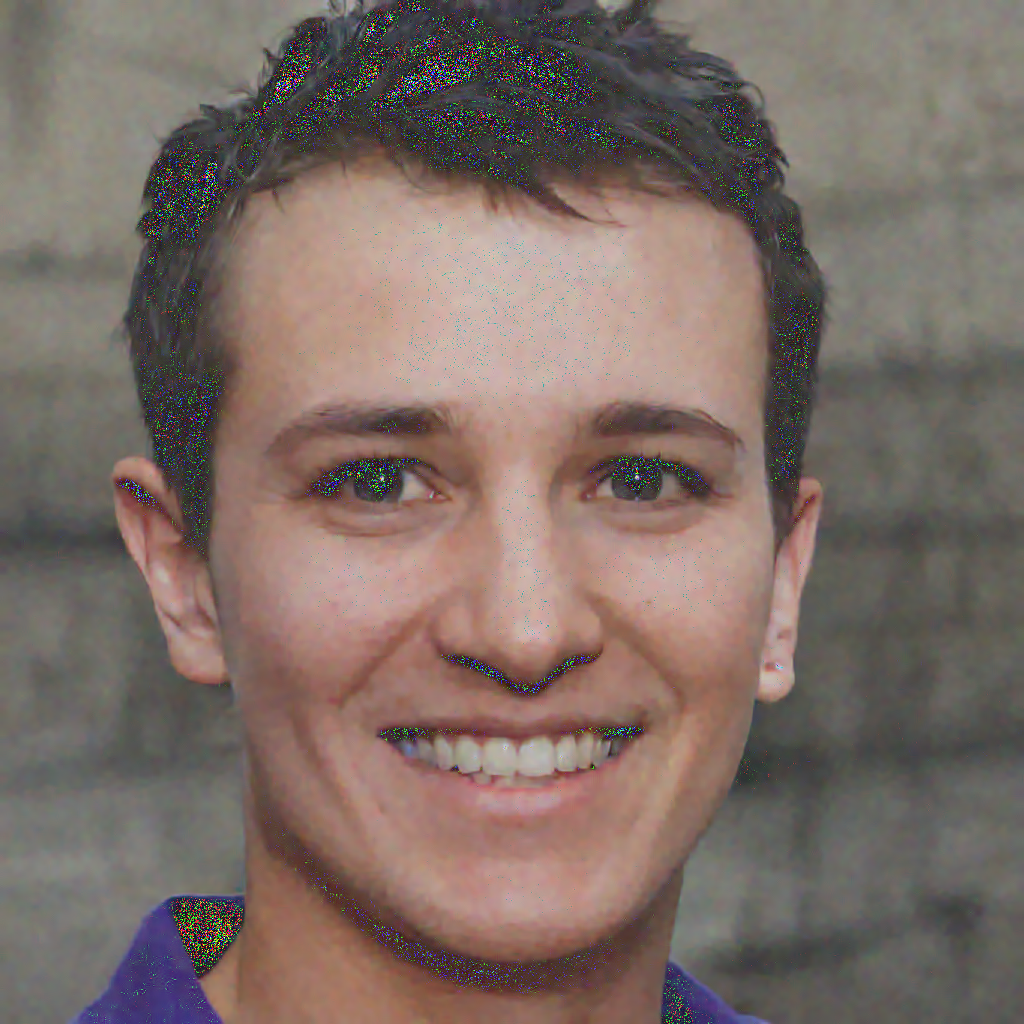}} & \hspace{-2mm}
\subfloat[Recovered 30.01dB][Ours (30.0dB)]{\includegraphics[width=0.4\textwidth]{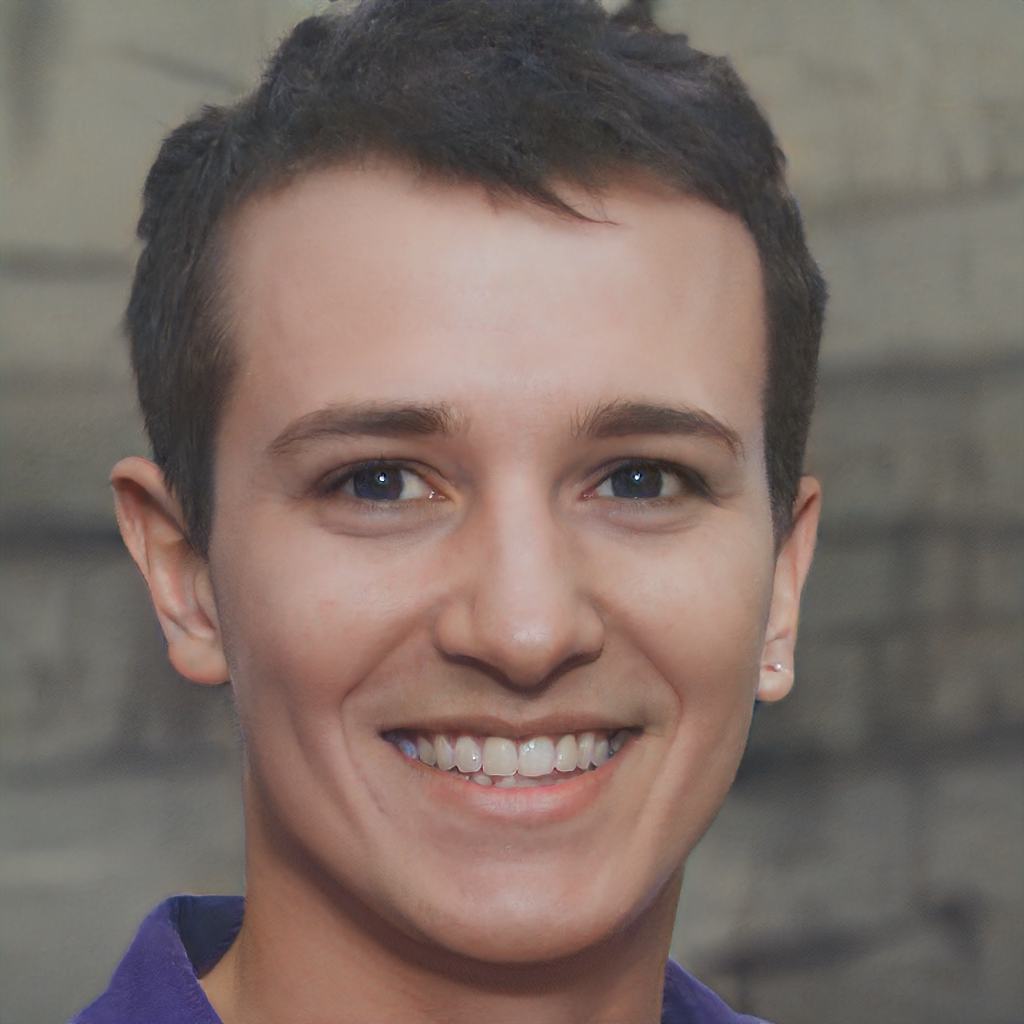}} \\
\subfloat[Original]{\includegraphics[width=0.4\textwidth]{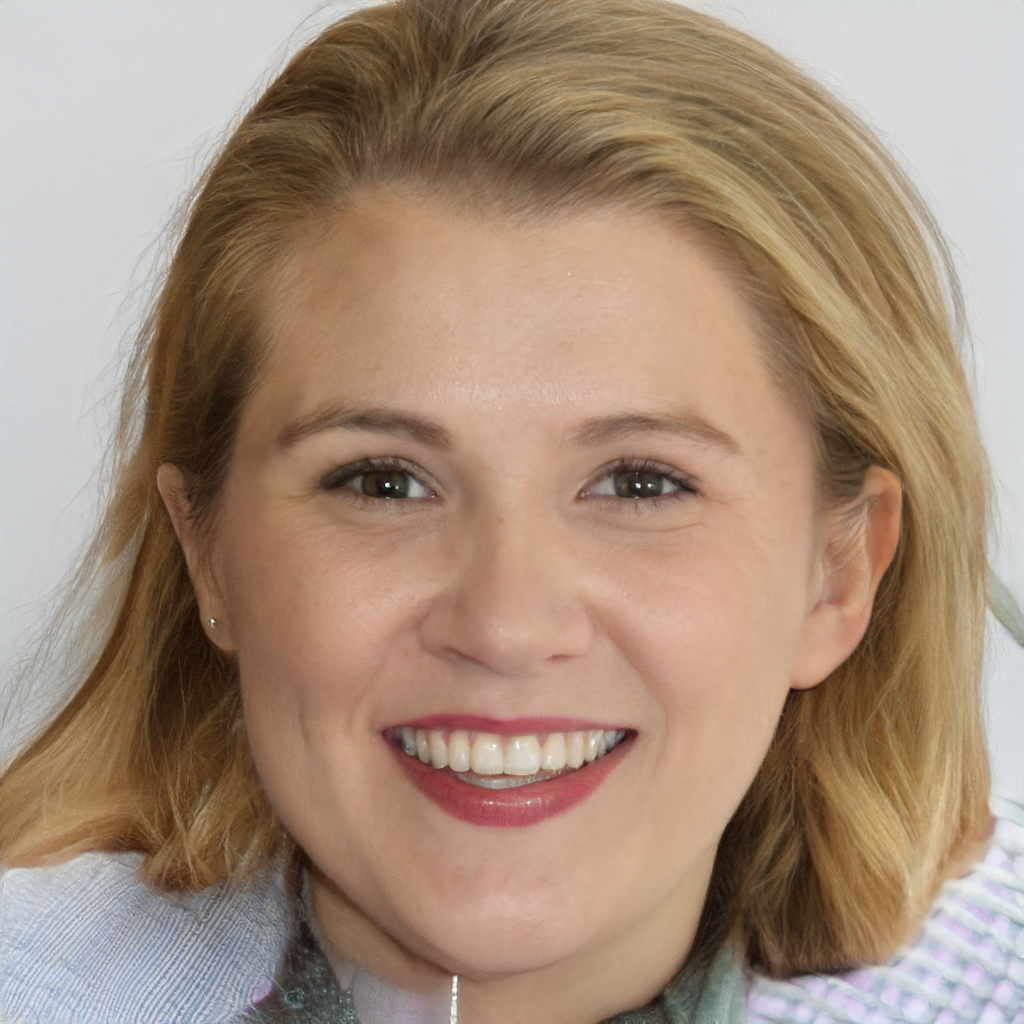}} & \hspace{-4mm}
\subfloat[Gaussian Noise 21.88 dB][{Noisy (21.8dB)}]{\includegraphics[width = 0.4\textwidth]{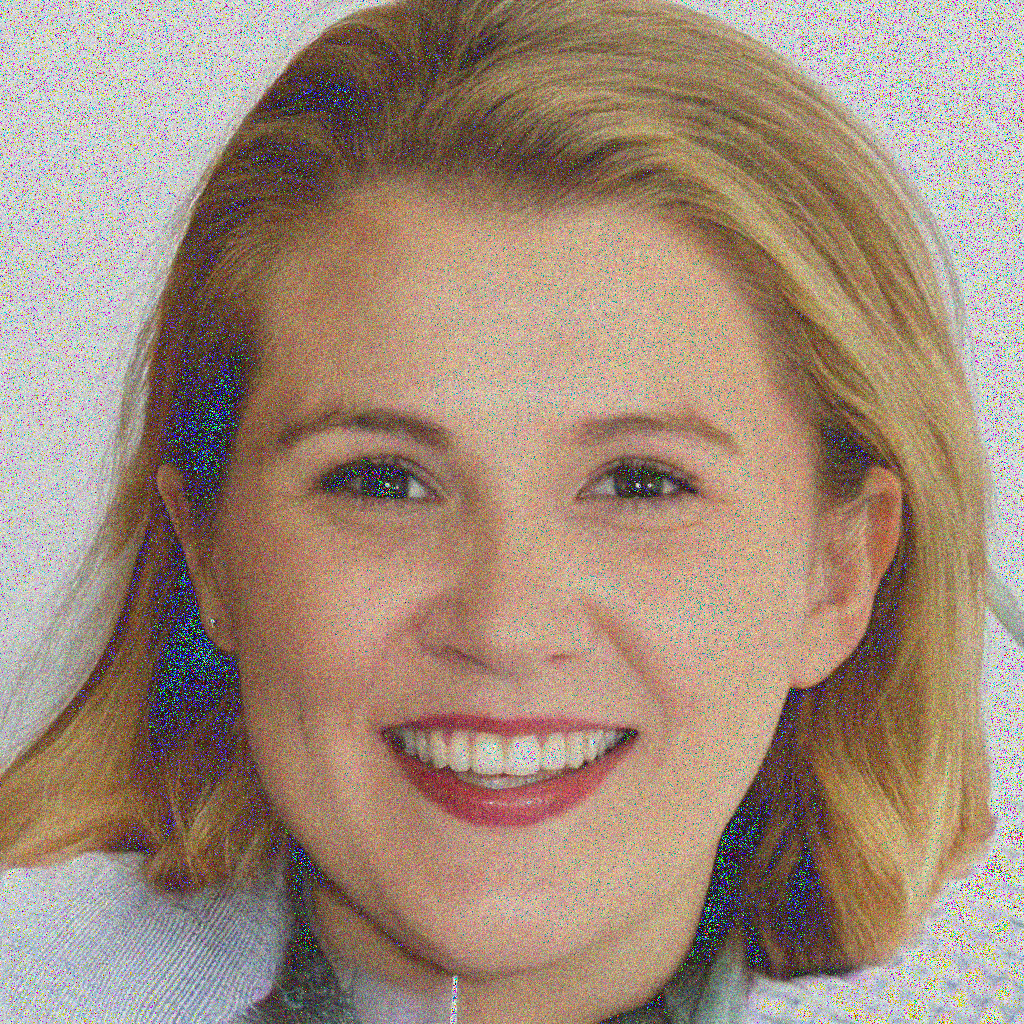}} & \hspace{-4mm}
\subfloat[MSE 23.08 dB][MSE (23.0dB)]{\includegraphics[width = 0.4\textwidth]{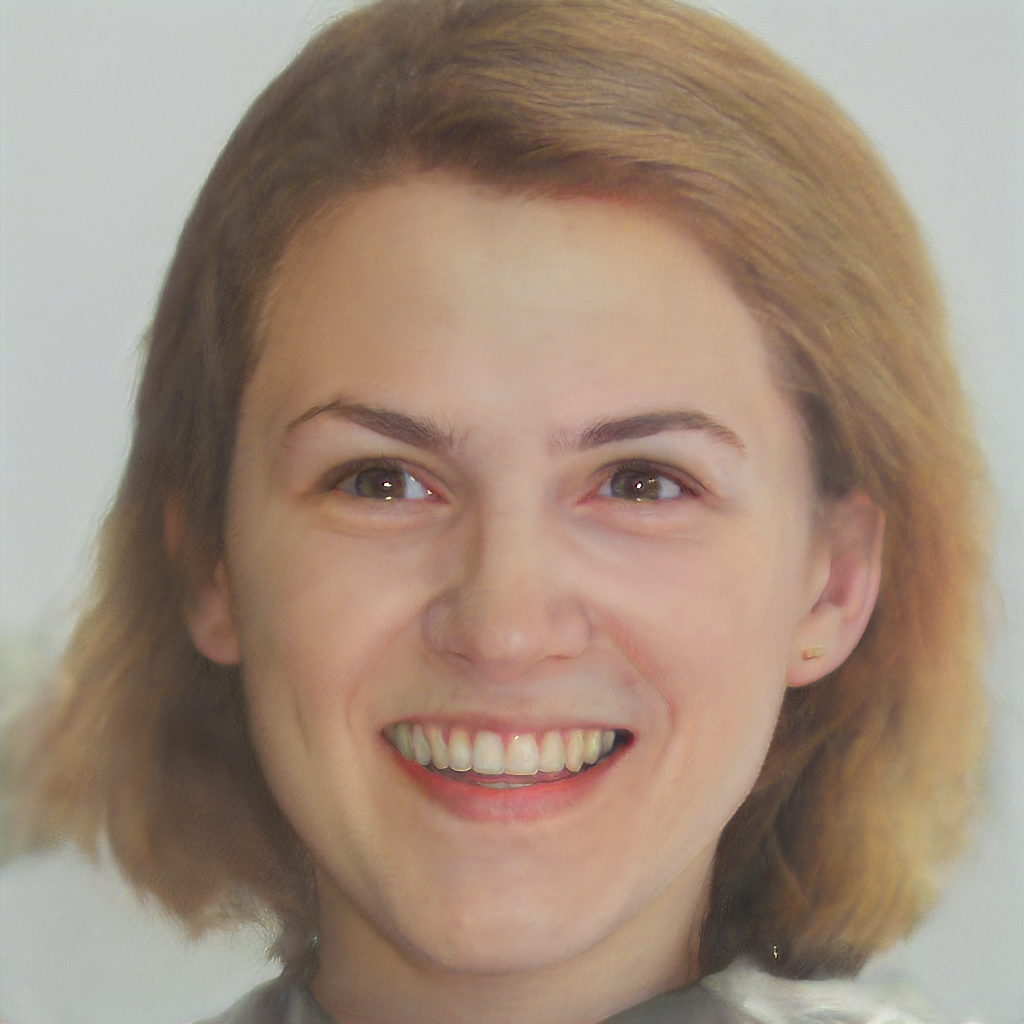}} & \hspace{-4mm}
\subfloat[BM3D24.4dB][BM3D (24.4dB)]{\includegraphics[width = 0.4\textwidth]{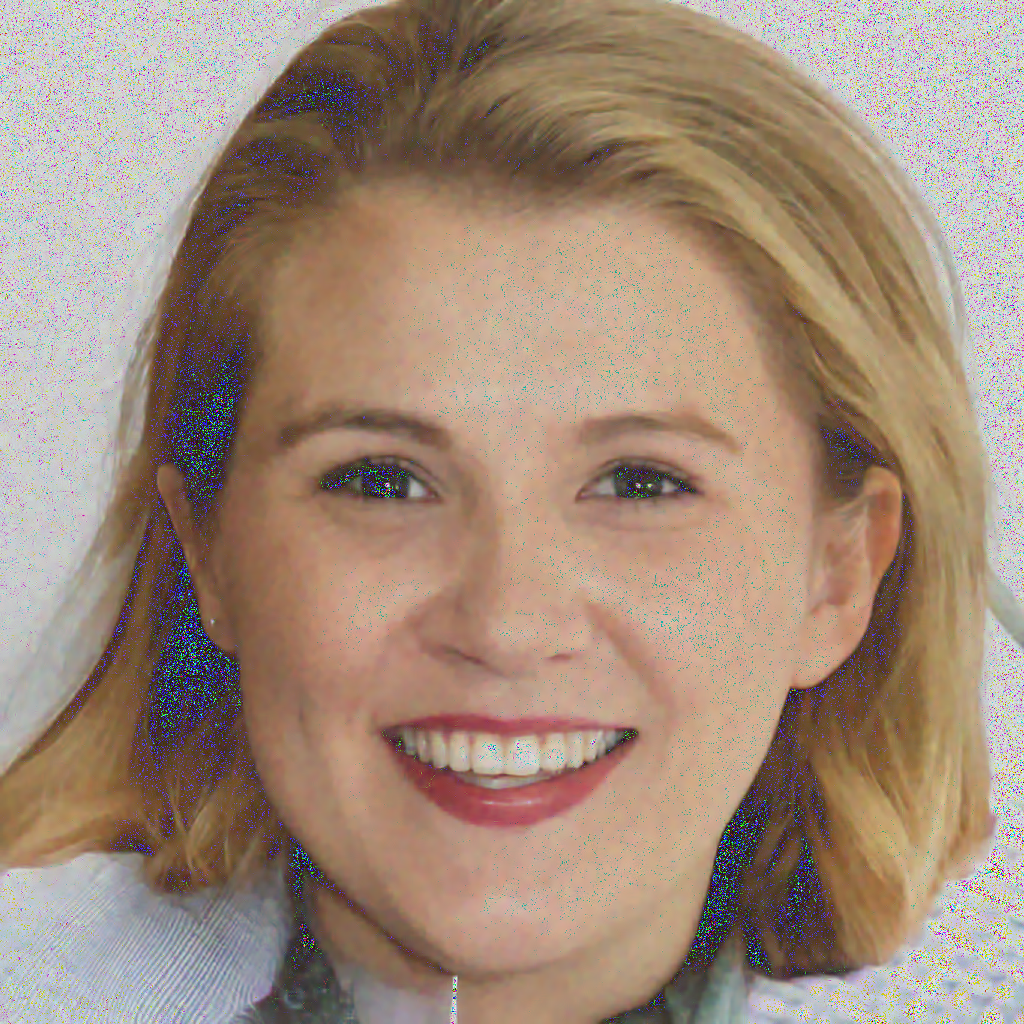}} & \hspace{-4mm}
\subfloat[Recovered 27.08 dB][Ours (27.1dB)]{\includegraphics[width = 0.4\textwidth]{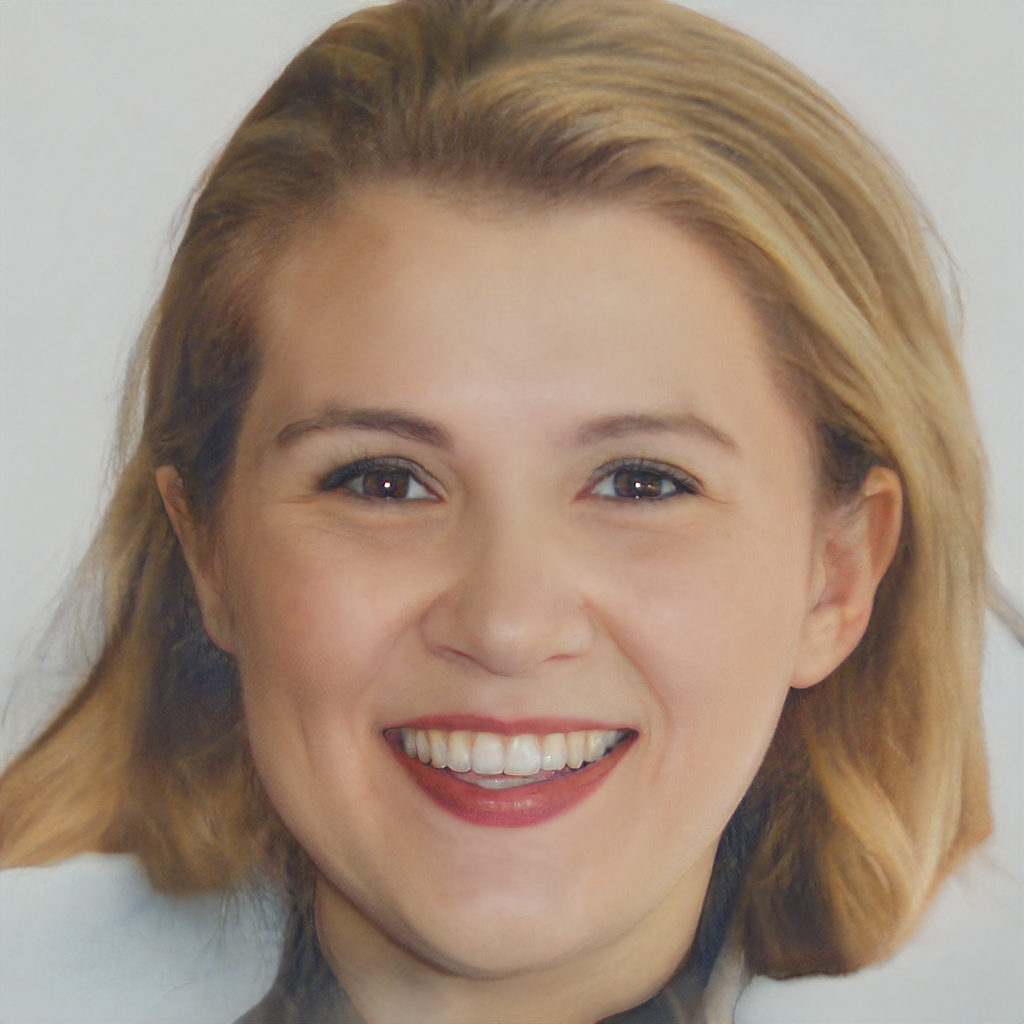}} \\
\end{tabular}
\end{adjustbox}
\caption{Results on the task of denoising. Gaussian noise $(\sigma=25$, known) is added to the original image and recovered with various methods. The MSE images indicate the reconstructed images obtained by inverting the noisy image.}
\label{denoising_results}
\end{figure*}

Figure \ref{denoising_results} shows visual results for the task of denoising. As shown, ILO gives superior visual reconstructions and better actual performance comparing to the (adapted for denoising) PULSE and the classical BM3D method.

\begin{figure*}[!htbp]
\captionsetup[subfigure]{labelformat=empty,font=ltpt}
\captionsetup[subfigure]{justification=centering}
\begin{adjustbox}{width=0.8\textwidth, center}
\begin{tabular}{ccc}
\centering
\subfloat[][]{\includegraphics[width=0.10\textwidth]{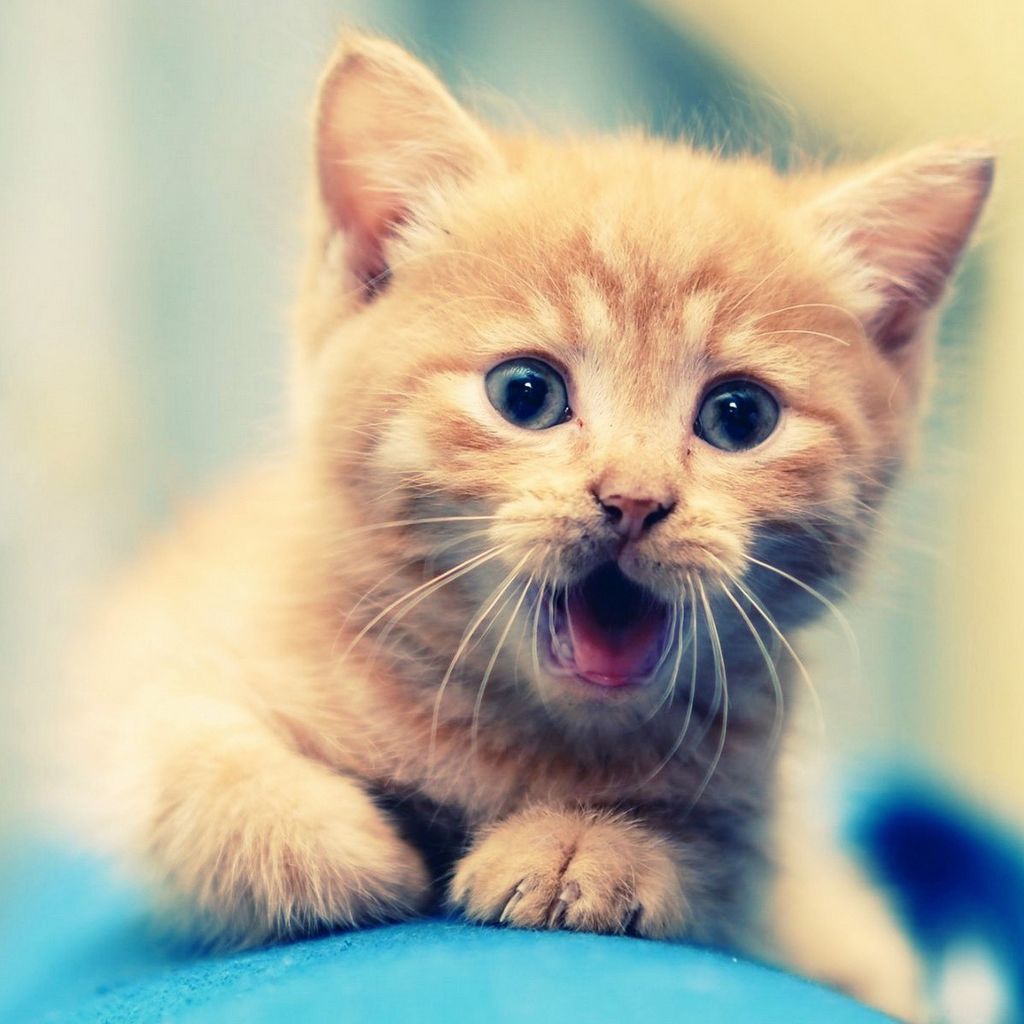}} & \hspace{-4mm}
\subfloat[][]{\includegraphics[width=0.10\textwidth]{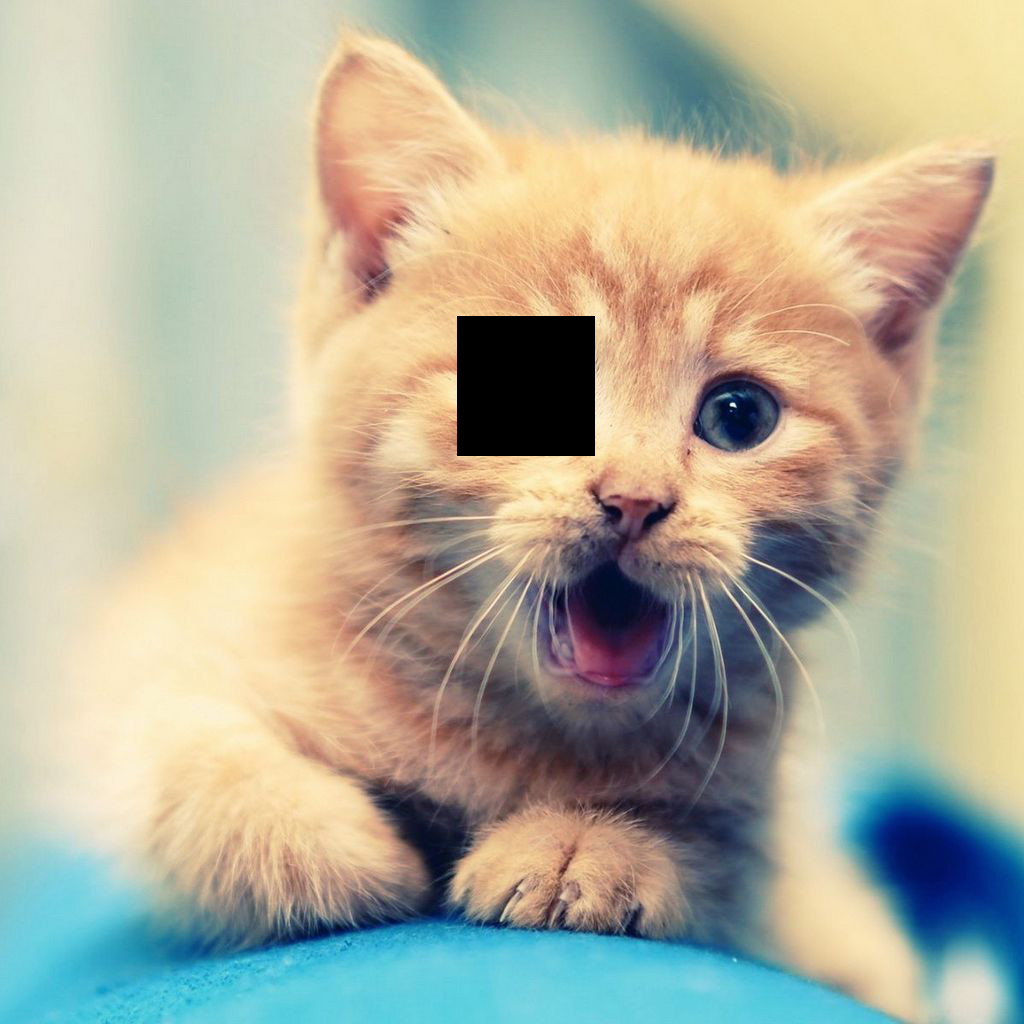}} & \hspace{-4mm}
\subfloat[][]{\includegraphics[width=0.10\textwidth]{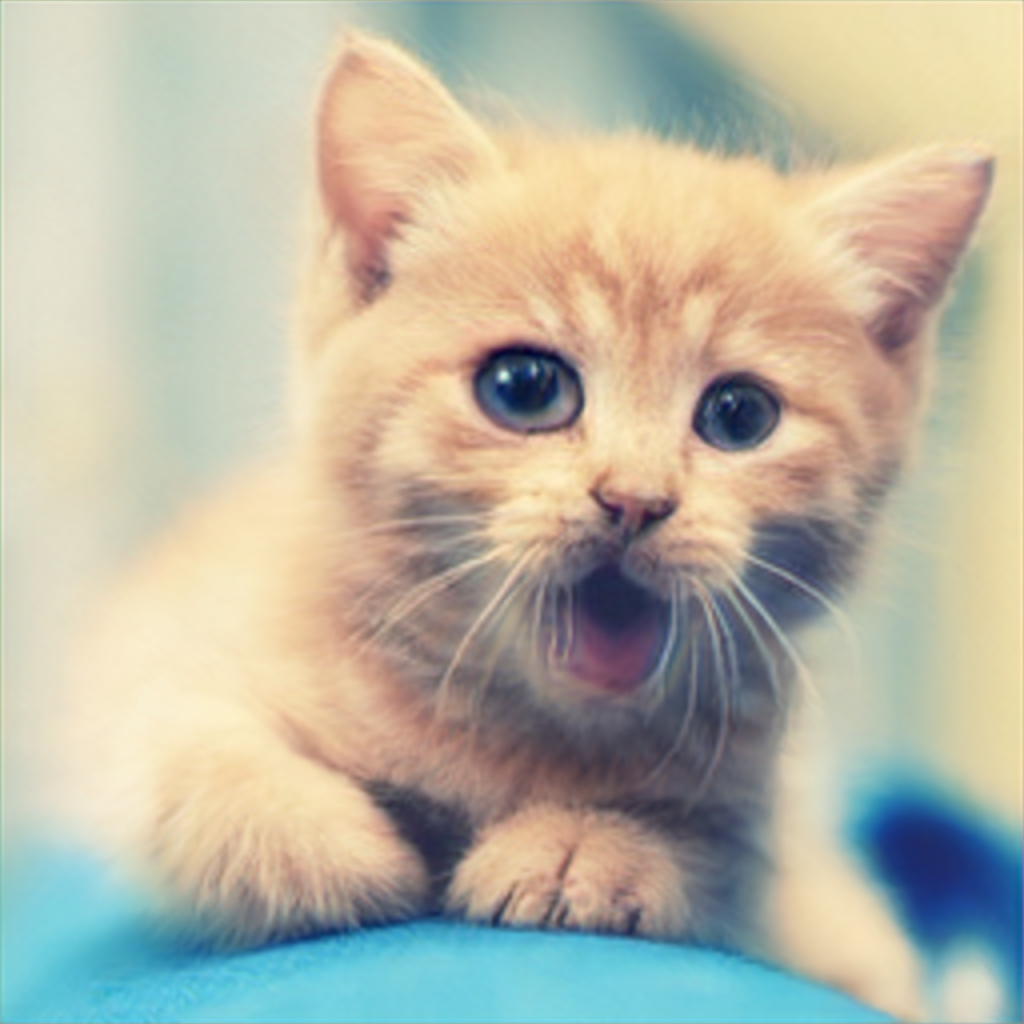}} \\
\subfloat[][]{\includegraphics[width=0.10\textwidth]{cats/cat.jpg}} & \hspace{-4mm}
\subfloat[][]{\includegraphics[width=0.10\textwidth]{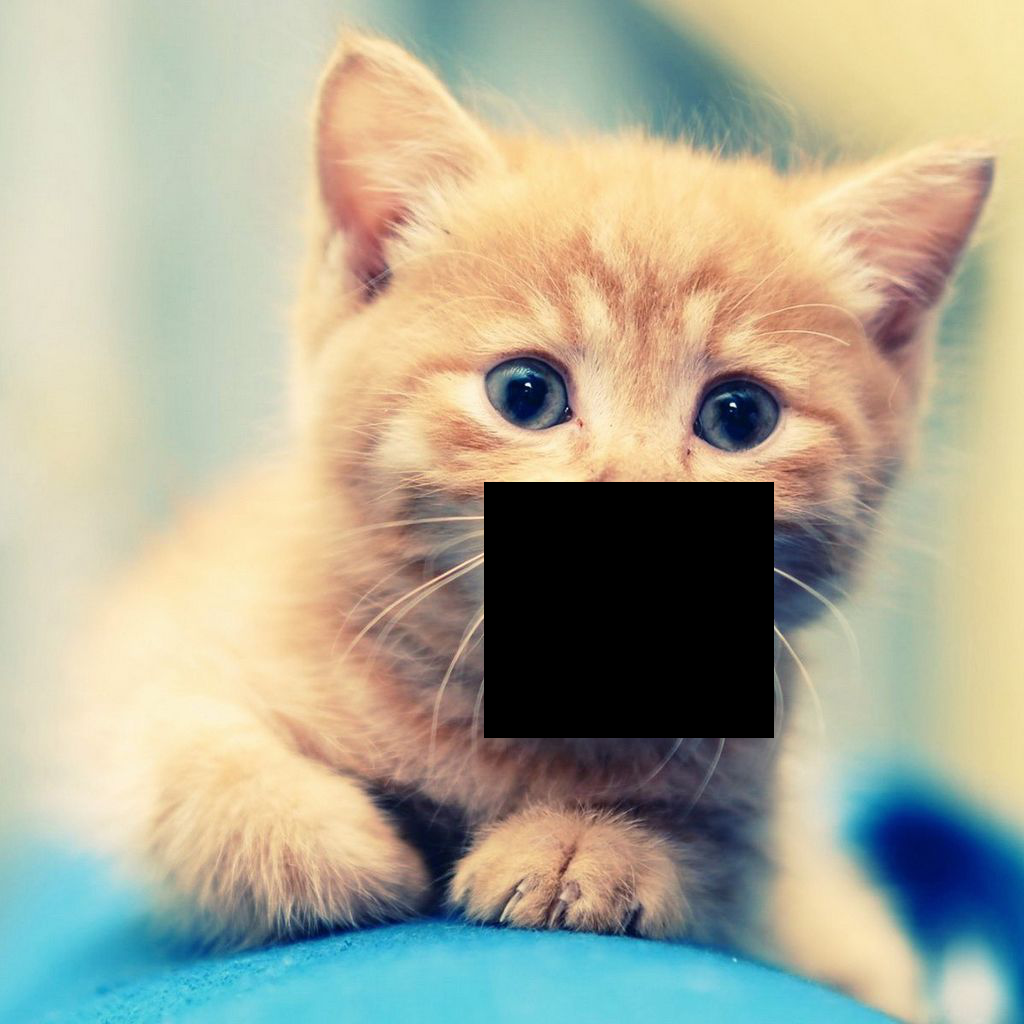}} & \hspace{-4mm}
\subfloat[][]{\includegraphics[width=0.10\textwidth]{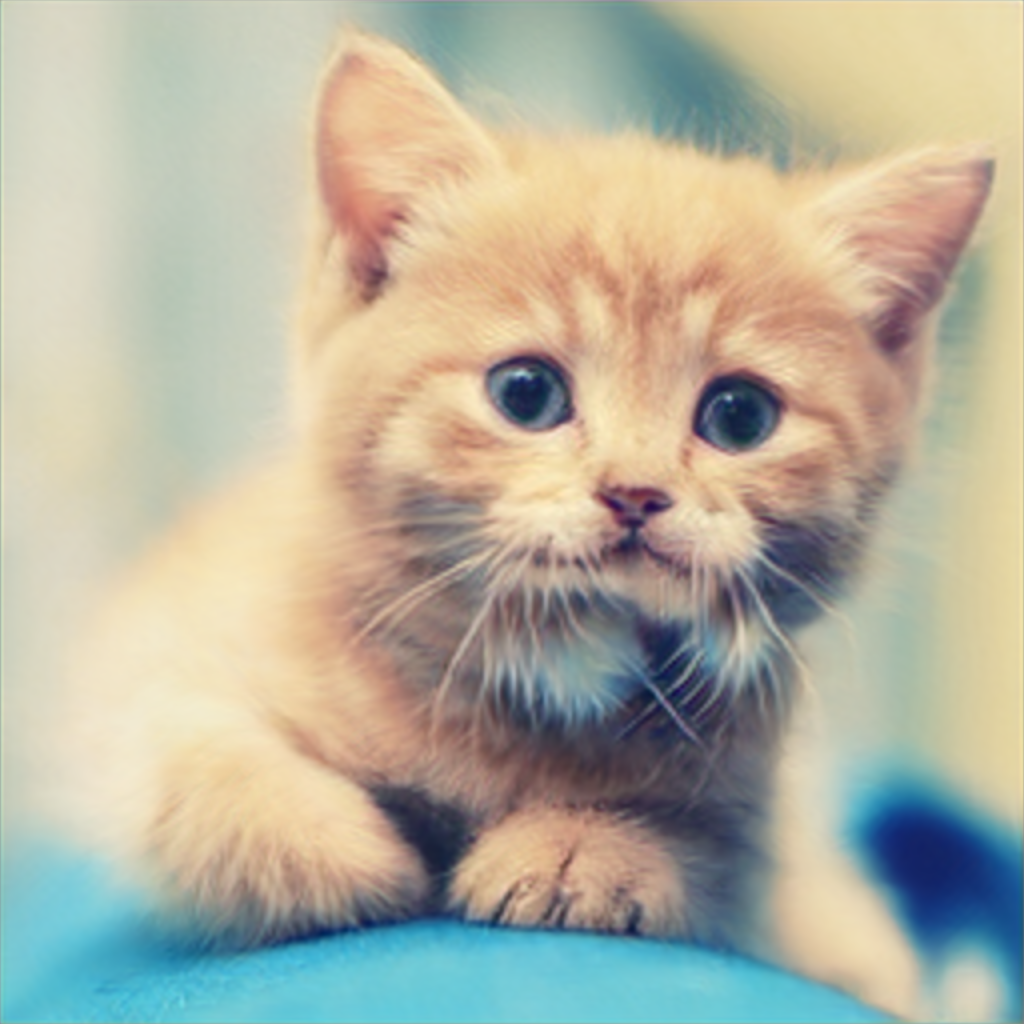}} \\
\end{tabular}
\end{adjustbox}
\caption{Inpainting using a StyleGAN trained to generate cat images. First column: Original image (never observed). Second column: Observed image. Third column: ILO reconstruction.}
\label{cats}
\end{figure*}

Finally, in order to show that our method can be successfully in other datasets as well, we perform inpainting experiments using a pre-trained StyleGAN-2 generators on cats. Results are shown in Figure \ref{cats}.

\FloatBarrier

\FloatBarrier

\subsection{Ethical Considerations}
Previous research has reported that StyleGAN leads to biased generations~\cite{pulse, jain2020imperfect, tan2020improving, salminen2020analyzing}. In practice, we observe that by extending the range of the generator we obtain more diverse generations. A similar finding has been reported by ~\citet{abdal2019image2stylegan}. Even though we observe less biased reconstructions, we encourage a lot more research on this topic.

Admittedly, our method makes the creation of DeepFakes~\cite{korshunov2018deepfakes} easier, since it expands the range of the generator. Arguably, the technology behind DeepFakes is already very powerful so the negative effect of this work will be diminishing. An interesting topic of research is whether existing defences against DeepFakes~\cite{matern2019exploiting, guera2018deepfake, nguyen2019deep, yang2020defending} are robust to images that lie outside of the range of the GAN.

Experiments with a robust classifier combined with similarity to a reference image can be abused to generate images that are offensive in various ways. In this paper we are only exploring with what is possible, but future work should consider detecting and preventing such abuse.

\subsection{Things that did not work}
We share some negative results we encountered during the process of writing this paper. Our goal is to inform the research community about some methods that failed so that future research can avoid them, reformulate them or even contradict our findings. We also suggest ways to mitigate some of the issues we experienced.

First, we observed that joint optimization of all noise vectors leads to poor visual reconstructions. Even in cases where the MSE loss to the unobserved image was going down, joint-optimization of all noise vectors was giving blurry and/or unrealistic reconstructions. Thus we believe that expanding the generator space without sequential optimization and constraints fails since it makes it too powerful.

We also tried to establish a criterion on how many steps to run per layer. In practice, we observed that a working heuristic is to move to the next layer when the observed MSE error flattens. Even though this idea works well for the first layers, it can lead to unrealistic reconstructions when applied to deeper layers of the generator. To mitigate this issue, we choose very small radii when optimizing in deep layers. Tuning the hyperparameters (learning rates, number of steps and optimization radii) for each of the layers can be a particularly toilsome procedure. Sadly, we observed that these parameters do not generalize across different tasks (even though they mostly generalize across different images).

On the theoretical side, we tried (unsuccessfully) to obtain similar results for an $l_2$ dilation of the range of the first generator. The main bottleneck is the measurements bound; for the $l_2$ ball, we cannot avoid linear dependence on the intermediate dimension. It is not clear yet whether a similar result for the $l_2$ case could be proved. In practice, we did not observe significant difference between projecting in $l_1$ or $l_2$ balls. Moreover, it is known that $l_1$ projection encourages sparsity in some cases, e.g. see LASSO ~\cite{lasso}. Establishing a connection between the $l_0$ and the $l_1$ solutions is left as future work.

\bibliography{references}

\begin{thebibliography}{69}
\providecommand{\natexlab}[1]{#1}
\providecommand{\url}[1]{\texttt{#1}}
\expandafter\ifx\csname urlstyle\endcsname\relax
  \providecommand{\doi}[1]{doi: #1}\else
  \providecommand{\doi}{doi: \begingroup \urlstyle{rm}\Url}\fi

\bibitem[Abadi et~al.(2016)Abadi, Barham, Chen, Chen, Davis, Dean, Devin,
  Ghemawat, Irving, Isard, et~al.]{abadi2016tensorflow}
Abadi, M., Barham, P., Chen, J., Chen, Z., Davis, A., Dean, J., Devin, M.,
  Ghemawat, S., Irving, G., Isard, M., et~al.
\newblock Tensorflow: A system for large-scale machine learning.
\newblock In \emph{12th $\{$USENIX$\}$ symposium on operating systems design
  and implementation ($\{$OSDI$\}$ 16)}, pp.\  265--283, 2016.

\bibitem[Abdal et~al.(2019)Abdal, Qin, and Wonka]{abdal2019image2stylegan}
Abdal, R., Qin, Y., and Wonka, P.
\newblock Image2stylegan: How to embed images into the stylegan latent space?
\newblock In \emph{Proceedings of the IEEE/CVF International Conference on
  Computer Vision}, pp.\  4432--4441, 2019.

\bibitem[Baraniuk et~al.(2008)Baraniuk, Davenport, DeVore, and
  Wakin]{baraniuk2008simple}
Baraniuk, R., Davenport, M., DeVore, R., and Wakin, M.
\newblock A simple proof of the restricted isometry property for random
  matrices.
\newblock \emph{Constructive Approximation}, 28\penalty0 (3):\penalty0
  253--263, 2008.

\bibitem[Bora et~al.(2017)Bora, Jalal, Price, and Dimakis]{bora2017compressed}
Bora, A., Jalal, A., Price, E., and Dimakis, A.~G.
\newblock Compressed sensing using generative models.
\newblock In \emph{International Conference on Machine Learning}, pp.\
  537--546. PMLR, 2017.

\bibitem[Chen et~al.(2008)Chen, Tang, and Leng]{chen2008prior}
Chen, G.-H., Tang, J., and Leng, S.
\newblock Prior image constrained compressed sensing (piccs): a method to
  accurately reconstruct dynamic ct images from highly undersampled projection
  data sets.
\newblock \emph{Medical physics}, 35\penalty0 (2):\penalty0 660--663, 2008.

\bibitem[Dabov et~al.(2006)Dabov, Foi, Katkovnik, and Egiazarian]{bm3d}
Dabov, K., Foi, A., Katkovnik, V., and Egiazarian, K.
\newblock Image denoising with block-matching and 3d filtering.
\newblock In \emph{Image Processing: Algorithms and Systems, Neural Networks,
  and Machine Learning}, volume 6064, pp.\  606414. International Society for
  Optics and Photonics, 2006.

\bibitem[Daras et~al.(2020)Daras, Odena, Zhang, and Dimakis]{ylg}
Daras, G., Odena, A., Zhang, H., and Dimakis, A.~G.
\newblock Your local gan: Designing two dimensional local attention mechanisms
  for generative models.
\newblock In \emph{Proceedings of the IEEE/CVF Conference on Computer Vision
  and Pattern Recognition}, 2020.

\bibitem[Deng et~al.(2009)Deng, Dong, Socher, Li, Li, and
  Fei-Fei]{imagenet_cvpr09}
Deng, J., Dong, W., Socher, R., Li, L.-J., Li, K., and Fei-Fei, L.
\newblock {ImageNet: A Large-Scale Hierarchical Image Database}.
\newblock In \emph{CVPR09}, 2009.

\bibitem[Dhar et~al.(2018)Dhar, Grover, and Ermon]{dhar2018modeling}
Dhar, M., Grover, A., and Ermon, S.
\newblock Modeling sparse deviations for compressed sensing using generative
  models.
\newblock In \emph{International Conference on Machine Learning}, pp.\
  1214--1223. PMLR, 2018.

\bibitem[Duarte et~al.(2008)Duarte, Davenport, Takhar, Laska, Sun, Kelly, and
  Baraniuk]{duarte2008single}
Duarte, M.~F., Davenport, M.~A., Takhar, D., Laska, J.~N., Sun, T., Kelly,
  K.~F., and Baraniuk, R.~G.
\newblock Single-pixel imaging via compressive sampling.
\newblock \emph{IEEE signal processing magazine}, 25\penalty0 (2):\penalty0
  83--91, 2008.

\bibitem[Duchi et~al.(2008)Duchi, Shalev-Shwartz, Singer, and
  Chandra]{duchi2008efficient}
Duchi, J., Shalev-Shwartz, S., Singer, Y., and Chandra, T.
\newblock Efficient projections onto the l 1-ball for learning in high
  dimensions.
\newblock In \emph{Proceedings of the 25th international conference on Machine
  learning}, pp.\  272--279, 2008.

\bibitem[Engstrom et~al.(2019)Engstrom, Ilyas, Salman, Santurkar, and
  Tsipras]{robustness}
Engstrom, L., Ilyas, A., Salman, H., Santurkar, S., and Tsipras, D.
\newblock Robustness (python library), 2019.
\newblock URL \url{https://github.com/MadryLab/robustness}.

\bibitem[G{\"u}era \& Delp(2018)G{\"u}era and Delp]{guera2018deepfake}
G{\"u}era, D. and Delp, E.~J.
\newblock Deepfake video detection using recurrent neural networks.
\newblock In \emph{2018 15th IEEE International Conference on Advanced Video
  and Signal Based Surveillance (AVSS)}, pp.\  1--6. IEEE, 2018.

\bibitem[Hand \& Voroninski(2018)Hand and Voroninski]{hand2018global}
Hand, P. and Voroninski, V.
\newblock Global guarantees for enforcing deep generative priors by empirical
  risk.
\newblock In \emph{Conference On Learning Theory}, pp.\  970--978. PMLR, 2018.

\bibitem[Hand et~al.(2018)Hand, Leong, and Voroninski]{hand2018phase}
Hand, P., Leong, O., and Voroninski, V.
\newblock Phase retrieval under a generative prior.
\newblock \emph{arXiv preprint arXiv:1807.04261}, 2018.

\bibitem[Hegde et~al.(2009)Hegde, Duarte, and Cevher]{hegde2009compressive}
Hegde, C., Duarte, M.~F., and Cevher, V.
\newblock Compressive sensing recovery of spike trains using a structured
  sparsity model.
\newblock In \emph{SPARS'09-Signal Processing with Adaptive Sparse Structured
  Representations}, 2009.

\bibitem[Hinrichs \& Vyb{\'\i}ral(2011)Hinrichs and
  Vyb{\'\i}ral]{hinrichs2011johnson}
Hinrichs, A. and Vyb{\'\i}ral, J.
\newblock Johnson-lindenstrauss lemma for circulant matrices.
\newblock \emph{Random Structures \& Algorithms}, 39\penalty0 (3):\penalty0
  391--398, 2011.

\bibitem[Jain et~al.(2020)Jain, Olmo, Sengupta, Manikonda, and
  Kambhampati]{jain2020imperfect}
Jain, N., Olmo, A., Sengupta, S., Manikonda, L., and Kambhampati, S.
\newblock Imperfect imaganation: Implications of gans exacerbating biases on
  facial data augmentation and snapchat selfie lenses.
\newblock \emph{arXiv preprint arXiv:2001.09528}, 2020.

\bibitem[Kabkab et~al.(2018)Kabkab, Samangouei, and
  Chellappa]{kabkab2018taskaware}
Kabkab, M., Samangouei, P., and Chellappa, R.
\newblock Task-aware compressed sensing with generative adversarial networks.
\newblock In \emph{AAAI}, 2018.

\bibitem[Karras et~al.(2019)Karras, Laine, and Aila]{stylegan}
Karras, T., Laine, S., and Aila, T.
\newblock A style-based generator architecture for generative adversarial
  networks.
\newblock \emph{2019 IEEE/CVF Conference on Computer Vision and Pattern
  Recognition (CVPR)}, Jun 2019.
\newblock \doi{10.1109/cvpr.2019.00453}.
\newblock URL \url{http://dx.doi.org/10.1109/CVPR.2019.00453}.

\bibitem[Karras et~al.(2020)Karras, Laine, Aittala, Hellsten, Lehtinen, and
  Aila]{stylegan2}
Karras, T., Laine, S., Aittala, M., Hellsten, J., Lehtinen, J., and Aila, T.
\newblock Analyzing and improving the image quality of stylegan.
\newblock \emph{2020 IEEE/CVF Conference on Computer Vision and Pattern
  Recognition (CVPR)}, Jun 2020.
\newblock \doi{10.1109/cvpr42600.2020.00813}.
\newblock URL \url{http://dx.doi.org/10.1109/cvpr42600.2020.00813}.

\bibitem[Keys(1981)]{keys1981cubic}
Keys, R.
\newblock Cubic convolution interpolation for digital image processing.
\newblock \emph{IEEE transactions on acoustics, speech, and signal processing},
  29\penalty0 (6):\penalty0 1153--1160, 1981.

\bibitem[Kingma \& Ba(2014)Kingma and Ba]{kingma2014adam}
Kingma, D.~P. and Ba, J.
\newblock Adam: A method for stochastic optimization.
\newblock \emph{arXiv preprint arXiv:1412.6980}, 2014.

\bibitem[Korshunov \& Marcel(2018)Korshunov and Marcel]{korshunov2018deepfakes}
Korshunov, P. and Marcel, S.
\newblock Deepfakes: a new threat to face recognition? assessment and
  detection.
\newblock \emph{arXiv preprint arXiv:1812.08685}, 2018.

\bibitem[Krahmer \& Ward(2011)Krahmer and Ward]{krahmer2011new}
Krahmer, F. and Ward, R.
\newblock New and improved johnson--lindenstrauss embeddings via the restricted
  isometry property.
\newblock \emph{SIAM Journal on Mathematical Analysis}, 43\penalty0
  (3):\penalty0 1269--1281, 2011.

\bibitem[Lee et~al.(2020)Lee, Liu, Wu, and Luo]{celebahq}
Lee, C.-H., Liu, Z., Wu, L., and Luo, P.
\newblock Maskgan: Towards diverse and interactive facial image manipulation.
\newblock In \emph{Proceedings of the IEEE/CVF Conference on Computer Vision
  and Pattern Recognition}, pp.\  5549--5558, 2020.

\bibitem[Lei et~al.(2019)Lei, Jalal, Dhillon, and Dimakis]{lei2019inverting}
Lei, Q., Jalal, A., Dhillon, I.~S., and Dimakis, A.~G.
\newblock Inverting deep generative models, one layer at a time.
\newblock In \emph{NeurIPS}, 2019.

\bibitem[Liu et~al.(2019)Liu, Jiang, Xiao, and Yang]{Liu_2019}
Liu, H., Jiang, B., Xiao, Y., and Yang, C.
\newblock Coherent semantic attention for image inpainting.
\newblock \emph{2019 IEEE/CVF International Conference on Computer Vision
  (ICCV)}, Oct 2019.
\newblock \doi{10.1109/iccv.2019.00427}.
\newblock URL \url{http://dx.doi.org/10.1109/ICCV.2019.00427}.

\bibitem[Liu \& Scarlett(2020)Liu and Scarlett]{inf_theoretic}
Liu, Z. and Scarlett, J.
\newblock Information-theoretic lower bounds for compressive sensing with
  generative models.
\newblock \emph{IEEE Journal on Selected Areas in Information Theory},
  1\penalty0 (1):\penalty0 292–303, May 2020.
\newblock ISSN 2641-8770.
\newblock \doi{10.1109/jsait.2020.2980676}.
\newblock URL \url{http://dx.doi.org/10.1109/JSAIT.2020.2980676}.

\bibitem[Liu et~al.(2018)Liu, Luo, Wang, and Tang]{celeba}
Liu, Z., Luo, P., Wang, X., and Tang, X.
\newblock Large-scale celebfaces attributes (celeba) dataset.
\newblock \emph{Retrieved August}, 15\penalty0 (2018):\penalty0 11, 2018.

\bibitem[Lucas et~al.(2018)Lucas, Iliadis, Molina, and
  Katsaggelos]{lucas2018using}
Lucas, A., Iliadis, M., Molina, R., and Katsaggelos, A.~K.
\newblock Using deep neural networks for inverse problems in imaging: beyond
  analytical methods.
\newblock \emph{IEEE Signal Processing Magazine}, 35\penalty0 (1):\penalty0
  20--36, 2018.

\bibitem[Lustig et~al.(2007)Lustig, Donoho, and Pauly]{lustig2007sparse}
Lustig, M., Donoho, D., and Pauly, J.~M.
\newblock Sparse mri: The application of compressed sensing for rapid mr
  imaging.
\newblock \emph{Magnetic Resonance in Medicine: An Official Journal of the
  International Society for Magnetic Resonance in Medicine}, 58\penalty0
  (6):\penalty0 1182--1195, 2007.

\bibitem[Lustig et~al.(2008)Lustig, Donoho, Santos, and
  Pauly]{lustig2008compressed}
Lustig, M., Donoho, D.~L., Santos, J.~M., and Pauly, J.~M.
\newblock Compressed sensing mri.
\newblock \emph{IEEE signal processing magazine}, 25\penalty0 (2):\penalty0
  72--82, 2008.

\bibitem[Mardani et~al.(2018)Mardani, Gong, Cheng, Vasanawala, Zaharchuk, Xing,
  and Pauly]{mardani2018deep}
Mardani, M., Gong, E., Cheng, J.~Y., Vasanawala, S.~S., Zaharchuk, G., Xing,
  L., and Pauly, J.~M.
\newblock Deep generative adversarial neural networks for compressive sensing
  mri.
\newblock \emph{IEEE transactions on medical imaging}, 38\penalty0
  (1):\penalty0 167--179, 2018.

\bibitem[Matern et~al.(2019)Matern, Riess, and
  Stamminger]{matern2019exploiting}
Matern, F., Riess, C., and Stamminger, M.
\newblock Exploiting visual artifacts to expose deepfakes and face
  manipulations.
\newblock In \emph{2019 IEEE Winter Applications of Computer Vision Workshops
  (WACVW)}, pp.\  83--92. IEEE, 2019.

\bibitem[Menon et~al.(2020)Menon, Damian, Hu, Ravi, and Rudin]{pulse}
Menon, S., Damian, A., Hu, S., Ravi, N., and Rudin, C.
\newblock Pulse: Self-supervised photo upsampling via latent space exploration
  of generative models.
\newblock \emph{2020 IEEE/CVF Conference on Computer Vision and Pattern
  Recognition (CVPR)}, 2020.
\newblock \doi{10.1109/cvpr42600.2020.00251}.
\newblock URL \url{http://dx.doi.org/10.1109/cvpr42600.2020.00251}.

\bibitem[Mousavi et~al.(2019)Mousavi, Dasarathy, and Baraniuk]{mousavi2019data}
Mousavi, A., Dasarathy, G., and Baraniuk, R.~G.
\newblock A data-driven and distributed approach to sparse signal
  representation and recovery.
\newblock In \emph{International Conference on Learning Representations}, 2019.

\bibitem[Nesterov(2003)]{nesterov2003introductory}
Nesterov, Y.
\newblock \emph{Introductory lectures on convex optimization: A basic course},
  volume~87.
\newblock Springer Science \& Business Media, 2003.

\bibitem[Nguyen et~al.(2019)Nguyen, Nguyen, Nguyen, Nguyen, and
  Nahavandi]{nguyen2019deep}
Nguyen, T.~T., Nguyen, C.~M., Nguyen, D.~T., Nguyen, D.~T., and Nahavandi, S.
\newblock Deep learning for deepfakes creation and detection: A survey.
\newblock \emph{arXiv preprint arXiv:1909.11573}, 2019.

\bibitem[Nocedal \& Wright(2006)Nocedal and Wright]{nocedal2006numerical}
Nocedal, J. and Wright, S.
\newblock \emph{Numerical optimization}.
\newblock Springer Science \& Business Media, 2006.

\bibitem[Ongie et~al.(2020)Ongie, Jalal, Metzler, Baraniuk, Dimakis, and
  Willett]{ongie2020deep}
Ongie, G., Jalal, A., Metzler, C.~A., Baraniuk, R.~G., Dimakis, A.~G., and
  Willett, R.
\newblock Deep learning techniques for inverse problems in imaging.
\newblock \emph{IEEE Journal on Selected Areas in Information Theory},
  1\penalty0 (1):\penalty0 39--56, 2020.

\bibitem[Pajot et~al.(2019)Pajot, de~Bezenac, and
  Gallinari]{pajot2018unsupervised}
Pajot, A., de~Bezenac, E., and Gallinari, P.
\newblock Unsupervised adversarial image reconstruction.
\newblock In \emph{International Conference on Learning Representations}, 2019.
\newblock URL \url{https://openreview.net/forum?id=BJg4Z3RqF7}.

\bibitem[Pandit et~al.(2019)Pandit, Sahraee, Rangan, and Fletcher]{Pandit_2019}
Pandit, P., Sahraee, M., Rangan, S., and Fletcher, A.~K.
\newblock Asymptotics of map inference in deep networks.
\newblock \emph{2019 IEEE International Symposium on Information Theory
  (ISIT)}, Jul 2019.
\newblock \doi{10.1109/isit.2019.8849316}.
\newblock URL \url{http://dx.doi.org/10.1109/ISIT.2019.8849316}.

\bibitem[Park et~al.(2020)Park, Smedemark-Margulies, Daniels, Yu, van~de Meent,
  and HAnd]{surgery}
Park, J.~Y., Smedemark-Margulies, N., Daniels, M., Yu, R., van~de Meent, J.-W.,
  and HAnd, P.
\newblock Generator surgery for compressed sensing.
\newblock In \emph{NeurIPS 2020 Workshop on Deep Learning and Inverse
  Problems}, 2020.
\newblock URL \url{https://openreview.net/forum?id=s2EucjZ6d2s}.

\bibitem[Paszke et~al.(2019)Paszke, Gross, Massa, Lerer, Bradbury, Chanan,
  Killeen, Lin, Gimelshein, Antiga, et~al.]{pytorch}
Paszke, A., Gross, S., Massa, F., Lerer, A., Bradbury, J., Chanan, G., Killeen,
  T., Lin, Z., Gimelshein, N., Antiga, L., et~al.
\newblock Pytorch: An imperative style, high-performance deep learning library.
\newblock \emph{arXiv preprint arXiv:1912.01703}, 2019.

\bibitem[Pathak et~al.(2016)Pathak, Krahenbuhl, Donahue, Darrell, and
  Efros]{pathak2016context}
Pathak, D., Krahenbuhl, P., Donahue, J., Darrell, T., and Efros, A.~A.
\newblock Context encoders: Feature learning by inpainting.
\newblock In \emph{Proceedings of the IEEE conference on computer vision and
  pattern recognition}, pp.\  2536--2544, 2016.

\bibitem[Pisier(1986)]{maurey}
Pisier, G.
\newblock Probabilistic methods in the geometry of banach spaces.
\newblock In \emph{Probability and analysis}, pp.\  167--241. Springer, 1986.

\bibitem[Qaisar et~al.(2013)Qaisar, Bilal, Iqbal, Naureen, and
  Lee]{qaisar2013compressive}
Qaisar, S., Bilal, R.~M., Iqbal, W., Naureen, M., and Lee, S.
\newblock Compressive sensing: From theory to applications, a survey.
\newblock \emph{Journal of Communications and networks}, 15\penalty0
  (5):\penalty0 443--456, 2013.

\bibitem[Raj et~al.(2019)Raj, Li, and Bresler]{raj2019ganbased}
Raj, A., Li, Y., and Bresler, Y.
\newblock Gan-based projector for faster recovery with convergence guarantees
  in linear inverse problems.
\newblock 2019.

\bibitem[Richardson et~al.(2020)Richardson, Alaluf, Patashnik, Nitzan, Azar,
  Shapiro, and Cohen-Or]{richardson2020encoding}
Richardson, E., Alaluf, Y., Patashnik, O., Nitzan, Y., Azar, Y., Shapiro, S.,
  and Cohen-Or, D.
\newblock Encoding in style: a stylegan encoder for image-to-image translation.
\newblock \emph{arXiv preprint arXiv:2008.00951}, 2020.

\bibitem[Salminen et~al.(2020)Salminen, Jung, Chowdhury, and
  Jansen]{salminen2020analyzing}
Salminen, J., Jung, S.-g., Chowdhury, S., and Jansen, B.~J.
\newblock Analyzing demographic bias in artificially generated facial pictures.
\newblock In \emph{Extended Abstracts of the 2020 CHI Conference on Human
  Factors in Computing Systems}, pp.\  1--8, 2020.

\bibitem[Santurkar et~al.(2019)Santurkar, Tsipras, Tran, Ilyas, Engstrom, and
  Madry]{santurkar2019image}
Santurkar, S., Tsipras, D., Tran, B., Ilyas, A., Engstrom, L., and Madry, A.
\newblock Image synthesis with a single (robust) classifier.
\newblock \emph{arXiv preprint arXiv:1906.09453}, 2019.

\bibitem[Shah \& Hegde(2018)Shah and Hegde]{Shah_2018}
Shah, V. and Hegde, C.
\newblock Solving linear inverse problems using gan priors: An algorithm with
  provable guarantees.
\newblock \emph{2018 IEEE International Conference on Acoustics, Speech and
  Signal Processing (ICASSP)}, Apr 2018.
\newblock \doi{10.1109/icassp.2018.8462233}.
\newblock URL \url{http://dx.doi.org/10.1109/ICASSP.2018.8462233}.

\bibitem[Song et~al.(2019)Song, Fan, and Lafferty]{song2019surfing}
Song, G., Fan, Z., and Lafferty, J.
\newblock Surfing: Iterative optimization over incrementally trained deep
  networks.
\newblock \emph{NeurIPS}, 2019.

\bibitem[Sudakov(1969)]{sudakov1969gaussian}
Sudakov, V.~N.
\newblock Gaussian measures, cauchy measures and $\varepsilon$-entropy.
\newblock In \emph{Soviet Math. Dokl}, volume~10, pp.\  310--313, 1969.

\bibitem[Sun \& Chen(2020)Sun and Chen]{Sun_2020}
Sun, W. and Chen, Z.
\newblock Learned image downscaling for upscaling using content adaptive
  resampler.
\newblock \emph{IEEE Transactions on Image Processing}, 29:\penalty0
  4027–4040, 2020.
\newblock ISSN 1941-0042.
\newblock \doi{10.1109/tip.2020.2970248}.
\newblock URL \url{http://dx.doi.org/10.1109/TIP.2020.2970248}.

\bibitem[Sun et~al.(2019)Sun, Liu, and Kamilov]{sun2019block}
Sun, Y., Liu, J., and Kamilov, U.~S.
\newblock Block coordinate regularization by denoising.
\newblock \emph{NeurIPS}, 2019.

\bibitem[Sun et~al.(2020)Sun, Liu, and Kamilov]{Sun2020}
Sun, Y., Liu, J., and Kamilov, U.~S.
\newblock Block coordinate regularization by denoising.
\newblock \emph{IEEE Transactions on Computational Imaging}, 6:\penalty0
  908–921, 2020.
\newblock ISSN 2573-0436.
\newblock \doi{10.1109/tci.2020.2996385}.
\newblock URL \url{http://dx.doi.org/10.1109/TCI.2020.2996385}.

\bibitem[Tan et~al.(2020)Tan, Shen, and Zhou]{tan2020improving}
Tan, S., Shen, Y., and Zhou, B.
\newblock Improving the fairness of deep generative models without retraining,
  2020.

\bibitem[Tian et~al.(2020)Tian, Fei, Zheng, Xu, Zuo, and Lin]{Tian_2020}
Tian, C., Fei, L., Zheng, W., Xu, Y., Zuo, W., and Lin, C.-W.
\newblock Deep learning on image denoising: An overview.
\newblock \emph{Neural Networks}, 131:\penalty0 251–275, Nov 2020.
\newblock ISSN 0893-6080.
\newblock \doi{10.1016/j.neunet.2020.07.025}.
\newblock URL \url{http://dx.doi.org/10.1016/j.neunet.2020.07.025}.

\bibitem[Tibshirani(1996)]{lasso}
Tibshirani, R.
\newblock Regression shrinkage and selection via the lasso.
\newblock \emph{Journal of the Royal Statistical Society: Series B
  (Methodological)}, 58\penalty0 (1):\penalty0 267--288, 1996.

\bibitem[Tripathi et~al.(2018)Tripathi, Lipton, and
  Nguyen]{tripathi2018correction}
Tripathi, S., Lipton, Z.~C., and Nguyen, T.~Q.
\newblock Correction by projection: Denoising images with generative
  adversarial networks.
\newblock \emph{arXiv preprint arXiv:1803.04477}, 2018.

\bibitem[Wainwright(2019)]{wainwright2019high}
Wainwright, M.~J.
\newblock \emph{High-dimensional statistics: A non-asymptotic viewpoint},
  volume~48.
\newblock Cambridge University Press, 2019.

\bibitem[Wu et~al.(2019)Wu, Rosca, and Lillicrap]{wu2019deep}
Wu, Y., Rosca, M., and Lillicrap, T.
\newblock Deep compressed sensing, 2019.

\bibitem[Xiao et~al.(2018)Xiao, Li, Zhu, He, Liu, and Song]{xiao2018generating}
Xiao, C., Li, B., Zhu, J.-Y., He, W., Liu, M., and Song, D.
\newblock Generating adversarial examples with adversarial networks.
\newblock \emph{arXiv preprint arXiv:1801.02610}, 2018.

\bibitem[Yang et~al.(2020)Yang, Ding, Chen, and Li]{yang2020defending}
Yang, C., Ding, L., Chen, Y., and Li, H.
\newblock Defending against gan-based deepfake attacks via transformation-aware
  adversarial faces.
\newblock \emph{arXiv preprint arXiv:2006.07421}, 2020.

\bibitem[Yang et~al.(2019)Yang, Zhang, Tian, Wang, Xue, and Liao]{yang2019deep}
Yang, W., Zhang, X., Tian, Y., Wang, W., Xue, J.-H., and Liao, Q.
\newblock Deep learning for single image super-resolution: A brief review.
\newblock \emph{IEEE Transactions on Multimedia}, 21\penalty0 (12):\penalty0
  3106--3121, 2019.

\bibitem[Yu et~al.(2018)Yu, Lin, Yang, Shen, Lu, and Huang]{yu2018generative}
Yu, J., Lin, Z., Yang, J., Shen, X., Lu, X., and Huang, T.~S.
\newblock Generative image inpainting with contextual attention.
\newblock In \emph{Proceedings of the IEEE conference on computer vision and
  pattern recognition}, pp.\  5505--5514, 2018.

\bibitem[Yu et~al.(2019)Yu, Lin, Yang, Shen, Lu, and Huang]{Yu_2019}
Yu, J., Lin, Z., Yang, J., Shen, X., Lu, X., and Huang, T.
\newblock Free-form image inpainting with gated convolution.
\newblock \emph{2019 IEEE/CVF International Conference on Computer Vision
  (ICCV)}, Oct 2019.
\newblock \doi{10.1109/iccv.2019.00457}.
\newblock URL \url{http://dx.doi.org/10.1109/ICCV.2019.00457}.

\end{thebibliography}
\bibliographystyle{icml2021}
\end{document}